\documentclass[10pt,twocolumn,letterpaper]{article}

\usepackage{cvpr}              % To produce the CAMERA-READY version

\usepackage{graphicx}
\usepackage{amsmath}
\usepackage{amssymb}
\usepackage{booktabs}
\usepackage{multicol,multirow}
\usepackage[pagebackref,breaklinks,colorlinks,bookmarks=false]{hyperref}

\DeclareMathOperator*{\argmin}{arg\, min}

\def\cvprPaperID{9349} % *** Enter the CVPR Paper ID here
\def\confName{CVPR}
\def\confYear{2022}

\usepackage{xcolor}
\usepackage[ruled,vlined,linesnumbered]{algorithm2e}

\begin{document}
\title{Image Based Reconstruction of Liquids from 2D Surface Detections} %
\author{Florian Richter, Ryan K. Orosco, and Michael C. Yip \\
University of California San Diego\\
{\tt\small \{frichter, rorosco,  yip\}@ucsd.edu}
}

\maketitle
\thispagestyle{empty}
\appendix

\begin{abstract}
In this work, we present a solution to the challenging problem of reconstructing liquids from image data.
The challenges in reconstructing liquids, which is not faced in previous reconstruction works on rigid and deforming surfaces, lies in the inability to use depth sensing and color features due the variable index of refraction, opacity, and environmental reflections.
Therefore, we limit ourselves to only surface detections (i.e. binary mask) of liquids as observations and do not assume any prior knowledge on the liquids properties.
A novel optimization problem is posed which reconstructs the liquid as particles by minimizing the error between a rendered surface from the particles and the surface detections while satisfying liquid constraints.
Our solvers to this optimization problem are presented and no training data is required to apply them.
We also propose a dynamic prediction to seed the reconstruction optimization from the previous time-step.
We test our proposed methods in simulation and on two new liquid datasets which we open source\footnote{Will release upon publication} so the broader research community can continue developing in this under explored area.
\end{abstract}

\section{Introduction}

To successfully navigate in and interact with the 3D world we live in, a 3D geometric understanding is required.
The importance of this requirement can be seen by the numerous advancements in reconstruction methods from cameras, which is the ideal sensor due to its information richness and cheap cost.
Solutions for surface based reconstruction have been proposed for a variety of scenarios such as rigid, unknown environments \cite{newcombe2011kinectfusion} with dynamic objects \cite{keller2013real}.
The rigidness assumption has also been lifted to handle deformable surfaces \cite{newcombe2015dynamicfusion, innmann2016volumedeform}.
Breakthrough developments from the reconstruction community have fed into downstream applications such as robotic manipulation \cite{varley2017shape} and surgical tissue tracking \cite{li2020super}.

\begin{figure}[t]
    \centering
    \includegraphics[trim=0cm 0cm 12cm 0cm, clip, width=0.48\linewidth]{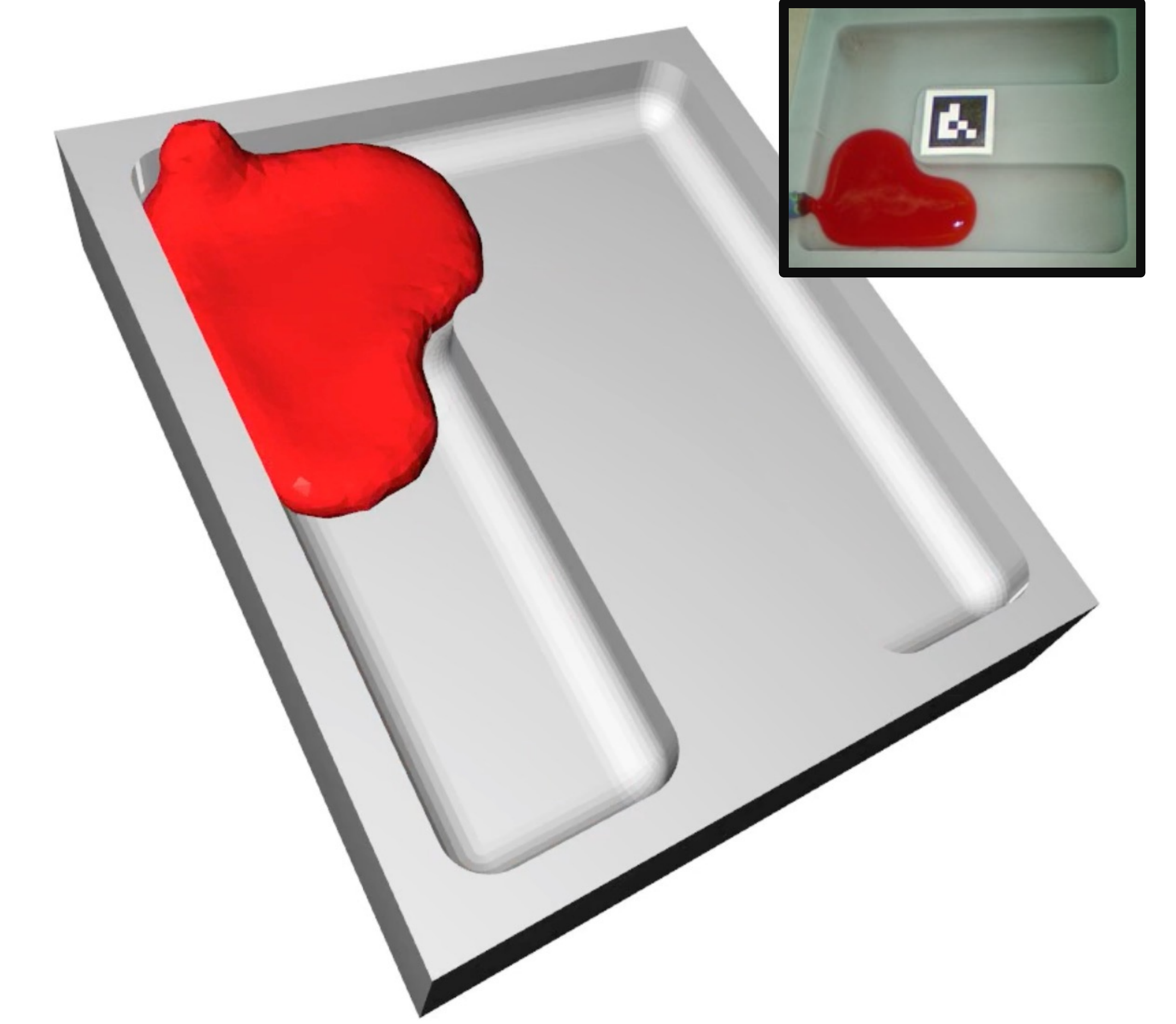}
    \includegraphics[trim=0cm 0cm 10cm 0cm, clip, width=0.48\linewidth]{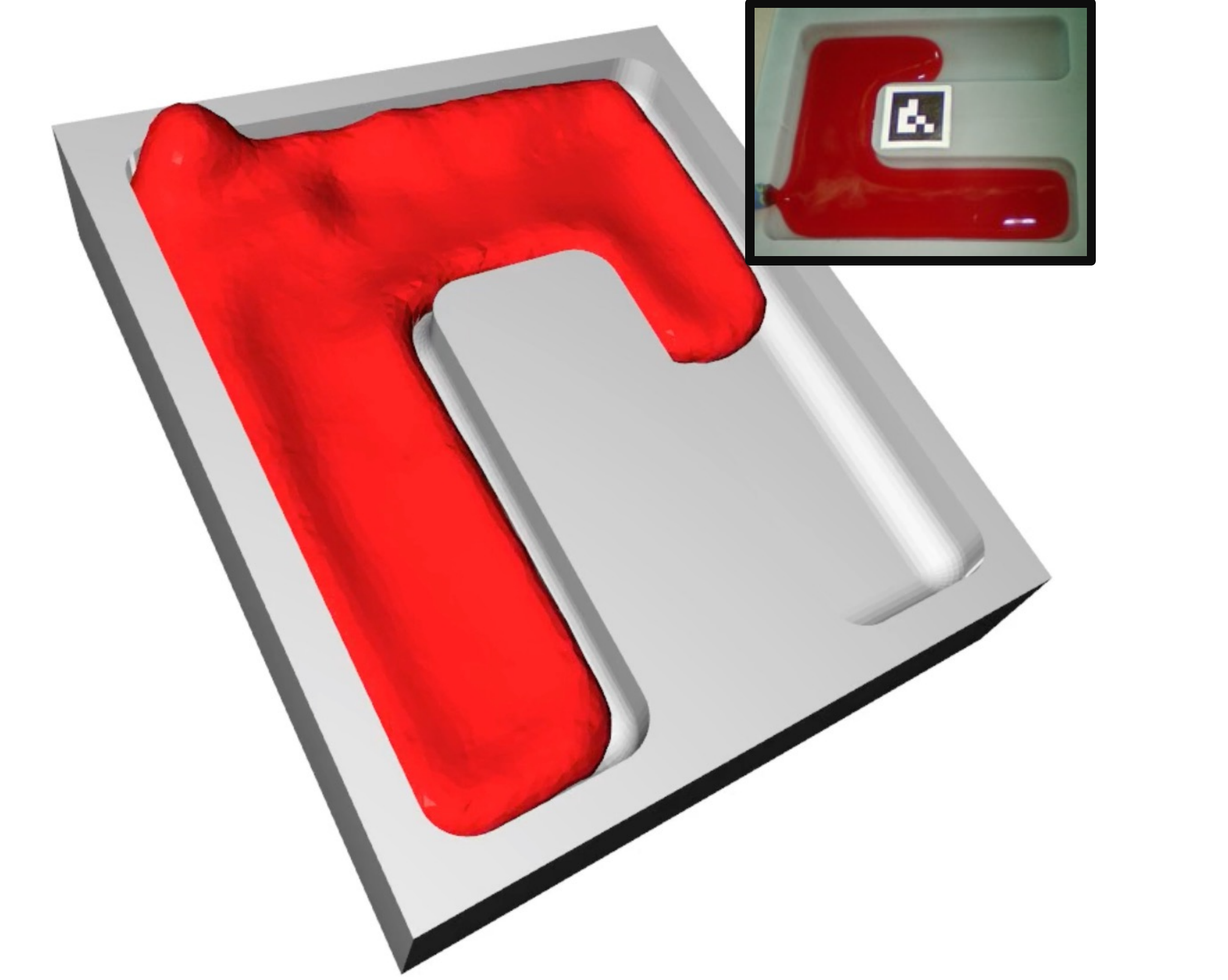}
    \includegraphics[trim=0cm 0cm 12cm 0cm, clip, width=0.48\linewidth]{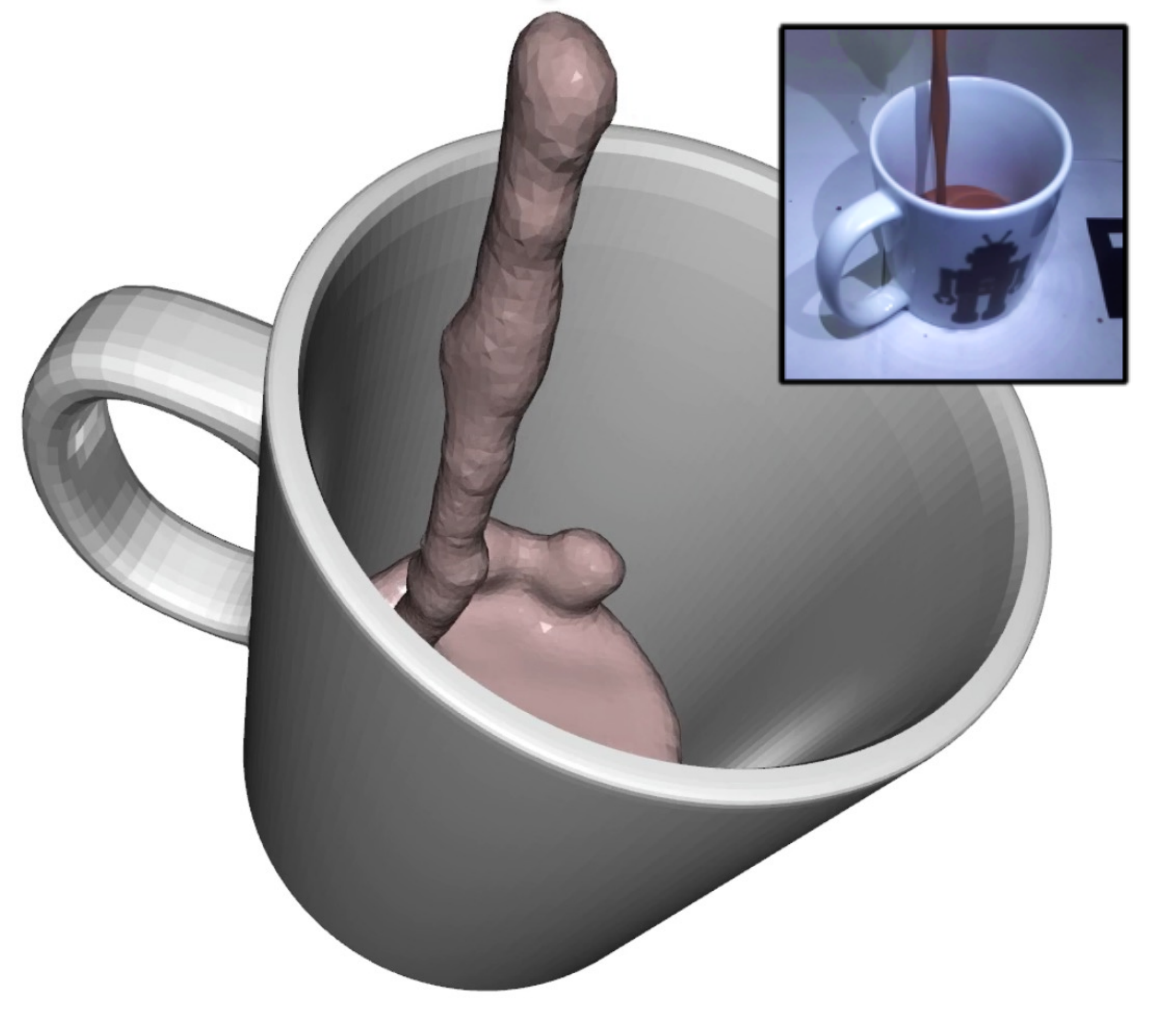}
    \includegraphics[trim=0cm 0cm 10cm 0cm, clip, width=0.48\linewidth]{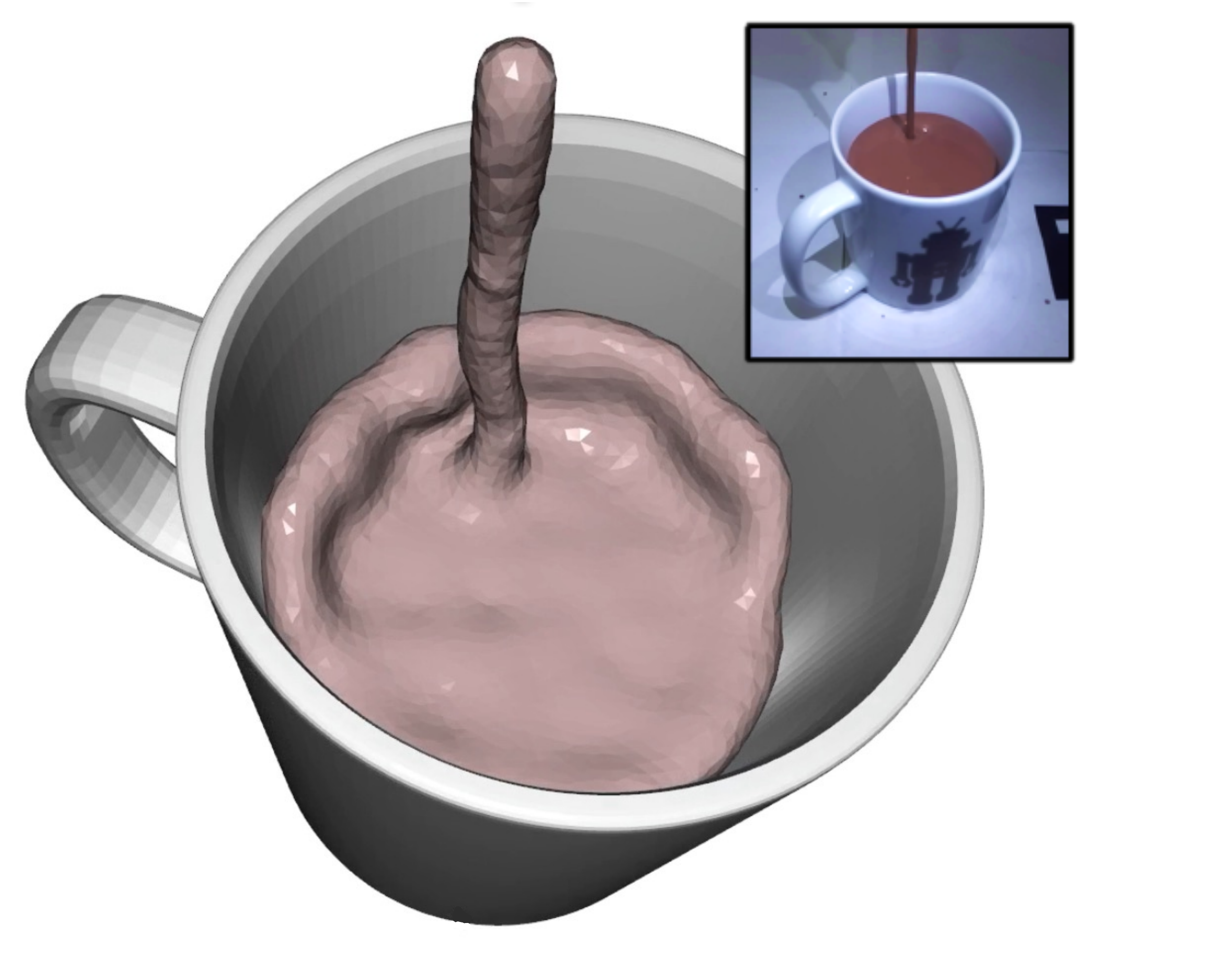}
    \caption{The top and bottom row figures shows the output of our proposed method for reconstructing liquid from an endoscopic camera and a human pouring chocolate milk into a cup respectively.
    Our reconstruction approach minimizes the 2D surface detection loss while simultaneously satisfying liquid constraints without the need for any prior training data.
    The result is an effective reconstruction technique for liquids that has been validated on simulated and real-life data as shown here.
    }
    \label{fig:cover}
\end{figure}

Reconstruction of more complex scenes, such as fluids however remains an under explored area.
Fluids, unlike rigid and deforming objects, are typically turbulent and can exhibit translating, shearing, and rotation motions \cite{pope2001turbulent}.
The well established Navier-Stokes equations which describe fluid motion have been applied to generate effective graphic renderings of fluids \cite{bridson2015fluid}.
The motions of fluids also differs depending on if it is a gas or liquid.
Gasses are compressible and reconstruction from images has been explored \cite{franz2021global}.
Liquids, unlike gasses, are in-compressible and for everyday human interactions, rely on a container and gravity to form their shape (e.g. a mug holding coffee).
By fully reconstructing liquids in 3D, automation efforts which replicate human tasks interacting with liquids can be significantly improved such as robot bar tending \cite{wu2020can}, autonomous blood suction during surgeries \cite{huang2021model}, and sewage service \cite{truong2011study}.
However, the challenge of reconstructing liquids from images remained unexplored and simplifying heuristics or end-to-end models were used to guide these automation efforts.

% We propose an approach to reconstruct liquids only using the same knowledge humans do.
We propose an approach to reconstruct and track liquids from videos using minimal information.
This results in the first technique to reconstruct liquids with only knowledge of the collision environment, gravity direction, and 2D surface detections.
The observations are limited to 2D surface detections (i.e. binary mask) because a liquids color varies widely based on their refraction index, opacity, and environment.
Furthermore, common depth sensors (e.g. Microsoft's Kinect or Intel's RealSense) will behave inconsistently due to the unknown refraction index.
By limiting the observation data to only 2D surface detections, our proposed reconstruction method can be directly applied to any detected liquid and does not require any prior information on the liquid (e.g. no training data is required).
To this end, our contributions are:
\begin{enumerate}
    \item a novel optimization problem for reconstructing liquids with a particle representation which accounts for liquid constraints,
    \item seeding the optimization with a dynamics prediction based on the previous time-step,
    \item and a branching strategy to dynamically adjust the number of particles in the reconstructed liquid.
    % \item and estimation of the liquid emission source in the reconstructed scene.
\end{enumerate}
The complete solution only relies on sequential data and was extended with a source estimation technique to show its adaptability for future applications with liquids.
To baseline our proposed method, new liquid datasets are collected and open sourced so future researchers can further develop in this under-explored area.

\section{Related Works}

\subsection{Fluid Reconstruction}

Several sensor modalities have been used historically to capture fluid flow in science and engineering.
Schlieren imaging \cite{dalziel2000whole, atcheson2008time, atcheson2009evaluation}, Particle Image Velocimetry \cite{grant1997particle}, laser scanners \cite{hawkins2005acquisition}, and structured light \cite{gu2012compressive} have all been developed for capturing fluids.
These specialized sensors however are not common place and often expensive, hence making them less ideal than visible spectrum sensors.
This lead to a lot of developement in the field of visible light tomography where a combination of 2D image projections of a fluid are used to reconstruct it in 3D.
Recent developments in the field have effectively registered fluids with simulation based fluid dynamics \cite{eckert2019scalarflow} and require only a few camera perspectives for effective reconstruction \cite{zang2020tomofluid, franz2021global}.
These approaches however do not consider in-compressible fluids, liquids, and only focus on gasses in free space (i.e. no collision).

\subsection{Liquid Detection and Simulation Registration}

While direct reconstruction of liquids has not been done before, there has been work in detecting liquids in the image frame and registering with a simulation.
Pools of water have been detected for unmanned ground vehicles \cite{rankin2010daytime, rankin2011daytime}, and flowing blood has been detected during surgeries for autonomous, robotic suction \cite{richter2021autonomous}.
Liquids during a pouring task have also been detected using optical flow \cite{yamaguchi2016stereo} and Deep Neural Networks \cite{schenck2018perceiving}.
The scope of this paper is on reconstructing liquids, and these detection methods could be utilized to feed into our proposed method by supplying the observations of the liquids surface.
Mottaghi et al. were able to estimate a liquid's volume in a container from images directly \cite{mottaghi2017see}.
Registration of a liquid simulation with the real world has also been conducted for robot pouring \cite{ guevara2017adaptable, schenck2017visual, schenck2018spnets}.
However, these techniques require prior information about the liquid being reconstructed, such as knowing the volume of the liquid before hand.
Meanwhile in this work, we only assume prior knowledge of the gravity direction and collision environment and use a novel branching strategy to dynamically adjust the volume of the reconstructed liquid.
Nevertheless, we integrated Schenck and Fox's most recent simulation registration work \cite{schenck2018spnets} to the best of our ability into our reconstruction approach for comparison.

\section{Methods}

\begin{figure}[t]
    \centering
    \vspace{2mm}
    \includegraphics[trim=2.35cm 8.1cm 22.9cm 2.5cm, clip, width=0.9\linewidth]{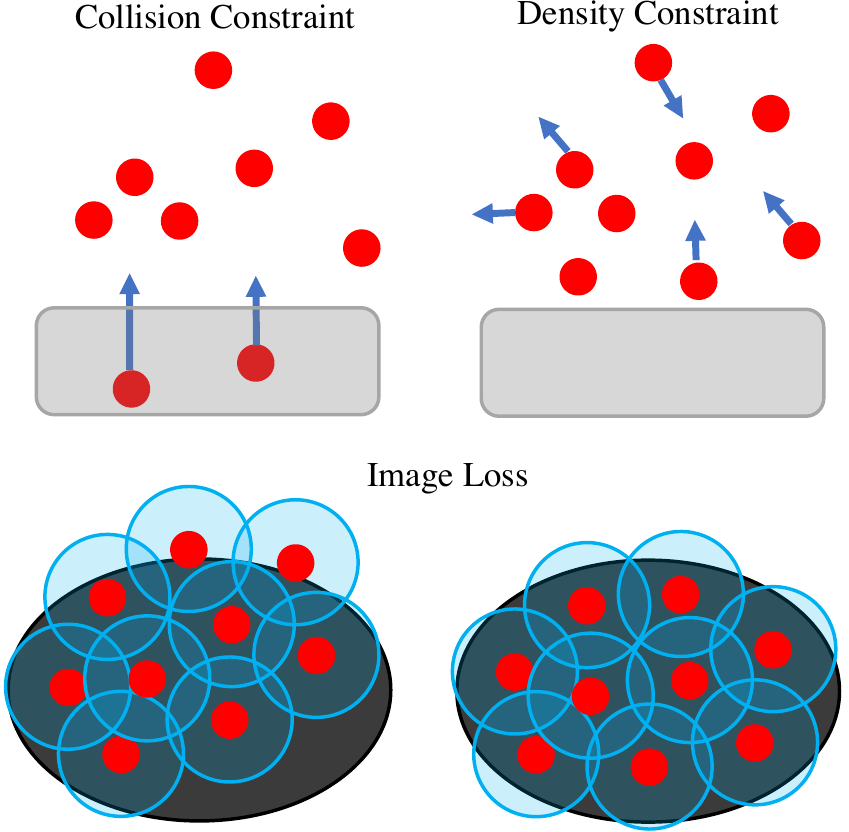}
    \caption{A visualization of solving (\ref{eq:optimization_problem}) in order of top-left, top-right, and bottom where the particle locations are drawn in red. The collision constraint pushes particles out of collision (drawn in light grey), the density constraints ensures incompressibility for liquids by maintaining a constant density, and finally the image loss between the detected surface (drawn in black) and a surface rendering (drawn in semi-transparent blue) is minimized.}
    \label{fig:constraints_visualize}
\end{figure}

 \begin{algorithm}[t]
    \caption{Reconstruct Liquid at time $t$}
    \label{alg:main_outline}
    \SetKwInOut{Input}{Input}
    \SetKwInOut{Output}{Output}
    \Input{Previous liquid particle positions and velocities, $\mathbf{p}_{t-1}, \mathbf{v}_{t-1}$, and image, $\mathbb{I}_t$}
    \Output{Updated liquid particle positions and velocities $\mathbf{p}_{t}, \mathbf{v}_{t}$}
    \tcp{Particle Prediction}
    \label{alg:equation_of_motion}
    $\mathbf{p}_t \leftarrow \mathbf{p}_{t-1} + \mathbf{v}_{t-1} \Delta t + \frac{1}{2} \mathbf{g} \Delta t^2$\\
    % \tcp{Insert source particles}
    % $\mathbf{p}_{s} \leftarrow getSourceParticles(\hat{\mathbf{s}}_t, \hat{f}_t)$ \\
    % \label{alg:insert_source_particles_1}
    % $\mathbf{p}_t \leftarrow [\mathbf{p}_t, \mathbf{p}_{s}]$\\
    \label{alg:insert_source_particles_2}
    \For{$n_o$ iterations}{
    \For{$n_j$ iterations}{
    \tcp{Apply Position Constraints}
    \For{$n_c$ iterations}{
        \label{alg:start_of_optimization}
        $\Delta \mathbf{p}_c \leftarrow solveCollision(\mathbf{p}_t)$\\
        \label{alg:collision_constraint_1}
        $\mathbf{p}_t \leftarrow \mathbf{p}_t + \Delta \mathbf{p}_c$ \\
        \label{alg:collision_constraint_2}
        $\Delta \mathbf{p}_\rho \leftarrow solveDensity(\mathbf{p}_t)$\\
        \label{alg:density_constraint_1}
        $\mathbf{p}_t \leftarrow \mathbf{p}_t + \Delta \mathbf{p}_\rho$ \\
        \label{alg:density_constraint_2}
        }
    \tcp{Minimize Image Loss}
    \For{$n_i$ iterations}{
        $\hat{\mathbb{I}}(\mathbf{p}_t) \leftarrow
        renderSurface(\mathbf{p}_t)$\\
        \label{alg:render_surface}
        $\mathbf{p}_t \leftarrow \mathbf{p}_t +  \alpha_\mathbb{I} \left( \partial \mathcal{L}\left(\mathbb{I}_t, \hat{\mathbb{I}}(\mathbf{p}_t)\right) / \partial \mathbf{p}_{t} \right)$\\
        \label{alg:optimize_image_loss}
        }
    }
    \tcp{Adjust Particle Count}
    \uIf{local\_minima\_conditions}{
        \label{alg:duplicate_remove_particles_1}
        $\mathbf{p}_t \leftarrow duplicateOrRemoveParticle(\mathbf{p}_t)$
        \label{alg:duplicate_remove_particles_2}
    }
    }
    \tcp{Update Particle Velocities}
    $\mathbf{v}_{t} \leftarrow \left( \mathbf{p}_{t} - \mathbf{p}_{t-1} \right)/ \Delta t$ \\
    \label{alg:compute_induced_velocity}
    $\mathbf{v}_{t} \leftarrow dampVelocityAndApplyViscocity(\mathbf{p}_{t}, \mathbf{v}_{t})$\\
    \label{alg:dampen_and_viscocity}
    \Return{ $\mathbf{p}_t, \mathbf{v}_t$ }
 \end{algorithm}

Let $\mathbf{p}_t = \{ \mathbf{p}^i_t \}_{i=1}^N$ be the set of particles in $\mathbb{R}^3$ representing the reconstructed liquid at time $t$.
To estimate the particle locations, and hence reconstruct the liquid, we assume only knowledge of a binary masked image which identified the liquids surface, $\mathbb{I}_t$.
The estimation for the particles is done by minimizing a loss between the detected surface and a reconstruction of the liquid surface from the particles, $\hat{\mathbb{I}}(\cdot)$.
Written explicitly, the optimization problem is:
\begin{equation}
\begin{split}
        \argmin_{ \mathbf{p}_t } \mathcal{L}\left( \mathbb{I}_t, \hat{\mathbb{I}}(\mathbf{p}_t) \right) \;\;\;\;
        \text{s.t. } \mathbf{C}(\mathbf{p}_t) = 0
\end{split}
\label{eq:optimization_problem}
\end{equation}
where liquid constraints, $\mathbf{C}(\cdot)$ , are applied to the particles positions so they behave like a liquid.
The position constraints considered here are density and collision, and a visual explanation is shown in Fig. \ref{fig:constraints_visualize}.
Solving position constraints and deriving velocities from them has produced stable, particle based simulations for large time-step sizes \cite{muller2007position, macklin2013position}.
% Position constraints and velocity time-step prediction  has produced, stable, liquid-like simulation for graphics .
Similarly, we leverage the liquid-like dynamics induced by position constraints for effective liquid reconstruction from video sequences (i.e. going from $t$ to $t+1$).

The following methods detail our solution to the optimization problem shown in (\ref{eq:optimization_problem}) and an outline is shown in Algorithm \ref{alg:main_outline}.
First, the position constraints, $\mathbf{C}(\cdot)$, and their respective solvers are described.
Second, the rendered surface, $\hat{\mathbb{I}}(\cdot)$ and its gradient with respective to the particle positions to minimize the loss is explained.
The constraint solvers and surface loss gradient are applied in a projective gradient descent scheme to solve (\ref{eq:optimization_problem}) as shown in lines \ref{alg:start_of_optimization} to \ref{alg:optimize_image_loss} of Algorithm \ref{alg:main_outline}.
Third, finding the number of particles, $N$, to reconstruct the liquid and a strategy of where to add or remove the particles is detailed.
Lastly, prediction of the particles from time-step $t$ to $t+1$ is defined to reconstruct from videos of detected liquids, $\mathbb{I}_1, \dots, \mathbb{I}_{T}$.

\subsection{Position Constraints for Liquid Particles}

The two position constraints used to reconstruct the liquid when optimizing (\ref{eq:optimization_problem}) are collision and density.
The collision constraint ensures that none of the particles representing the reconstructed liquid are in collision with the scene.
Let $C_c(\cdot)$ be the collision constraint for a particle, and it is expressed as:
\begin{equation}
    C_c(\mathbf{p}^i) = \text{relu}( - SDF(\mathbf{p}^i) )
    \label{eq:collision_constraint}
\end{equation}
where $\text{relu}(\cdot)$ is the rectified linear unit function and $SDF(\cdot)$ is the signed distance function of the scene.
The collision constraint is satisfied when it is at 0, which occurs by having all of the particles out of collision (i.e. no more negative $SDF$ values at the particle positions).

To push the particles out of collision and satisfy the collision constraint, finite difference is used to approximate a gradient of (\ref{eq:collision_constraint}) and the particles are moved along the gradient step.
This is computed for particle $\mathbf{p}^i$ as follows:
\begin{equation}
    \label{eq:collision_constraint_solution}
    \Delta \mathbf{p}^i_c = C_c(\mathbf{p}^i) \sum \limits_{\mathbf{k} \in K} w_k SDF(\mathbf{p}^i + d \mathbf{k}) 
\end{equation}
where $K$ is the set of finite sample directions (e.g. [$\pm$1, 0, 0], [0, $\pm$1, 0], [0, 0, $\pm$1]), $w_k$ is the finite difference weight, and $d$ is the steps size for the sample directions.
The finite difference weights are computed optimally \cite{fornberg1988generation} and scaled such that the resulting vector from the summation is normalized.
The normalization is done so the particles are moved up to the current collision depth, $C_c(\mathbf{p}^i)$, and not in collision free space.
The collision constraint is iteratively solved and applied to the particles as shown in lines \ref{alg:collision_constraint_1} and \ref{alg:collision_constraint_2} in Algorithm \ref{alg:main_outline}.

The second constraint, density, ensures that the liquid is in-compressible.
The density of particle based representations for liquids can be expressed using the same technique as Smoothed Particle Hydrodynamics (SPH) \cite{gingold1977smoothed, lucy1977numerical}.
SPH simulations compute physical properties from hydrodynamics, such as density, using interpolation techniques with kernel operators centered about the particle locations.
Similarly, we compute the density at particle $\mathbf{p}^i$
\begin{equation}
    \rho^i(\mathbf{p}) = \sum \limits_{j=1}^N W( || \mathbf{p}^i - \mathbf{p}^j ||, h)
    \label{eq:density_expression}
\end{equation}
where $W(\cdot, h)$ is a smoothing kernel operator with radius $h$.
This is the same as SPH simulations except without the mass term because each particle is set to represent an equal amount of mass in the reconstructed liquid.
A density constraint for the $i$-th particle using (\ref{eq:density_constraint}) can be written as:
\begin{equation}
    C^i_{\rho}(\mathbf{p}) = \frac{\rho_i(\mathbf{p})}{\rho_0} - 1
    \label{eq:density_constraint}
\end{equation}
where $\rho_0$ is the resting density of the liquid being reconstructed \cite{bodin2011constraint}.
This density constraint is satisfied at 0 which occurs when the reconstructed liquid is achieves resting density at each of the particle locations.
% Since each particle represents an equal amount of liquid, the resting density, $\rho_0$, can be interpreted as the resolution of the reconstruction (e.g. increasing $\rho_0$ implies more particles are needed to represent the same liquid).

Newton steps along the constraint's gradient are iteratively taken to satisfy the density constraint in (\ref{eq:density_constraint}).
Each Newton step, $\Delta \mathbf{p}_{\rho}$, is calculated as: 
\begin{equation}
    \label{eq:density_newton_step}
    \Delta \mathbf{p}_{\rho} = - \nabla \mathbf{C}_{\rho} (\mathbf{p}) \left( \nabla \mathbf{C}^\top_{\rho}( \mathbf{p}) \nabla \mathbf{C}_{\rho}( \mathbf{p}) + \epsilon_\rho \mathbf{I} \right)^{-1} \mathbf{C}_{\rho}( \mathbf{p})
\end{equation}
where $\mathbf{C}_{\rho} (\cdot) = [C^1_{\rho}(\cdot), \dots, C^N_{\rho}(\cdot)]^\top$, the partials are
\begin{equation}
\begin{split}
  \frac{ \partial C^i_{\rho} (\mathbf{p})} { \partial \mathbf{p}^i } &= \frac{1}{\rho_0} \sum \limits_{j = 1}^N  \frac{ (\mathbf{p}^i - \mathbf{p}^j)  }{|| \mathbf{p}^i - \mathbf{p}^j || } W'(||\mathbf{p}^i - \mathbf{p}^j ||, h) \\
  \frac{ \partial C^i_{\rho} (\mathbf{p})} { \partial \mathbf{p}^j } &= \frac{ (\mathbf{p}^i - \mathbf{p}^j)  }{\rho_0 || \mathbf{p}^i - \mathbf{p}^j || } W'(||\mathbf{p}^i - \mathbf{p}^j ||, h)
  \end{split}
\end{equation}
where $W'( \cdot, h)$ is the derivative of smoothing kernel operator in (\ref{eq:density_expression}), and $\epsilon_\rho\mathbf{I} \in \mathbb{R}^{N \times N}$ stabilizes the inversion with a damping factor $\epsilon_\rho$.
Enforcing incompressibility in SPH simulations, similar to the proposed density constraint here, when particles have a small number of neighbors is known to cause particle clustering \cite{monaghan2000sph}.
Therefore, we use Monaghan's solution by adding the following artificial pressure term to $\Delta \mathbf{p}_\rho$:
\begin{equation}
    \mathbf{s}_{corr}^i = -\frac{\lambda_s}{\rho_0} \sum \limits_{j=1}^N \left( \frac{W(||\mathbf{p}^i - \mathbf{p}^j||, h)}{W(\lambda_\mathbf{p}, h)} \right)^{\lambda_n} \frac{ \partial C^i_{\rho} (\mathbf{p})} { \partial \mathbf{p}^j }
\end{equation}
for the $i$-th particle where $\lambda_s, \lambda_\mathbf{p}, \lambda_n$ are set according to the original work \cite{monaghan2000sph}.
% After a few iterations of adjusting the liquid particle locations with $\Delta \mathbf{p}_\rho$, the density constraint can be consistently satisfied.
The density constraint is iteratively solved with the artificial pressure term and applied to the particles as shown in lines \ref{alg:density_constraint_1} and \ref{alg:density_constraint_2} in Algorithm \ref{alg:main_outline}.

\subsection{Differentiable Liquid Surface Rendering}

The loss being minimized in (\ref{eq:optimization_problem}) to reconstruct the liquid is between the detected surface, $\mathbb{I}$, and the reconstructed surface, $\hat{\mathbb{I}}(\cdot)$.
This is equivalent to the differentiable rendering problem formulation, which multiple solutions have been proposed for \cite{kato2020differentiable}.
The differentiable renderer we employ is Pulsar which renders each particle as a sphere \cite{lassner2021pulsar} because it is currently state-of-the-art for point-based geometry rendering and requires no training data to get a gradient of the rendered image with respect to the particle locations when not using its shader.
The loss used to minimize the difference between the detected surface and rendered surface is the Symmetric Mean Absolute Percentage Error (SMAPE):
\begin{equation}
    \mathcal{L}\left(\mathbb{I}, \hat{\mathbb{I}}(\mathbf{p} ) \right)= \frac{1}{N_p} \sum \limits_{u,v \in \mathbb{I}} \frac{|\mathbb{I}_{u,v} - \hat{\mathbb{I}}_{u,v}(\mathbf{p})|}{|\mathbb{I}_{u,v}| + |\hat{\mathbb{I}}_{u,v}(\mathbf{p})| + \epsilon_s}
\end{equation}
where $N_p$ is the number of pixels on the image and $\epsilon_s$ is used to stabilize the division.
SMAPE was chosen because the $\ell$-1 loss was used in the original Pulsar work \cite{lassner2021pulsar} and SMAPE is a symmetric version of an $\ell$-1 loss.
In Algorithm \ref{alg:main_outline}, lines \ref{alg:render_surface} and \ref{alg:optimize_image_loss} are where the differentiable renderer is integrated into our reconstruction technique with a gradient step size of $\alpha_\mathbb{I}$.

\subsection{Adding and Removal of Particles}
\label{section:adding_removing_particles}

The number of particles $N$ must be found to solve (\ref{eq:optimization_problem}), hence making this a mixed-integer optimization problem.
To solve for $N$, we use a branching strategy based on the following heuristic: if the rendered surface area is smaller than the detected surface area, duplicate a particle, $N+1$, and vice-versa to remove a particle, $N-1$.
The branching strategy is enabled after confirming a local-minima has been reached with the current number of particles.
This is determined by taking the mean image loss gradient and checking if it less than a threshold:
\begin{equation}
    \frac{1}{N} \sum \limits_{k=1}^{N} \left| \left| \frac{\partial \mathcal{L}\left(\mathbb{I}, \hat{\mathbb{I}}(\mathbf{p}) \right)}{\partial \mathbf{p}^k} \right| \right| \leq \gamma_s
\end{equation}
where $\gamma_s$ is the threshold and if the Intersection over Union (IoU) is less than a threshold:
\begin{equation}
    \frac{\mathbb{I} \cup \hat{\mathbb{I}}(\mathbf{p}) }{\mathbb{I} \cap \hat{\mathbb{I}}(\mathbf{p})} \leq \gamma_{I}
\end{equation}
where $\gamma_{I}$ is the threshold.
IoU is chosen over the SMAPE loss because Pulsar renders each sphere with a blending value so the rendered image will have values from $[0,1]$ hence increasing the SMAPE loss as more spheres are rendered even when the spheres make a perfect silhouette fit.
Meanwhile IoU directly measures silhouette fit which is in line with our heuristic for the branching strategy.
% The IoU metric is used rather than SMAPE loss because SMAPE with Pulsar will always output 
If these two criteria are satisfied, a local-minima due to the number of particles is assumed, and the branching decision of duplicating or removing a particle is triggered.
This branching logic is handled in lines \ref{alg:duplicate_remove_particles_1} and \ref{alg:duplicate_remove_particles_2} in Algorithm \ref{alg:main_outline}.

When duplicating or removing a particle, the collision constraint will remain unchanged and the density constraint will be increased when duplicating a particle and decreased when removing a particle.
Therefore, the particle selected to duplicate or remove is chosen to best satisfy the density constraint so the initial particle locations when solving (\ref{eq:optimization_problem}) after adjusting the particle count remains closest to the density constraint manifold.
Written explicitly and using the $\ell$-1 loss to describe closeness to the constraint manifold, the index of the particle to duplicate or remove is found by solving
\begin{equation}
    \label{eq:optimization_selecting_particle}
    \argmin_i \sum \limits_{k=1}^N |C^{k+}_\rho (\mathbf{p})| + |C^{i+}_\rho (\mathbf{p})| \;\;\;\; \argmin_i \sum \limits_{k \neq i}^N | C^{k-}_\rho (\mathbf{p}) |
\end{equation}
for duplication and removal respectively and $C^{k+}_\rho(\cdot)$, $C^{k-}_\rho(\cdot)$ are the density constraint evaluated at particle $k$ after duplicating and removing the $i$-th particle respectively.
The new density constraints are evaluated as:
\begin{align}
    &C^{k+}_{\rho} (\mathbf{p}) = \frac{1}{\rho_0} \sum \limits_{j=1}^{N+1} W(|| \mathbf{p}^{k} - \mathbf{p}^j ||, h) - 1 \\
    \label{eq:simplified_duplicate_density_constraint}
    &C^{k+}_{\rho} (\mathbf{p}) = C^k_{\rho}(\mathbf{p}) + \frac{1}{\rho_0} W(||\mathbf{p}^{k} - 
    \mathbf{p}^{N+1}||, h)
\end{align}
for duplicating the $i$-th particle (so $\mathbf{p}^{N+1} = \mathbf{p}^i$) and
\begin{align}
    &C^{k-}_\rho(\mathbf{p}) = \frac{1}{\rho_0} \sum \limits_{j \neq i}^{N} W(|| \mathbf{p}^{k} - \mathbf{p}^j ||, h) - 1 \\
    \label{eq:simplified_removal_density_constraint}
    &C^{k-}_\rho(\mathbf{p}) = C^k_\rho(\mathbf{p}) - \frac{1}{\rho_0}W(|| \mathbf{p}^{k} - \mathbf{p}^i ||, h)
\end{align}
for removing the $i$-th particle where $C^k_\rho(\mathbf{p})$ is the density constraint evaluated at particle $k$ before duplicating or removing a particle.
Finally, (\ref{eq:optimization_selecting_particle}) is solved by explicitly computing the loss for every potential $i$ (i.e. computing loss after duplicating or removing every particle) and choosing $i$ that yields the smallest loss, hence duplicating or removing particles that best satisfy the density constraint.
Note that this can be efficiently computed due to the simplifications derived in (\ref{eq:simplified_duplicate_density_constraint}) and (\ref{eq:simplified_removal_density_constraint}).

\subsection{Liquid Prediction}

In Position Based Fluid simulations, the constraints at every time step update the positions of the particles which in turn induces a velocity for the particles \cite{macklin2013position}.
These constraint induced velocities combined with other external forces such as gravity are used to move the particles forward in time for liquid-like motion of the particles.
A similar approach is used here to recreate the liquid-like motion through time and hence enable reconstruction from a video of observations, $\mathbb{I}_1 \dots, \mathbb{I}_T$.
Let $\mathbf{p}^{i*}_{t}$ and $\mathbf{p}^{i*}_{t-1}$ be the optimized particles from solving (\ref{eq:optimization_problem}) at time $t$ and $t-1$ respectively.
Then the induced velocity for time $t$ is:
\begin{equation}
    \mathbf{v}^i_t = (1 - \lambda_{d})(\mathbf{p}^{i*}_{t} - \mathbf{p}^{i*}_{t-1})/ \Delta t 
\end{equation}
where $\lambda_{d} \in [0,1]$ is the velocity dampening factor and $\Delta t$ is the time-step size.
For consistent motion, XSPH viscosity \cite{schechter2012ghost} is applied:
\begin{equation}
    \bar{\mathbf{v}}^i_t = \mathbf{v}^i_t + \lambda_{v} \sum \limits_{j=1}^N \frac{\mathbf{v}^j_t - \mathbf{v}^i_t}{\rho^j(\mathbf{p})} W(||\mathbf{p}^{i*}_{t} - \mathbf{p}^{j*}_{t} ||, h)
\end{equation}
where $\lambda_{v}$ dictates the amount of viscosity applied.
The induced velocity is computed after every timestep of liquid reconstruction as shown in lines \ref{alg:compute_induced_velocity} and \ref{alg:dampen_and_viscocity} in Algorithm \ref{alg:main_outline}.
The induced velocity and gravity are used to forward predicts the particles to $t+1$ using equations of motion:
\begin{equation}
    \mathbf{p}^i_{t+1} = \mathbf{p}^{i*}_t + \bar{\mathbf{v}}^i_t \Delta t +  \frac{1}{2}\mathbf{g} \Delta t^2
\end{equation}
where $\mathbf{g}$ is the gravity vector.
This forward prediction is done in line \ref{alg:equation_of_motion} of Algorithm \ref{alg:main_outline}.
The dampening and viscosity not only represent physical properties, but also provide tuning parameters to stabilize the initialization for the next timestep.
Dampening, $\lambda_d$, dictates how much to rely on the prediction and viscosity, $\lambda_v$, adjusts the consistency of the velocity.

\section{Experiments}

To show the effectiveness of our proposed liquid reconstruction method, we test on a simulated and two real-life datasets with a comparative study.
These experiments are explained in the coming sections, and first implementation details are given.
Secondly, a description of our datasets and how they are collected is presented.
Lastly, we explain our comparative study and the results of it on the datasets.

\subsection{Implementation Details}

All the arithmetic, e.g. Newton's density constraint step in (\ref{eq:density_newton_step}), are implemented with PyTorch for its GPU integration \cite{paszke2017automatic}.
The collision constraint, (\ref{eq:collision_constraint}), and its solution, (\ref{eq:collision_constraint_solution}), are implemented with SPNet's \textit{ConvSP} operator and its PyTorch wrapper \cite{schenck2018spnets}, Kernel $K =\{[\pm1,0,0], [0,\pm1,0], [0,0,\pm1] \}$, and step size $d$ is equal to the resolution of the $SDF(\cdot)$.
The resting density $\rho_0$ is generated by setting a resting distance between particles because that is more intuitive to adjust.
The resting distance between particles is converted to the resting density by packing 1000 particles in a sphere and computing the particle density of the sphere.
Then the density constraint parameters to solve (\ref{eq:density_constraint}) and (\ref{eq:density_newton_step}) are set to a resting distance of $0.6h$, $\epsilon_p = 10^{2}$, and $W(\cdot)$ is set to Poly6 and Spiky Kernels for density estimation and gradient steps respectively \cite{muller2003particle}.
Differentiable rendering is done with the PyTorch3D framework \cite{ravi2020accelerating} and $\epsilon_s = 10^{-2}$.
The thresholds for adding/removing particles are $\gamma_s = 10^{-3}$ and $\gamma_I=0.9$ respectively.
Velocity dampening and viscosity coefficients are set to $\lambda_d = 0.2$ and $\lambda_v = 0.75$ respectively.
The parameters in Algorithm (\ref{alg:main_outline}) are set to $n_o = 30$, $n_j = 2$, $n_c=5$, $n_i = 5$, and $\alpha_\mathbb{I} = 0.02$.
All datasets are stereo so an initial four particles can be placed at a stereo-computed, 3D location from the first liquid detections. 
The last parameters, $SDF(\cdot)$ resolution and particle interaction radius, $h$, are set depending on the dataset as they need to be adjusted depending on the scale of the reconstruction.

\subsection{Datasets}

\begin{figure}[t]
    \centering
    \includegraphics[trim=0cm 0cm 0cm 0cm, clip, width=0.48\linewidth]{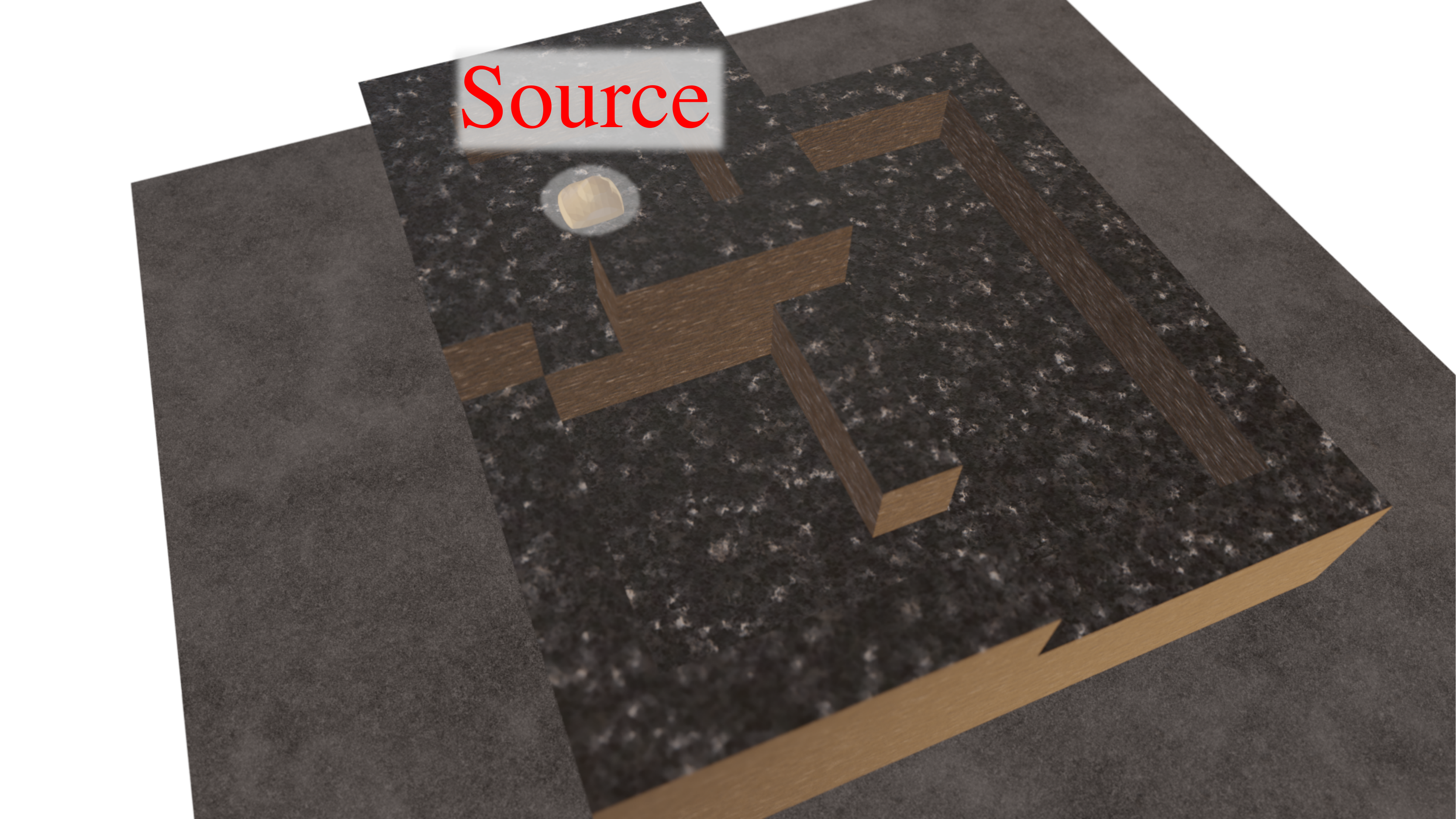}
    \includegraphics[trim=0cm 0cm 0cm 0cm, width=0.48\linewidth]{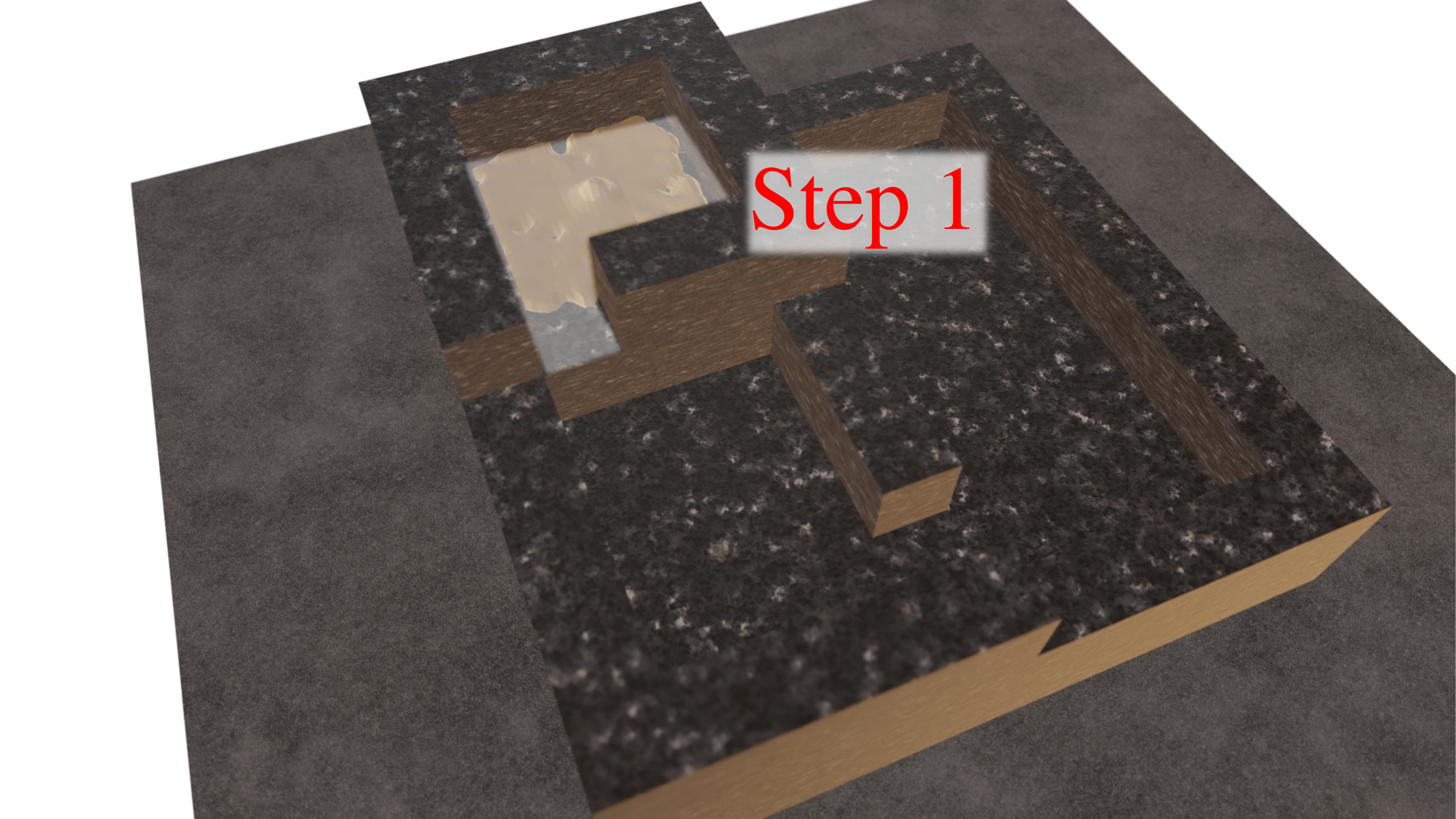}
    
    \vspace{0.5mm}
    
    \includegraphics[trim=0cm 0cm 0cm 0cm, width=0.48\linewidth]{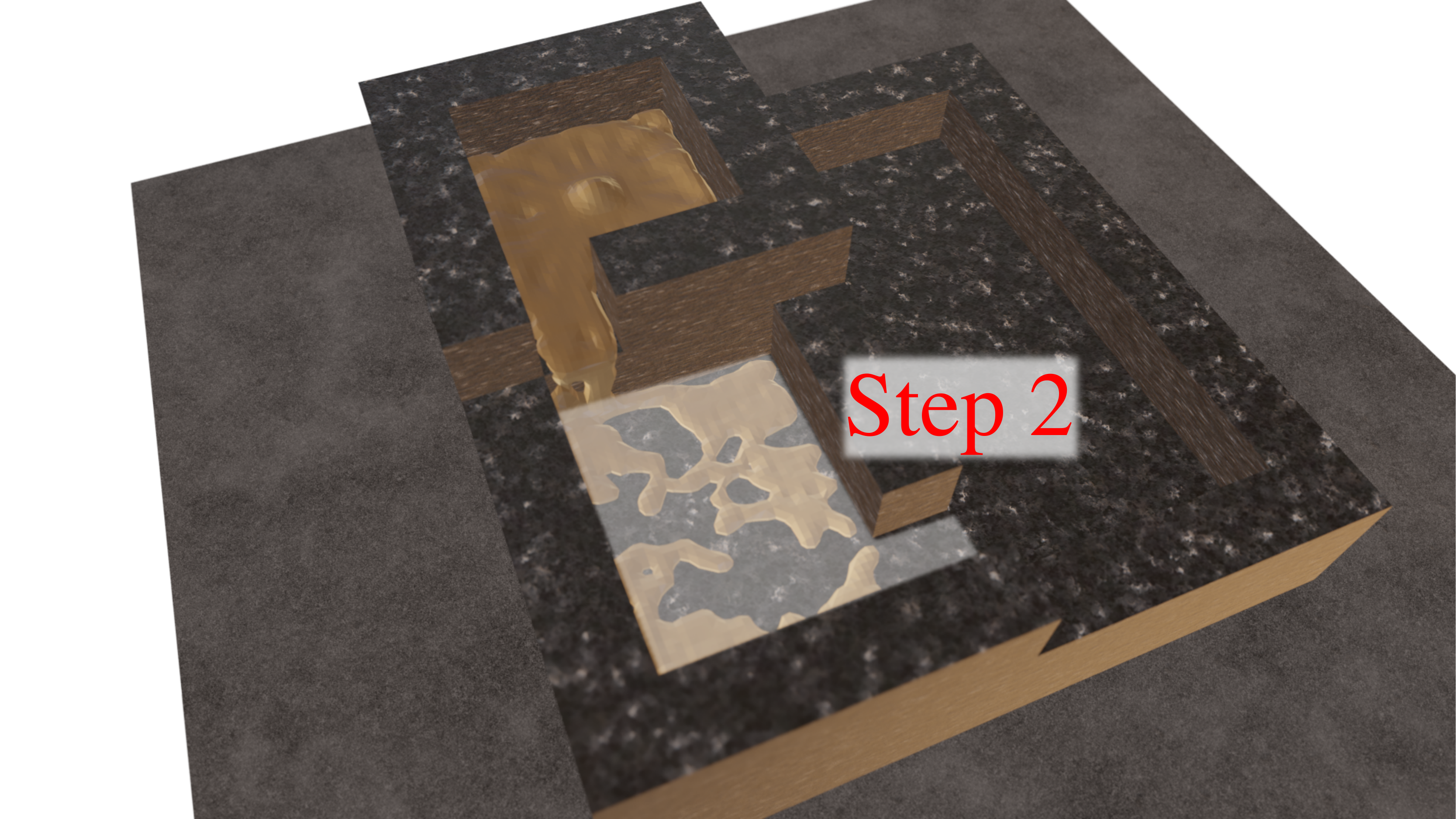}
    \includegraphics[trim=0cm 0cm 0cm 0cm, width=0.48\linewidth]{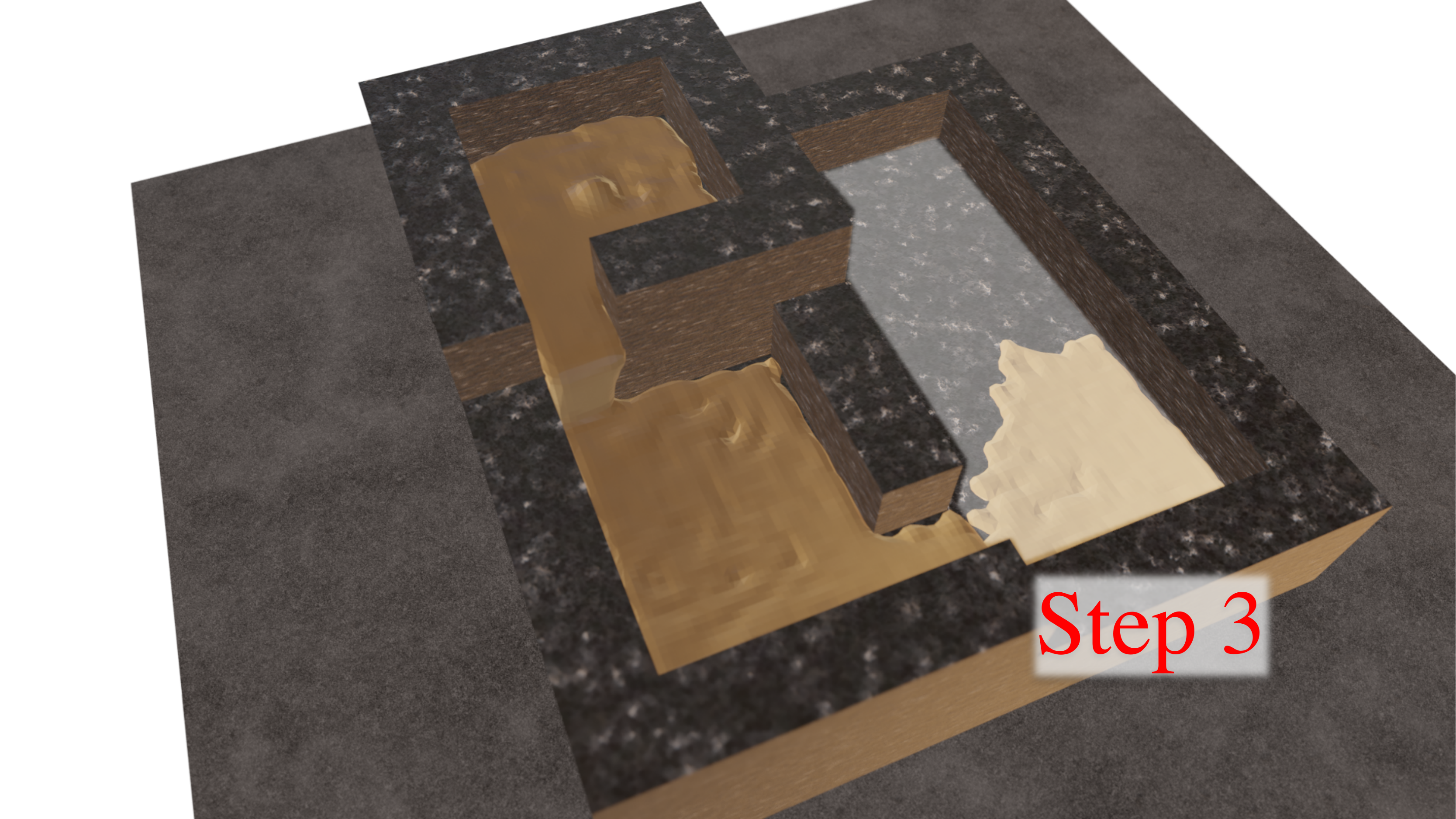}
    \caption{
    The sequence of figures (top-left, top-right, bottom-left, bottom-right) shows how liquid fills the simulated fountain.
    Take note how the first step fills in a consistent shape, but significant turbulence occurs when dropping to the second step making this a challenging component of the scene.
    Another challenge is by the time the third and final step of the fountain fills, a significant number of particles must be used for reconstruction due to the large volume, hence testing the scalability of the reconstruction method.
    }
    \label{fig:simulated_set_up}
\end{figure}

\textit{Simulated Fountain}: The first dataset is generated on a three step fountain, shown in Fig. \ref{fig:simulated_set_up}, with Blender \cite{blender}.
The liquid simulation uses all default values except the viscosity is set to 0.001.
The $SDF(\cdot)$ is generated from the fountain with a resolution of 1cm and the particle interaction radius, $h$, is set to 1cm.
The scene is rendered with 1080p at 24fps stereo cameras, and a mask of the rendered liquid is directly outputted from Blender.
For this simulated dataset, the ground-truth liquid mesh is available to evaluate our recontruction with.
The metric of 3D IoU is used to capture the shape accuracy of our reconstruction and computed as:
\begin{equation}
    \text{IoU}_{3D} = \frac{\mathbb{V} \cup \hat{\mathbb{V}}(\mathbf{p}) }{\mathbb{V} \cap \hat{\mathbb{V}}(\mathbf{p})}
\end{equation}
where $\mathbb{V}$ and $\hat{\mathbb{V}}(\mathbf{p})$ are voxel representation of the simulated and reconstructed liquid respectively.
The reconstructed liquid in voxel representation,  $\hat{\mathbb{V}}(\mathbf{p})$, is generated with the color field, shown in equation (\ref{eq:color_field}) in the supplementary material.
The voxel grid is computed at a resolution of 3cm.

\begin{figure}[t]
    \centering
    \includegraphics[trim=0cm 6.9cm 15.8cm 0cm, clip, width=0.52\linewidth]{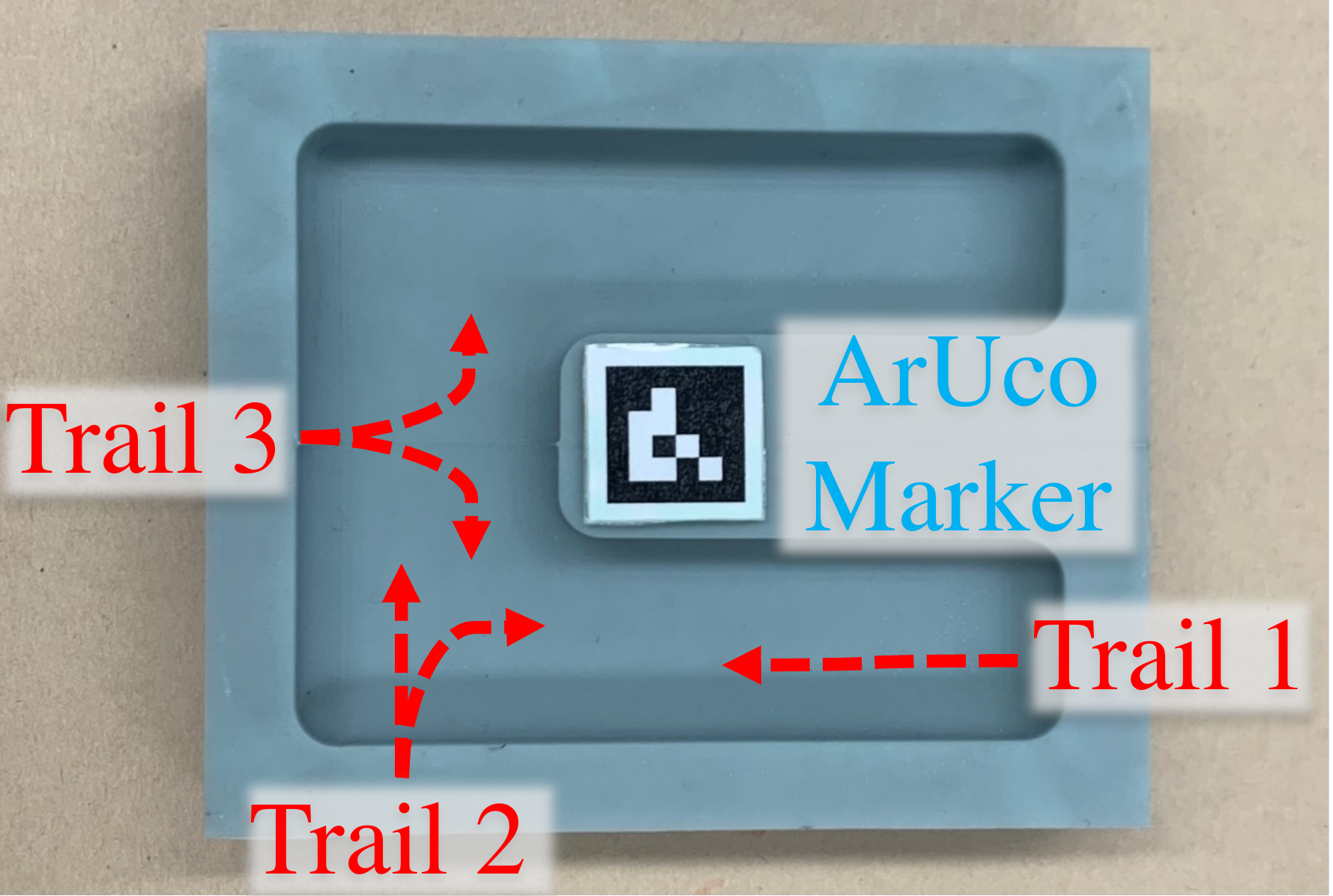}
    \includegraphics[trim=0cm 6.9cm 17.8cm 0cm, clip, width=0.46\linewidth]{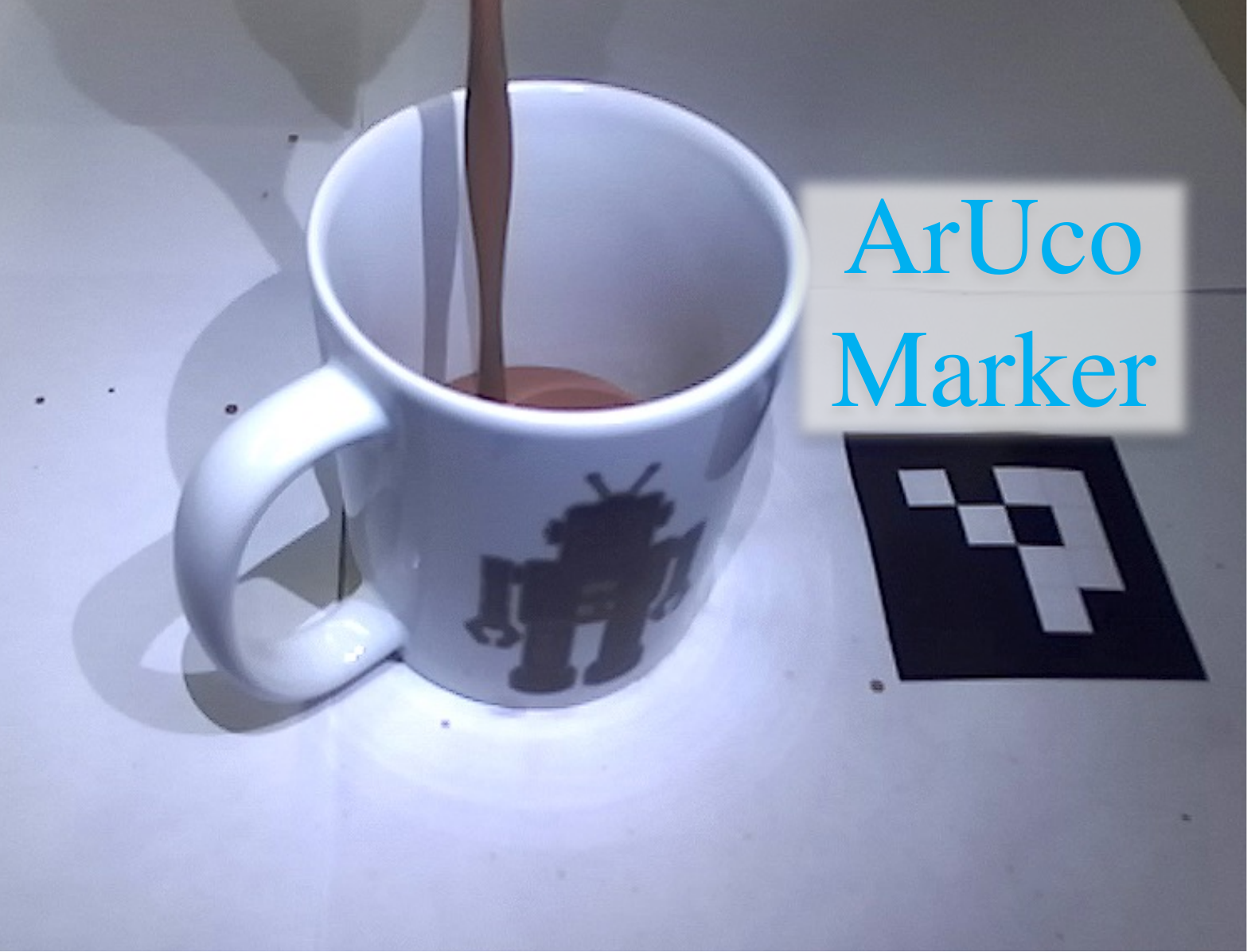}
    \caption{
    The left figure shows a top down view of the silicon cavity used for the Endoscopic Liquid dataset, and the liquid is injected with a syringe at the labelled points for three trials.
    The right figure shows a camera image from our Pouring Milk experiment set up. Notice that the milk is partially blocked by the mug, hence testing the reconstructions ability to handle occlusions.
    }
    \label{fig:experiment_set_up}
\end{figure}

\begin{figure*}
    \centering
    \includegraphics[trim=2.4cm 7.5cm 3cm 0.5cm, clip, width=0.7\linewidth]{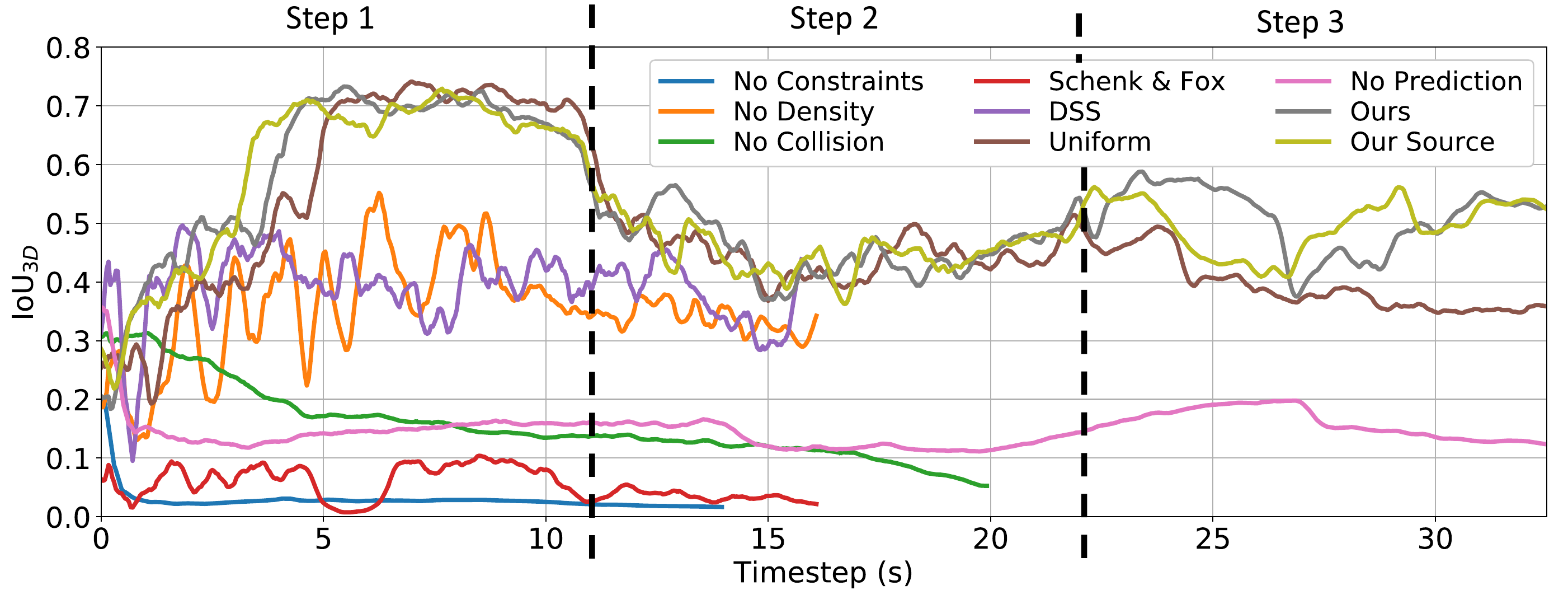}
    \hspace{5pt}
    \includegraphics[trim=0cm 0cm 13cm 0cm, clip, width=0.28\linewidth]{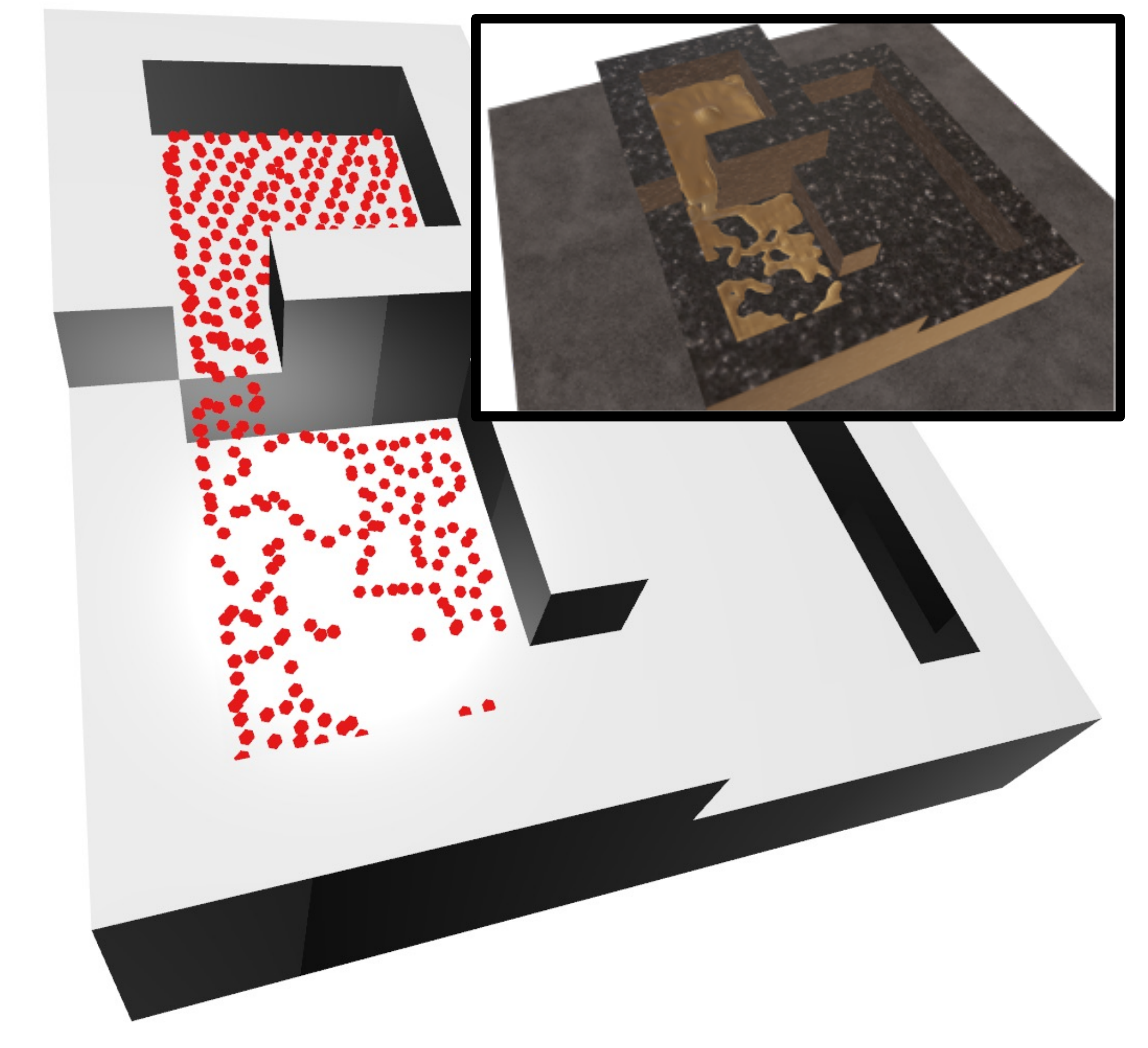}
    \caption{The plot on the left shows IoU$_{3D}$ results from the Simulated Fountain datasets along with time-marked points when the liquid reaches different steps in the scene.
    Note how our proposed methods and the uniform comparison are able to reach 70\% IoU$_{3D}$ in the first step, and retain a good reconstruction as the very long, and turbulent simulation continues.
    An example of our reconstruction approach during the turbulent period of the scene is shown on the right-hand figure.
    Meanwhile the compared approaches ran into memory limitations and crashed (required greater than 24GB of memory) or were unable to converge effectively.
    }
    \label{fig:simulated_results}
\end{figure*}

\begin{figure*}
    \centering
    %\vspace{2mm}
    
    \begin{subfigure}{0.155\linewidth}
        \setlength{\fboxsep}{0pt}
        \setlength{\fboxrule}{1.5pt}
        \fbox{\includegraphics[trim=0cm 4.5cm 0cm 0cm, clip, width=\linewidth]{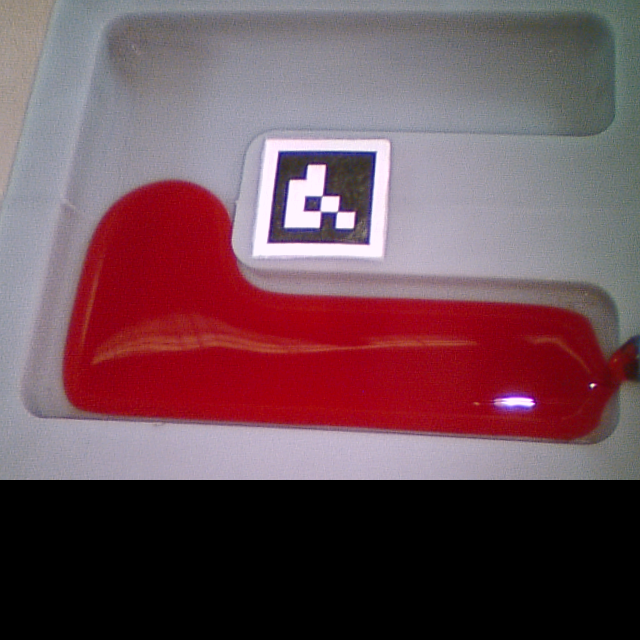}}
    \end{subfigure}%
    \hspace{2pt}
    \begin{subfigure}{0.155\linewidth}
        \centering
        \begin{subfigure}{1.0\textwidth}
        \includegraphics[trim=12cm 10cm 24cm 11cm, clip, width=\linewidth]{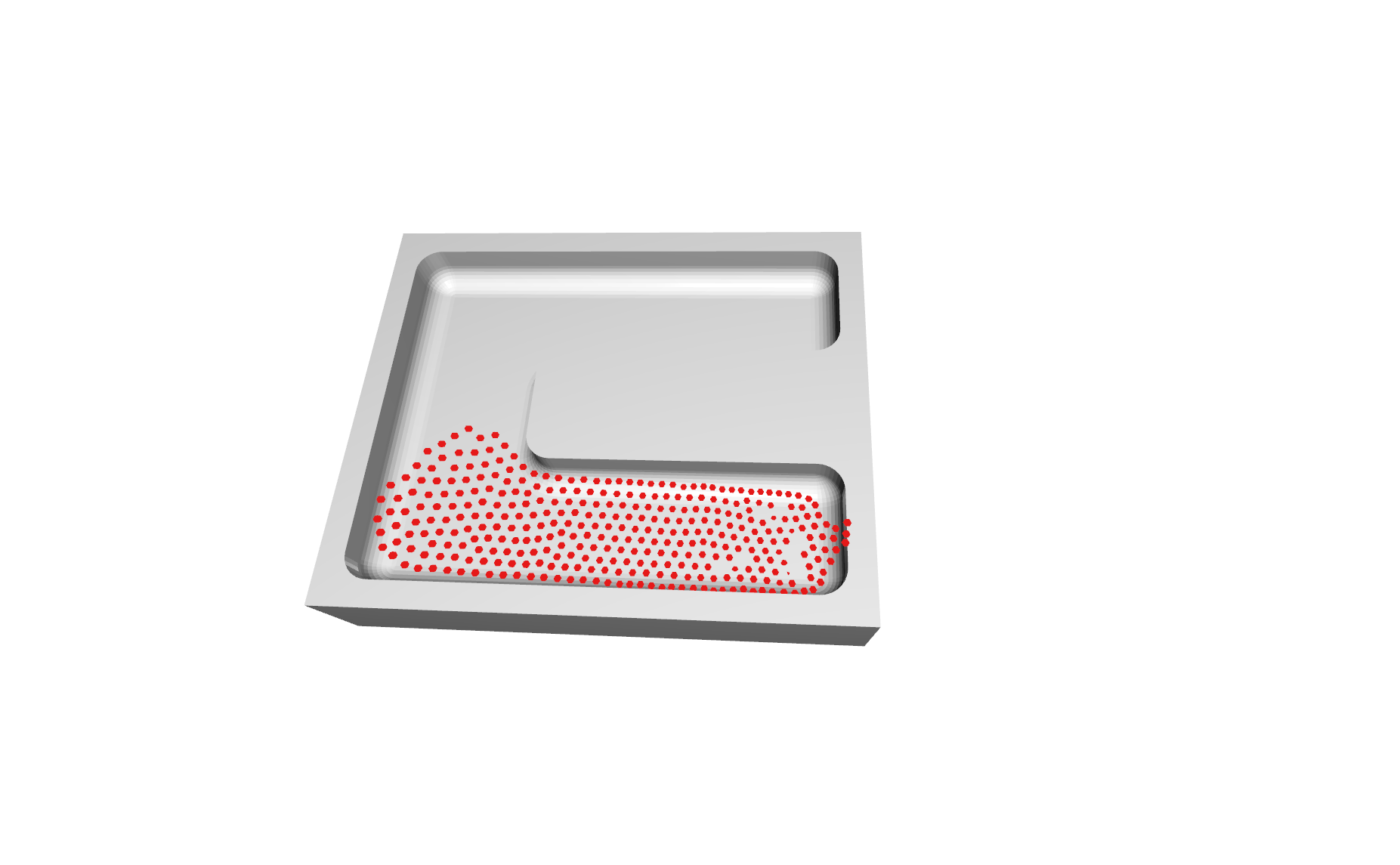}  
        \end{subfigure}
        
        \begin{subfigure}{1.0\textwidth}
        \includegraphics[trim=12cm 7cm 17cm 4cm, clip, width=\linewidth]{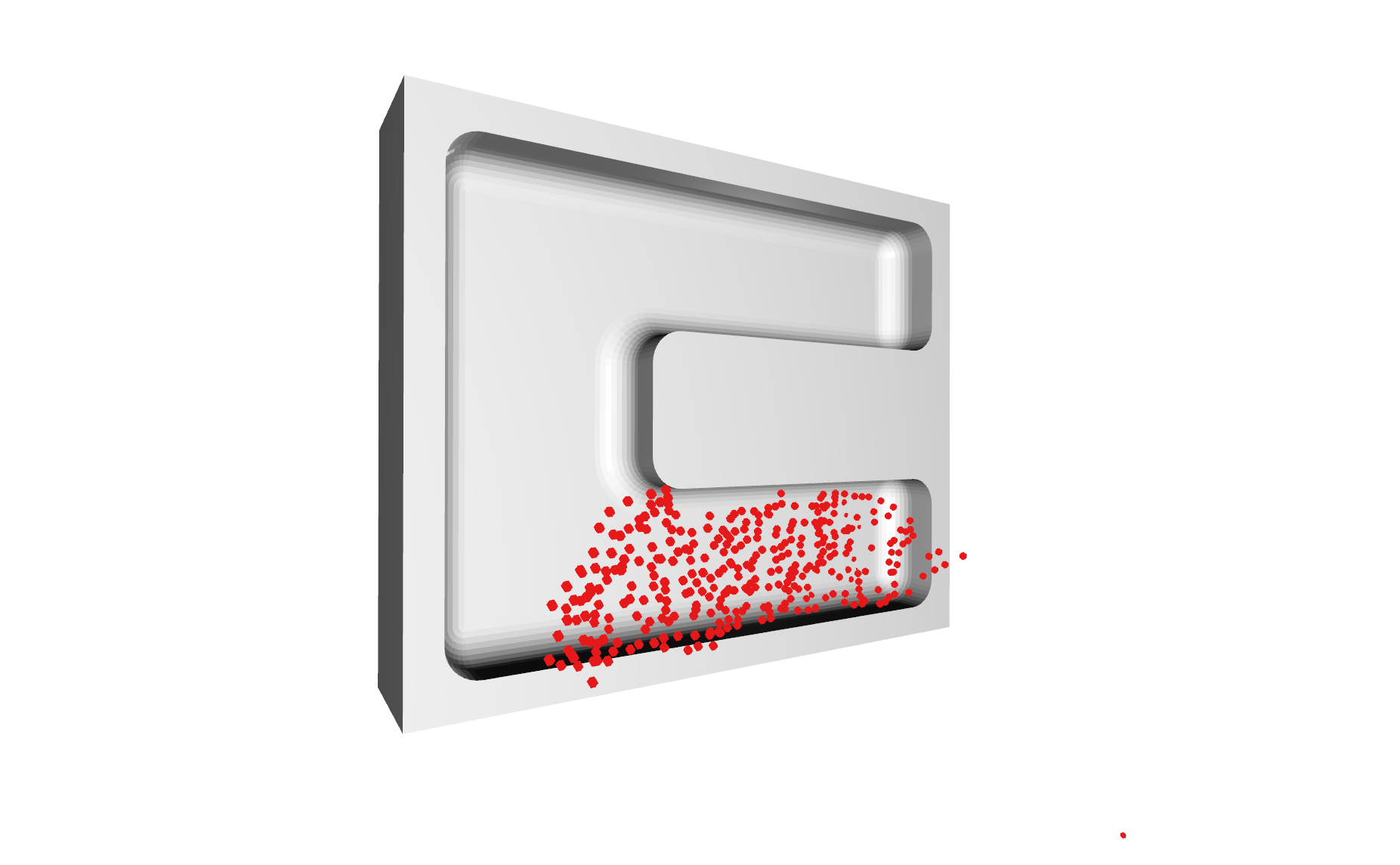}  
        \end{subfigure}
    \end{subfigure}%
    \hspace{2pt}
    \begin{subfigure}{0.155\linewidth}
        \centering
        \begin{subfigure}{1.0\textwidth}
        \includegraphics[trim=12cm 10cm 24cm 11cm, clip,width=\linewidth]{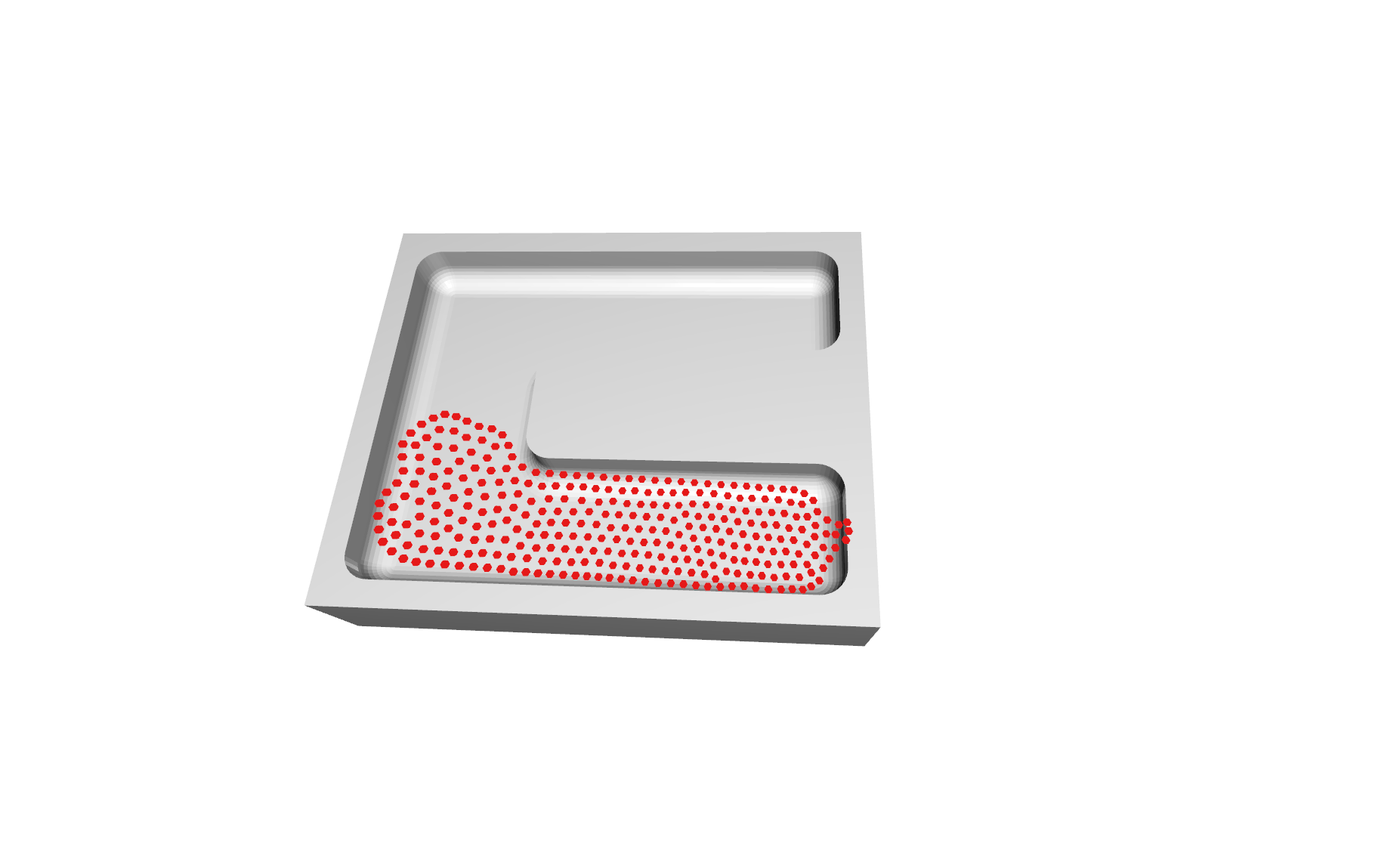}  
        \end{subfigure}
        
        \begin{subfigure}{1.0\textwidth}
        \includegraphics[trim=12cm 7cm 17cm 4cm, clip,width=\linewidth]{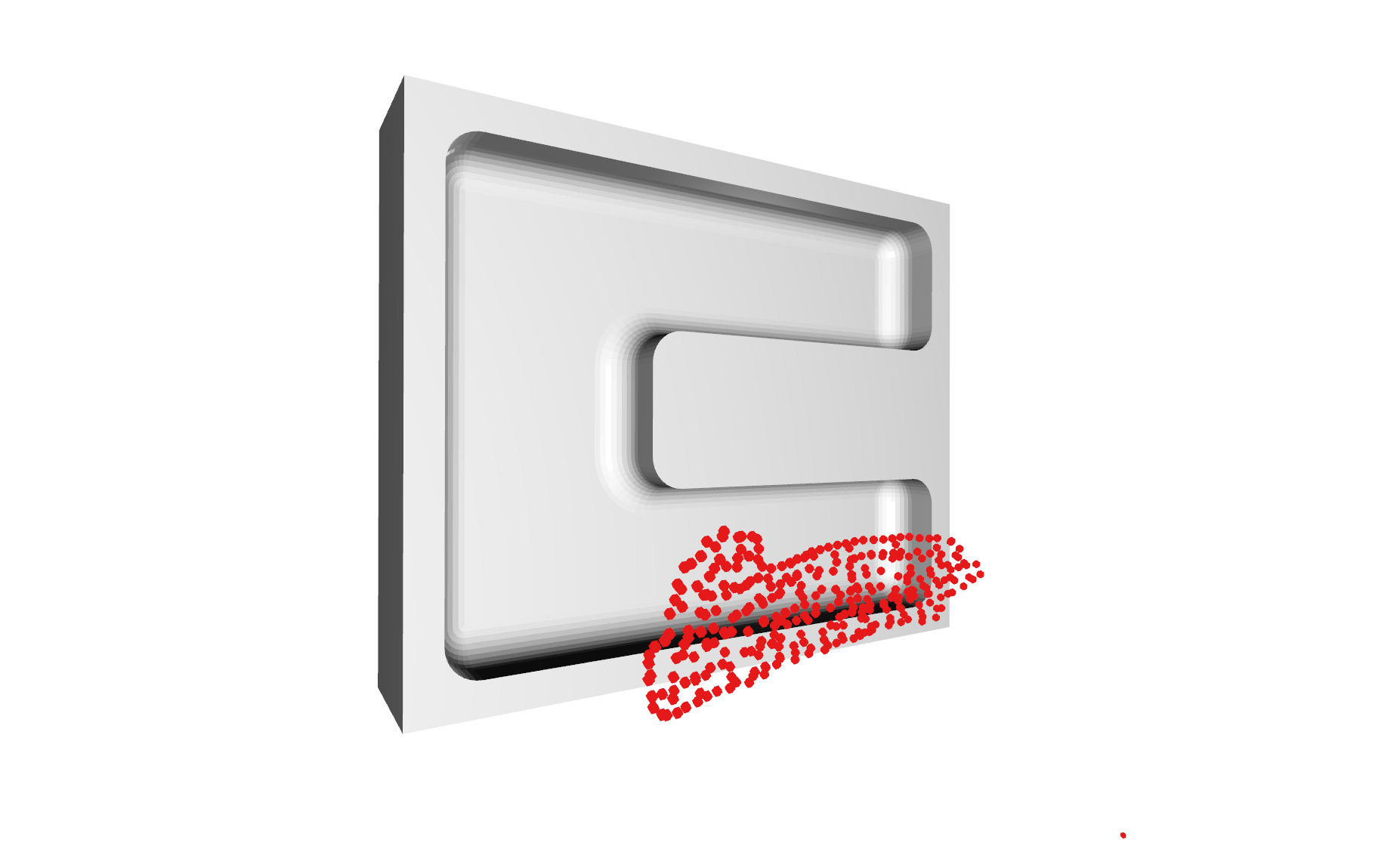}  
        \end{subfigure}
    \end{subfigure}%
    \hspace{2pt}
    \begin{subfigure}{0.155\linewidth}
        \centering
        \begin{subfigure}{1.0\textwidth}
        \includegraphics[trim=12cm 10cm 24cm 11cm, clip,width=\linewidth]{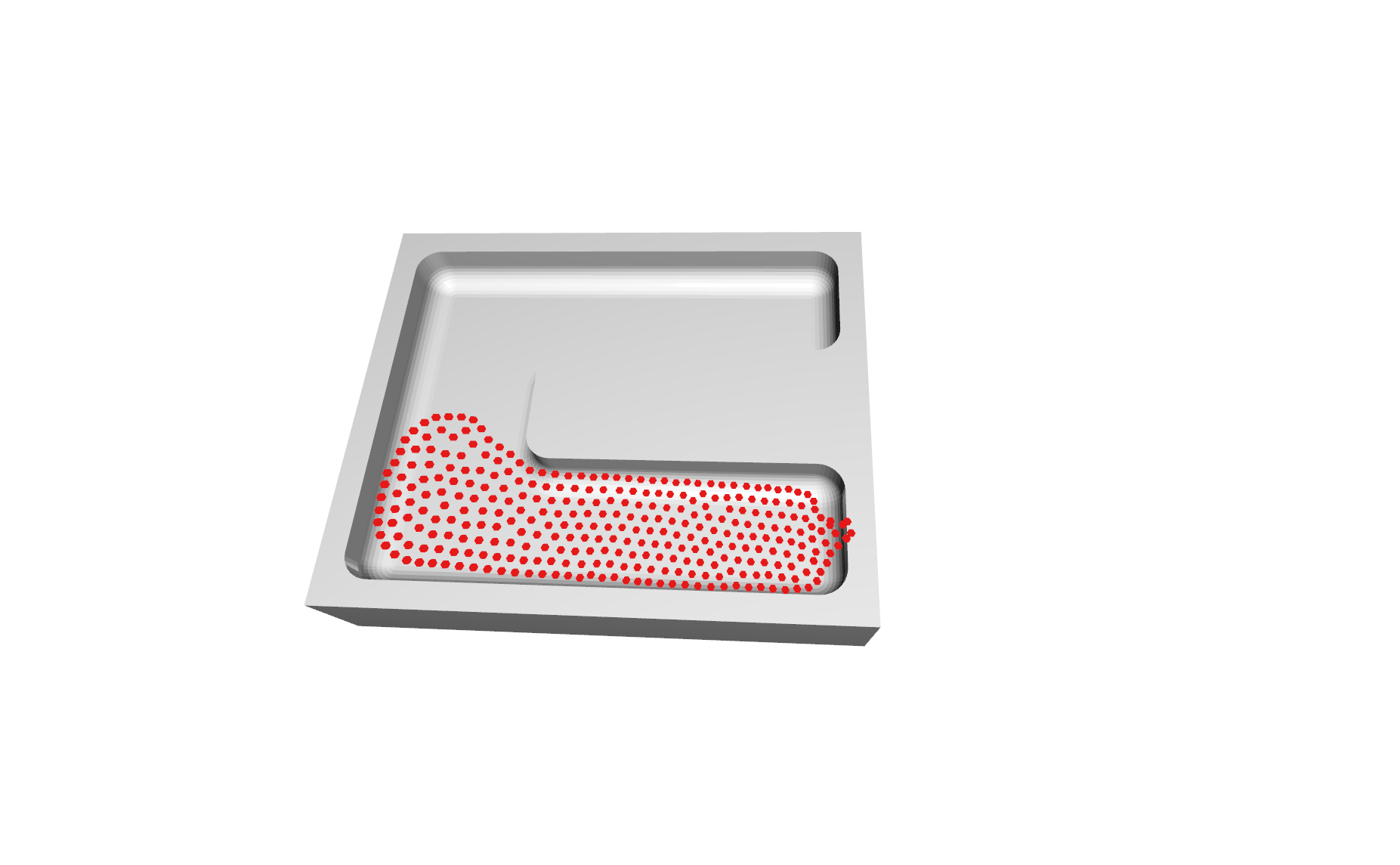}  
        \end{subfigure}
        
        \begin{subfigure}{1.0\textwidth}
        \includegraphics[trim=12cm 7cm 17cm 4cm, clip,width=\linewidth]{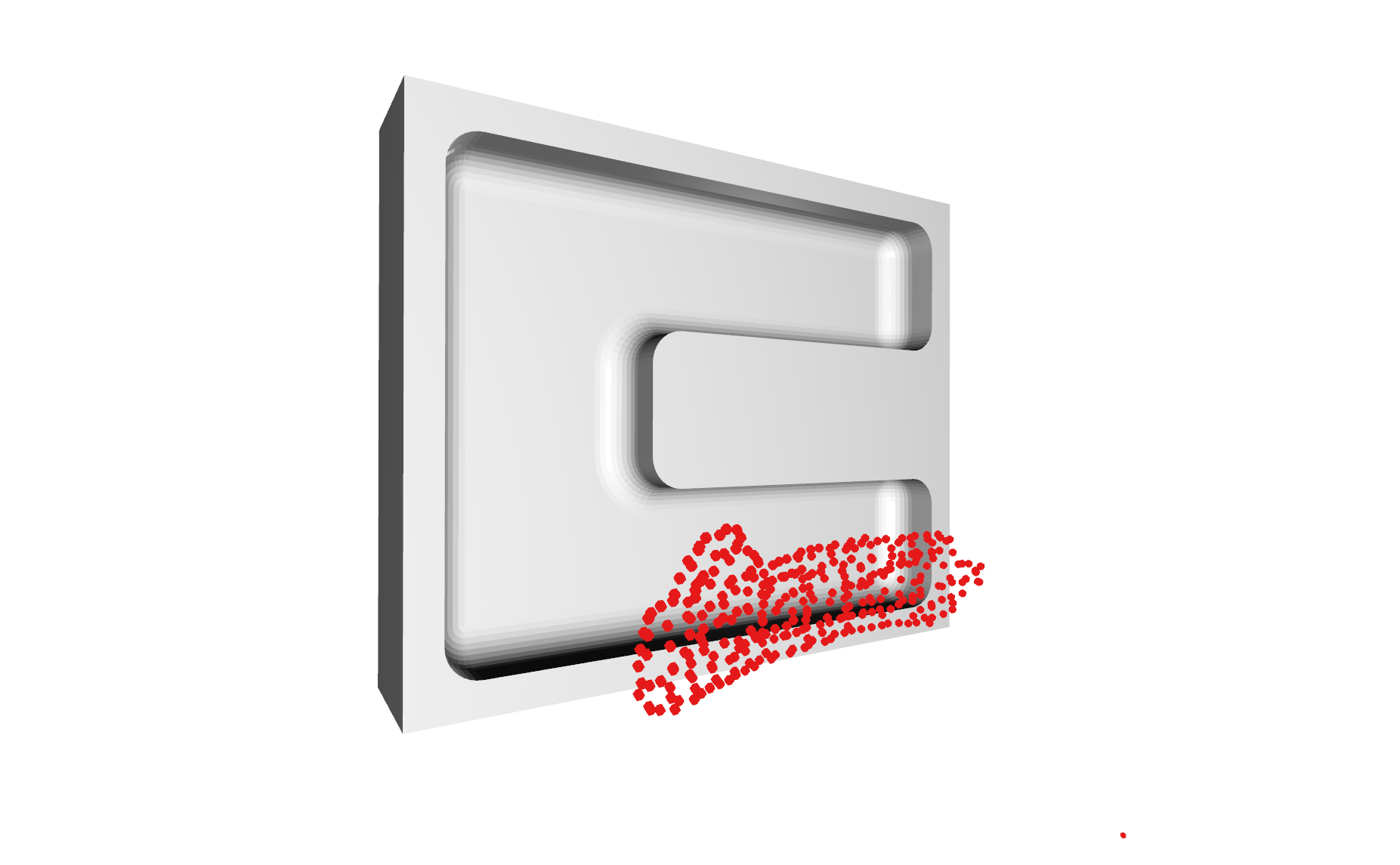}  
        \end{subfigure}
    \end{subfigure}%
    \hspace{2pt}
    \begin{subfigure}{0.155\linewidth}
        \centering
        \begin{subfigure}{1.0\textwidth}
        \includegraphics[trim=12cm 10cm 24cm 11cm, clip,width=\linewidth]{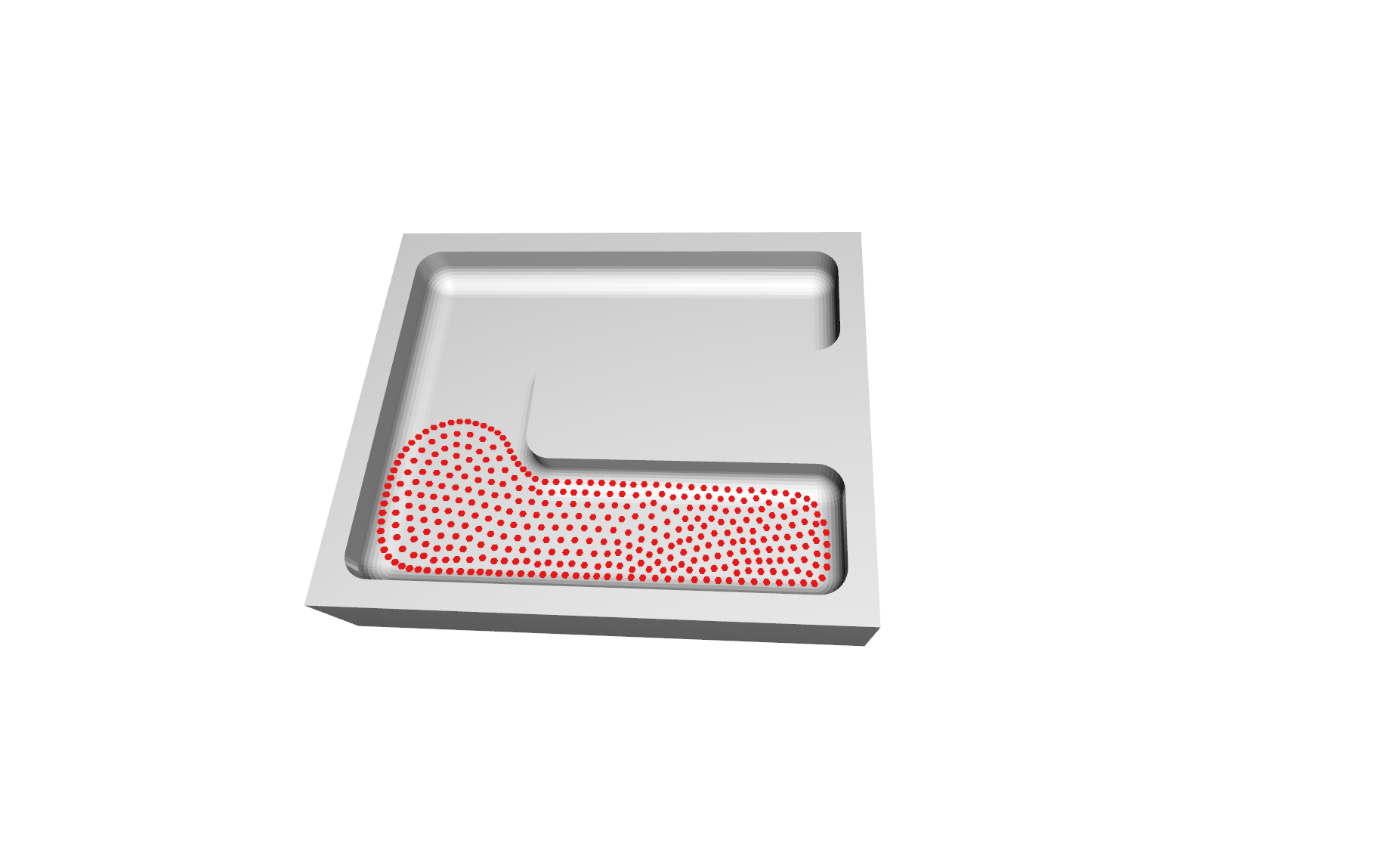}  
        \end{subfigure}
        
        \begin{subfigure}{1.0\textwidth}
        \includegraphics[trim=12cm 7cm 17cm 4cm, clip,width=\linewidth]{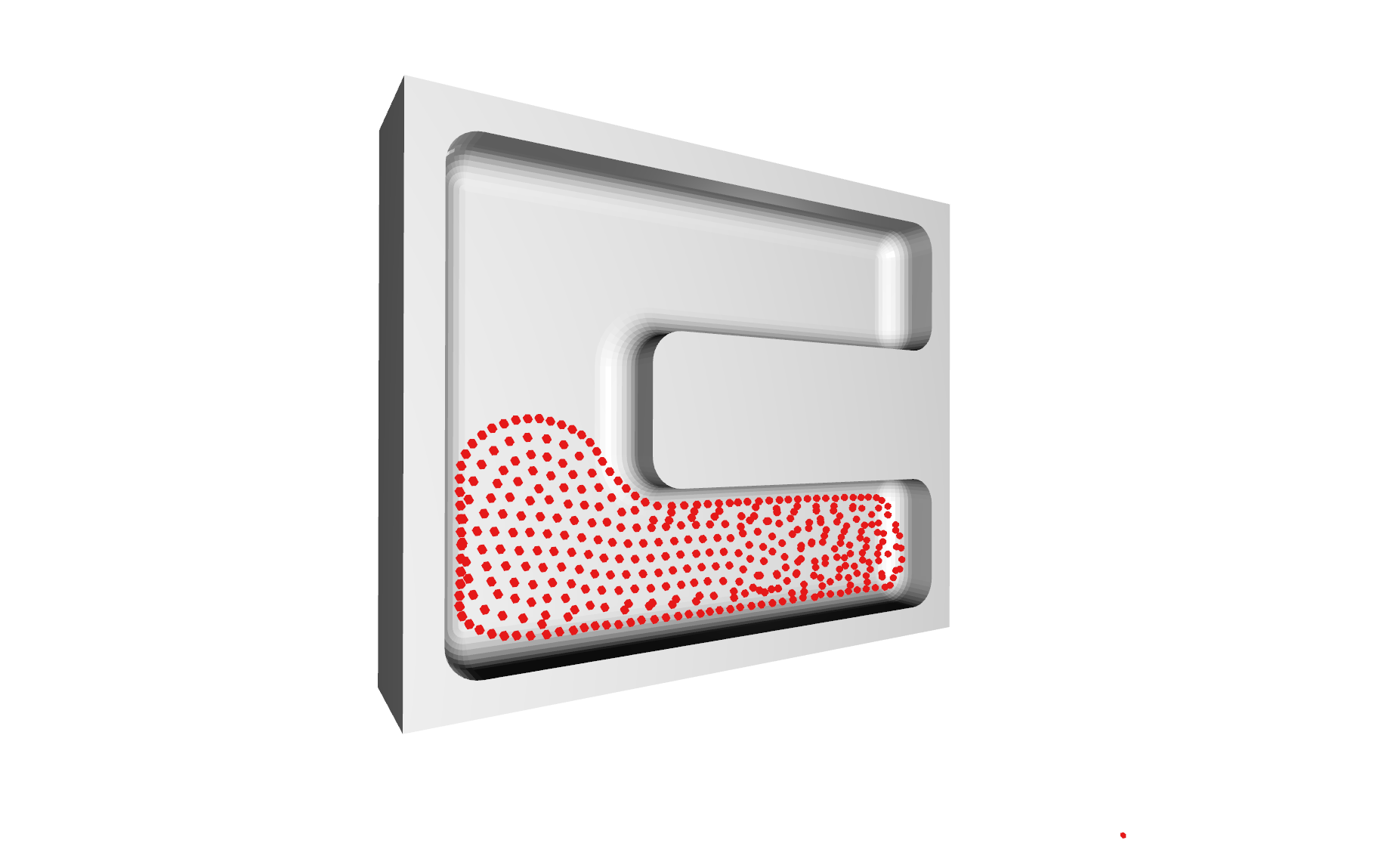}  
        \end{subfigure}
    \end{subfigure}%
    \hspace{1pt}
    \begin{subfigure}{0.155\linewidth}
        \centering
        \begin{subfigure}{1.0\textwidth}
        \includegraphics[trim=12cm 10cm 24cm 11cm, clip,width=\linewidth]{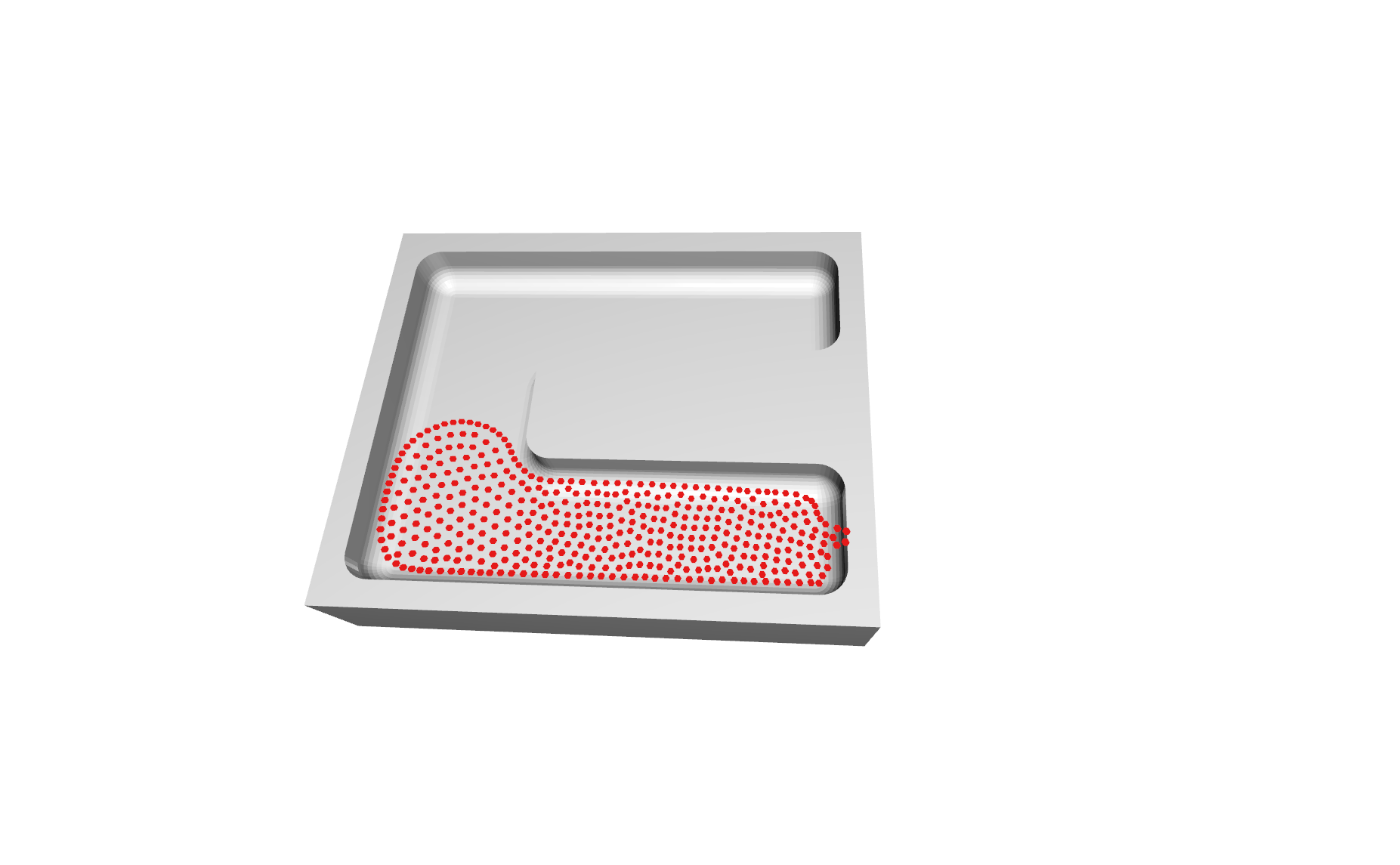}  
        \end{subfigure}
        
        \begin{subfigure}{1.0\textwidth}
        \includegraphics[trim=12cm 7cm 17cm 4cm, clip,width=\linewidth]{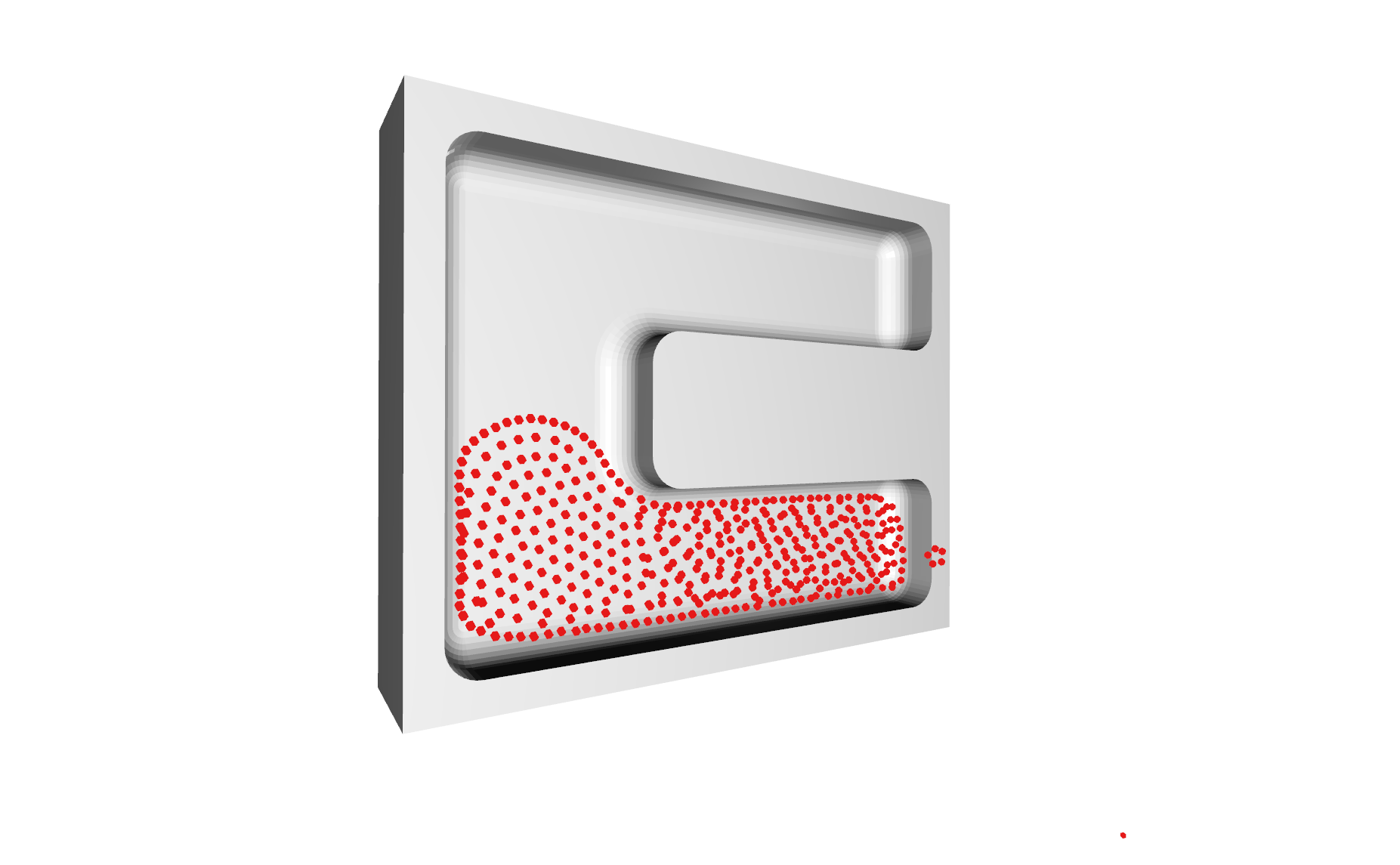}  
        \end{subfigure}
    \end{subfigure}%
    
    \caption{From left to right the image columns are an endoscopic image of liquid being reconstruction with: no constraints,  no collision constraint, no prediction, our approach, and our source estimation technique. The first row of renderings have the virtual camera positioned similar to the real endoscope showing how from that perspective, the particles in red line up with the real image of the liquid. The second row shows another rendered perspective and how our proposed approaches properly reconstruct the liquid in 3D. The first three comparisons are unable to properly reconstruct because they do not leverage a liquids dynamics (i.e. falling to gravity and colliding with the cavity).}
    \label{fig:cavity_qualitative_results}
\end{figure*}

\textit{Endoscopic Liquid}: The second dataset uses a custom silicon cavity that was molded with a 3D printed negative so a $SDF(\cdot)$ for it can be generated.
The cavity is 11cm by 9.5cm, $SDF(\cdot)$ resolution is set to 1mm, and particle interaction radius, $h$, is set to 5mm.
To transform the $SDF(\cdot)$ to the camera frame, which is the coordinate frame the particles are being optimized in, an ArUco Marker \cite{garrido2014automatic} is placed on the cavity in a known location.
Roughly 50ml of water is injected with a syringe at three different locations for three trails as depicted in Fig. \ref{fig:experiment_set_up}.
The water is mixed with red-coloring dye so color segmentation can be applied to detect the liquid surface.
The liquid video is recorded using a da Vinci Research Kit stereo-endoscope which is 1080p at 30fps \cite{kazanzides2014open}.

\textit{Pouring Milk}:
The third dataset is pouring chocolate milk by a human into a mug as shown in Fig. \ref{fig:experiment_set_up}.
The mug is 9cm high and has a 7cm diameter, the $SDF(\cdot)$ resolution is set to 1mm, and particle interaction radius, $h$, is set to 6.5mm.
The mug is placed on a sheet of paper with an ArUco Marker \cite{garrido2014automatic} in a marked location.
The Aruco Marker and known geometry of the paper provides the transformation to take the $SDF(\cdot)$ to the camera frame.
Color segmentation is used to detect the chocolate milk's liquid surface.
The liquid video is recorded at 720p 15fps using a ZED Stereo Camera from Stereo Labs.

\subsection{Comparative Study}

We show the effectiveness of our proposed method through a comparative study.
The configurations being compared are:
\begin{itemize}
    \setlength\itemsep{0.1em}
    \item \textit{No Constraints} \cite{lassner2021pulsar} (i.e. no density or collision constraints) and only image loss
    \item \textit{No Density} constraint
    \item \textit{No Collision} constraint
    \item \textit{Schenck \& Fox} \cite{schenck2018spnets} constraints instead of the density constraint we presented
    \item \textit{DSS} \cite{yifan2019differentiable} for rendering gradients rather than Pulsar
    \item \textit{Uniform} random selection for duplication or removal of particles instead of solving (\ref{eq:optimization_selecting_particle})
    \item \textit{No Prediction} of particles (line \ref{alg:equation_of_motion} in Algorithm \ref{alg:main_outline})
    \item \textit{Our} complete approach
    \item \textit{Our Source} estimation which adds particles at a source location and detailed in the next sub-section
\end{itemize}
\begin{figure*}
    \centering
    %\vspace{2mm}
    
    \begin{subfigure}{0.155\linewidth}
        \setlength{\fboxsep}{0pt}
        \setlength{\fboxrule}{1.5pt}
        \fbox{\includegraphics[trim=10cm 25cm 10cm 1cm, clip, width=\linewidth]{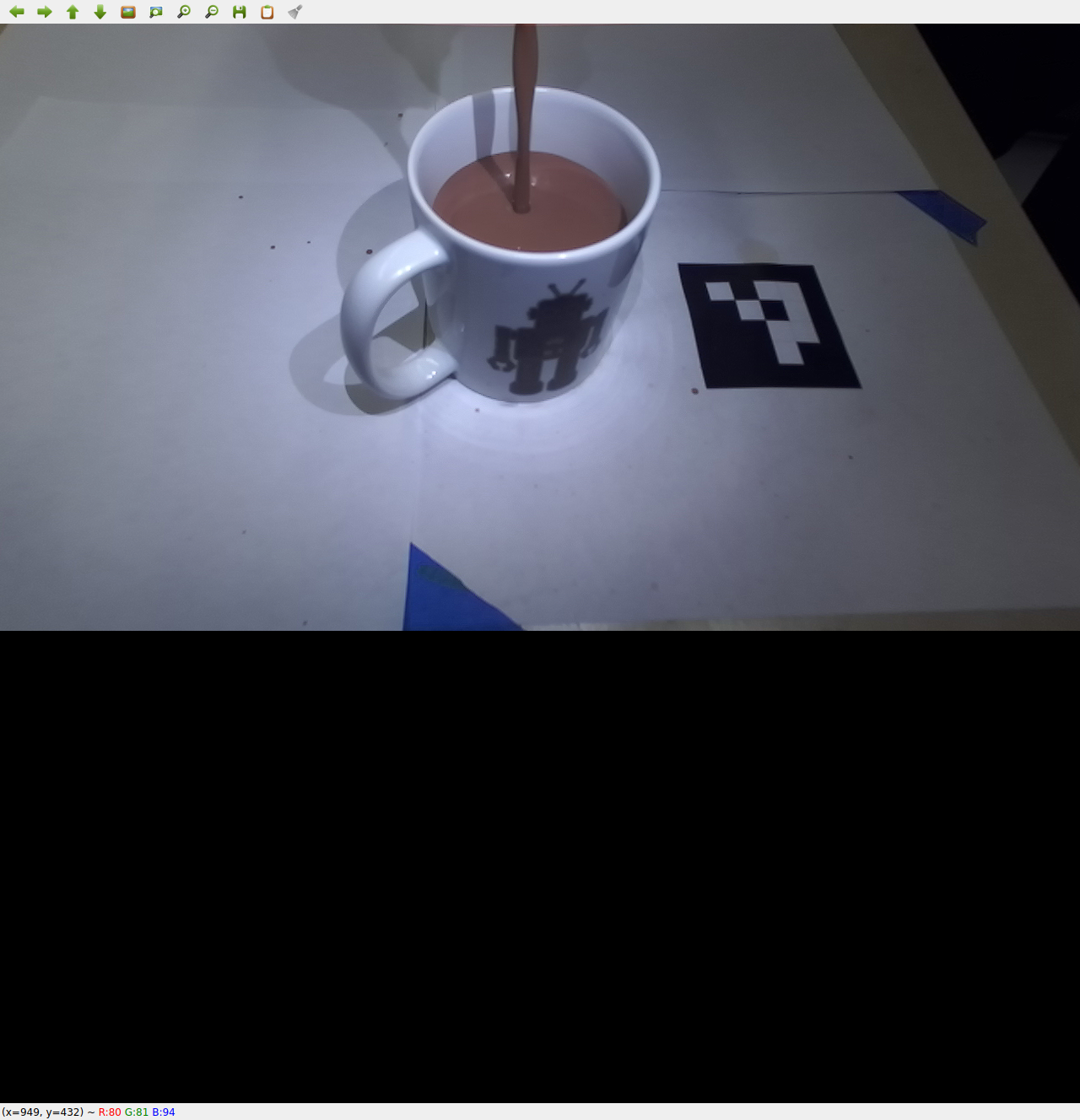}}
    \end{subfigure}%
    \hspace{2pt}
    \begin{subfigure}{0.155\linewidth}
        \centering
        \begin{subfigure}{1.0\textwidth}
        \includegraphics[trim=13cm 5cm 19cm 0cm, clip, width=\linewidth]{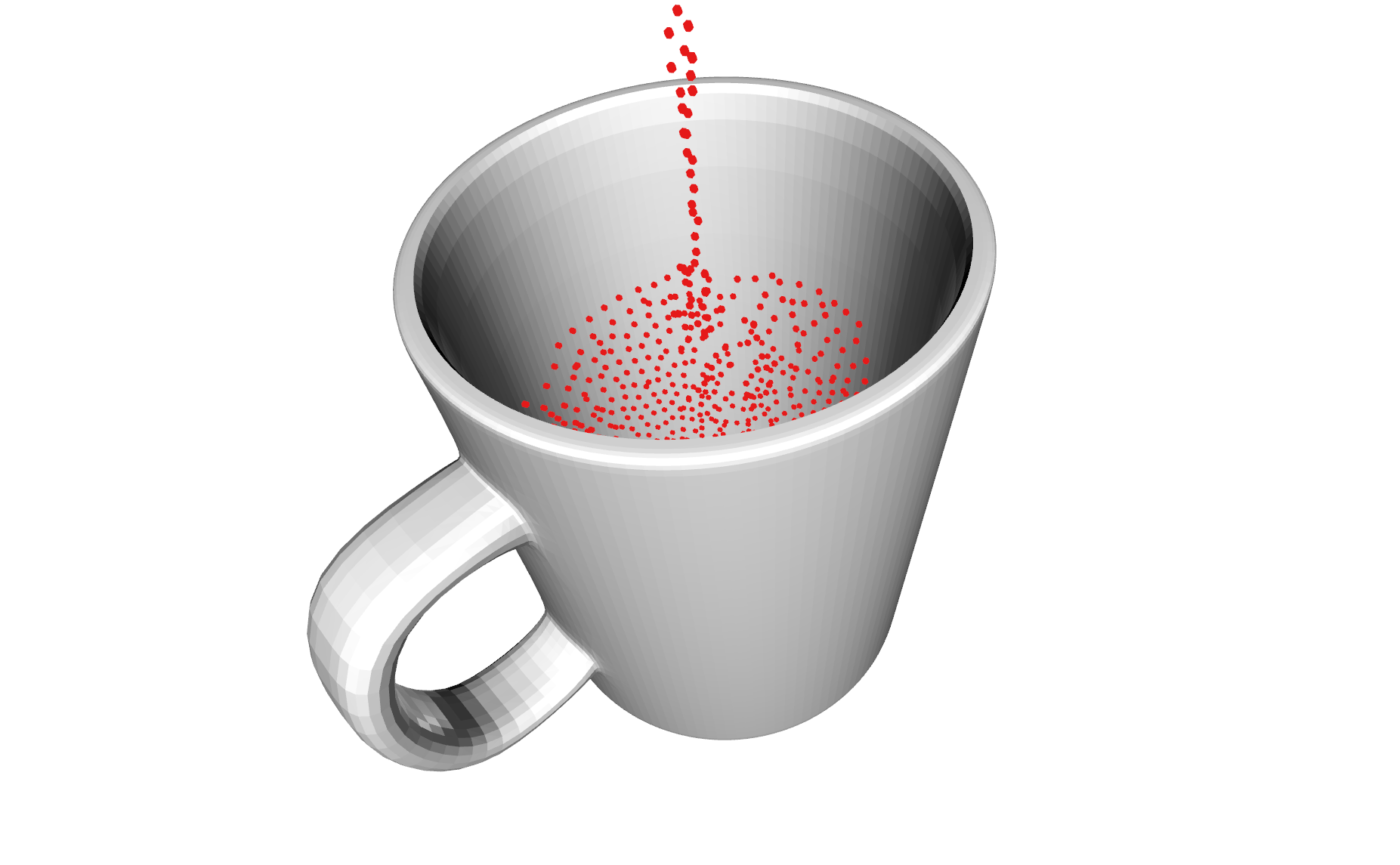}  
        \end{subfigure}
        
        \begin{subfigure}{1.0\textwidth}
        \includegraphics[trim=18cm 6cm 16cm 2cm, clip, width=\linewidth]{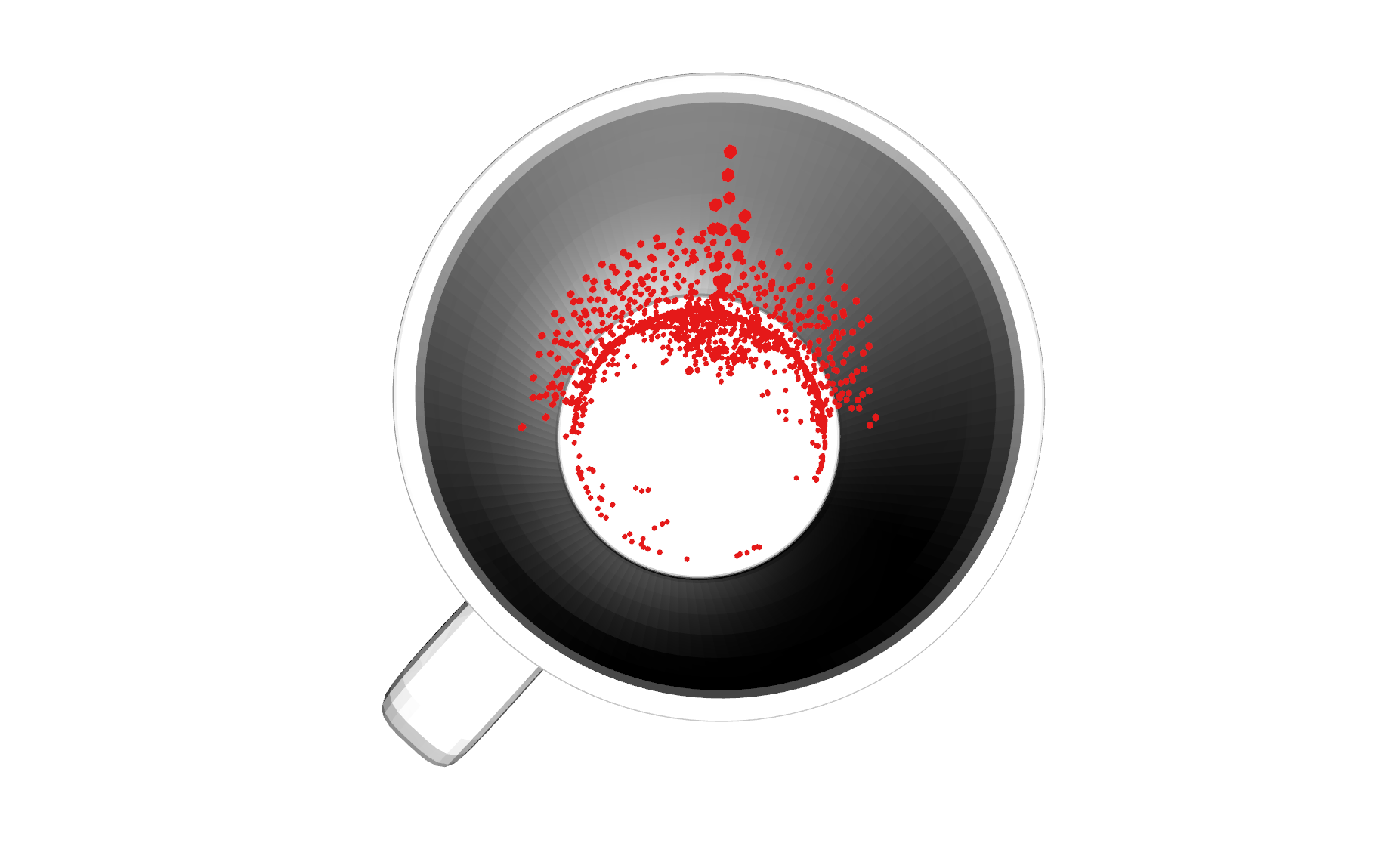}  
        \end{subfigure}
    \end{subfigure}%
    \hspace{2pt}
    \begin{subfigure}{0.155\linewidth}
        \centering
        \begin{subfigure}{1.0\textwidth}
        \includegraphics[trim=13cm 5cm 19cm 0cm, clip,width=\linewidth]{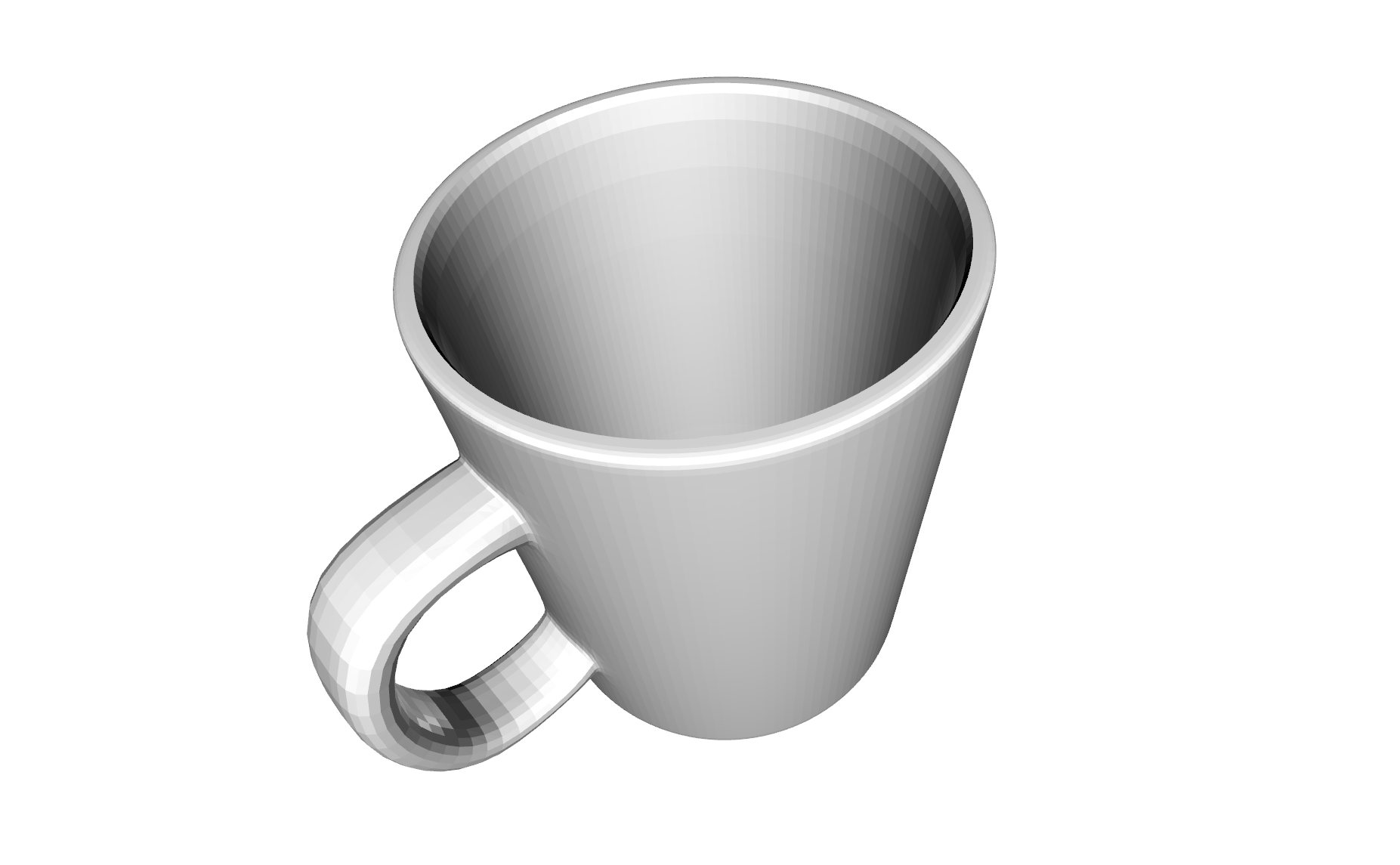}  
        \end{subfigure}
        
        \begin{subfigure}{1.0\textwidth}
        \includegraphics[trim=18cm 6cm 16cm 2cm, clip,width=\linewidth]{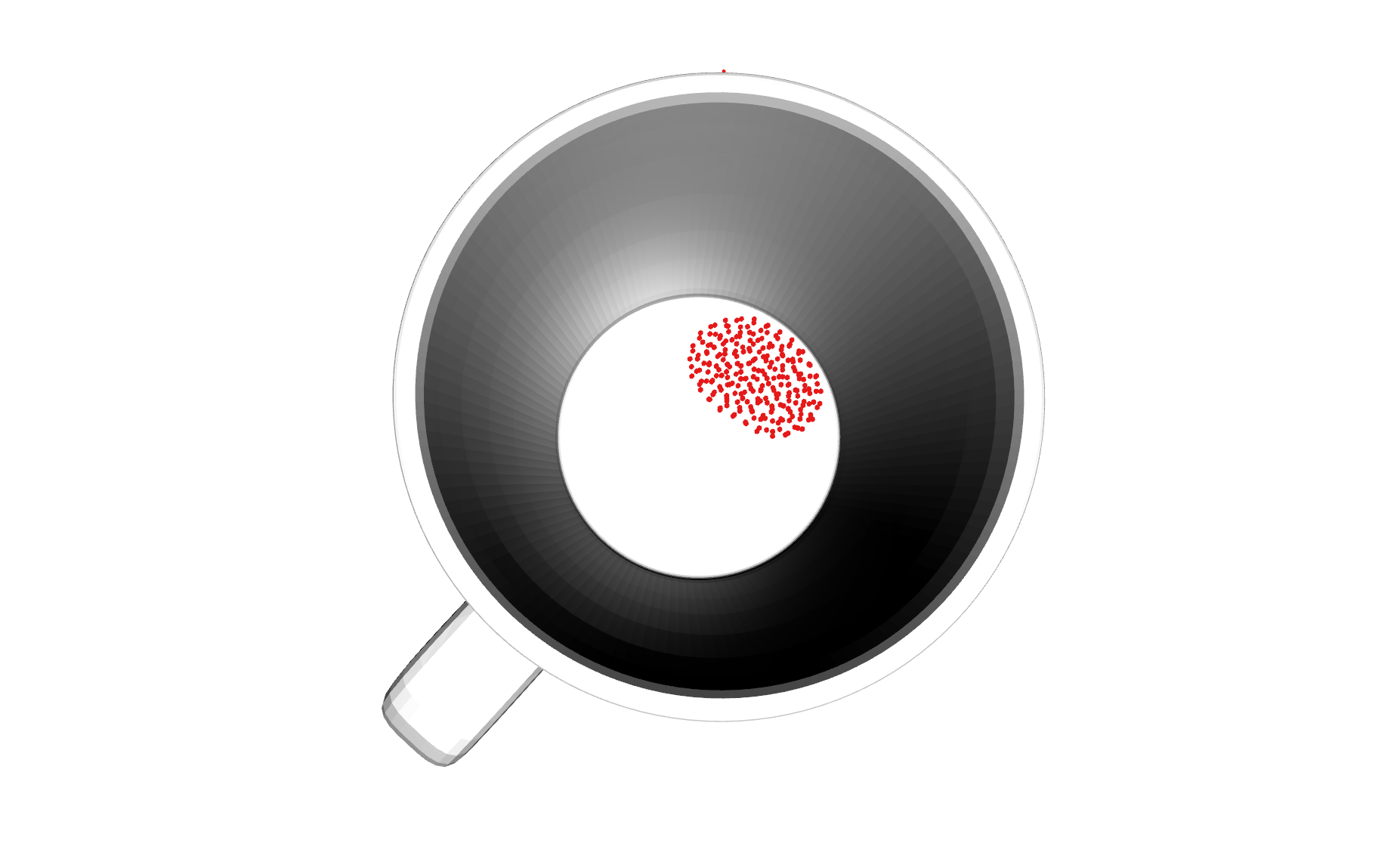}  
        \end{subfigure}
    \end{subfigure}%
    \hspace{2pt}
    \begin{subfigure}{0.155\linewidth}
        \centering
        \begin{subfigure}{1.0\textwidth}
        \includegraphics[trim=13cm 5cm 19cm 0cm, clip,width=\linewidth]{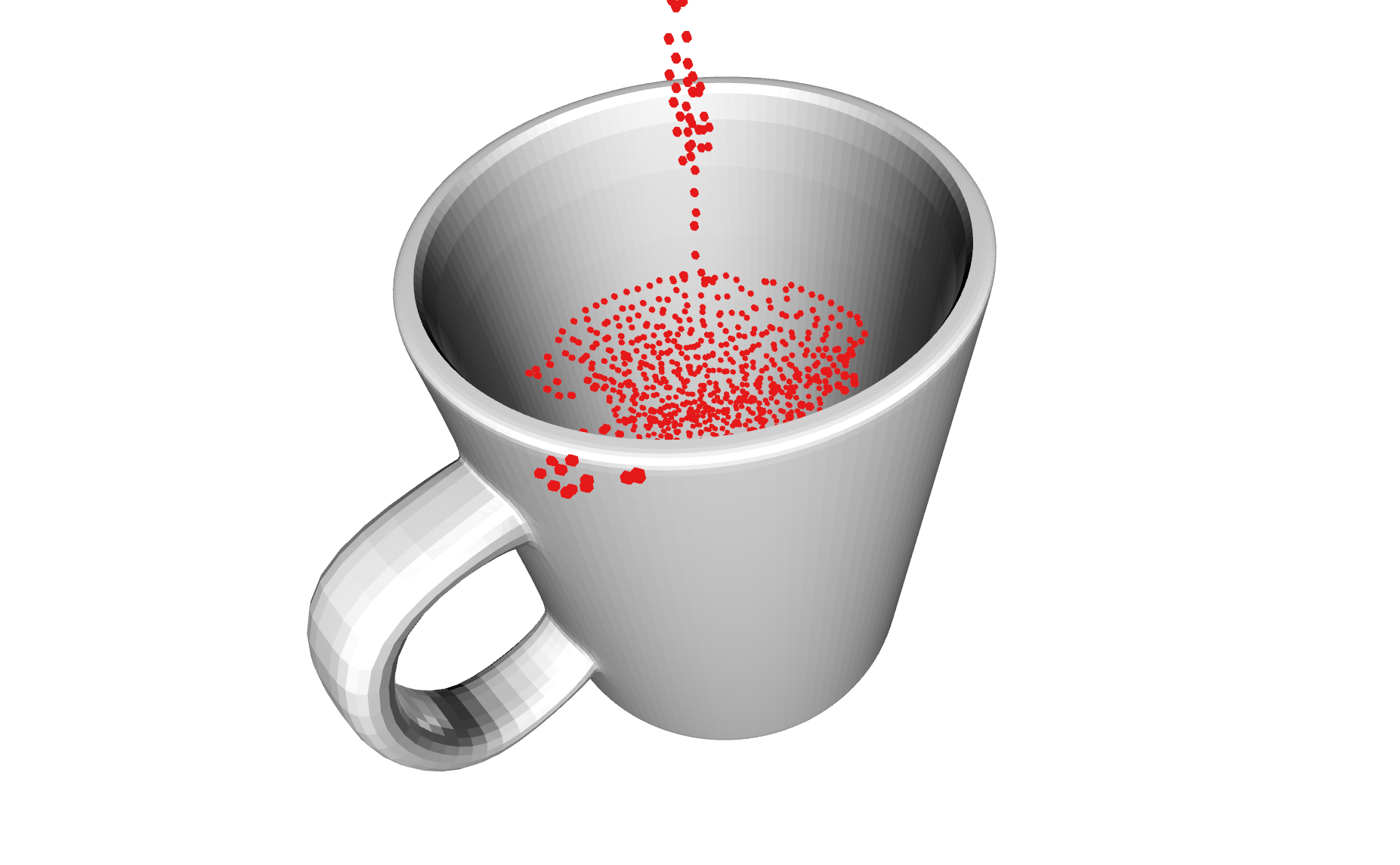}  
        \end{subfigure}
        
        \begin{subfigure}{1.0\textwidth}
        \includegraphics[trim=18cm 6cm 16cm 2cm, clip,width=\linewidth]{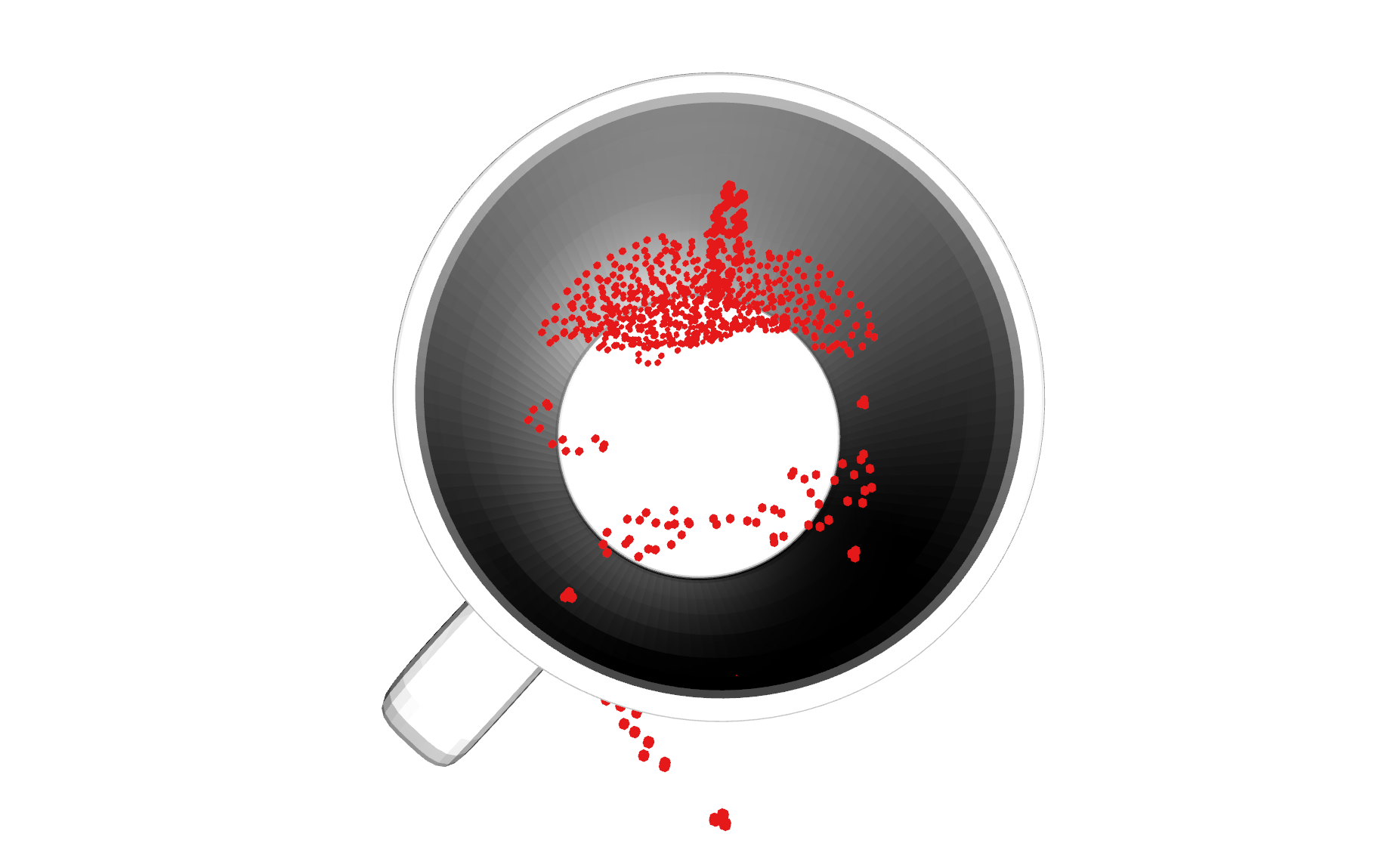}  
        \end{subfigure}
    \end{subfigure}%
    \hspace{2pt}
    \begin{subfigure}{0.155\linewidth}
        \centering
        \begin{subfigure}{1.0\textwidth}
        \includegraphics[trim=13cm 5cm 19cm 0cm, clip,width=\linewidth]{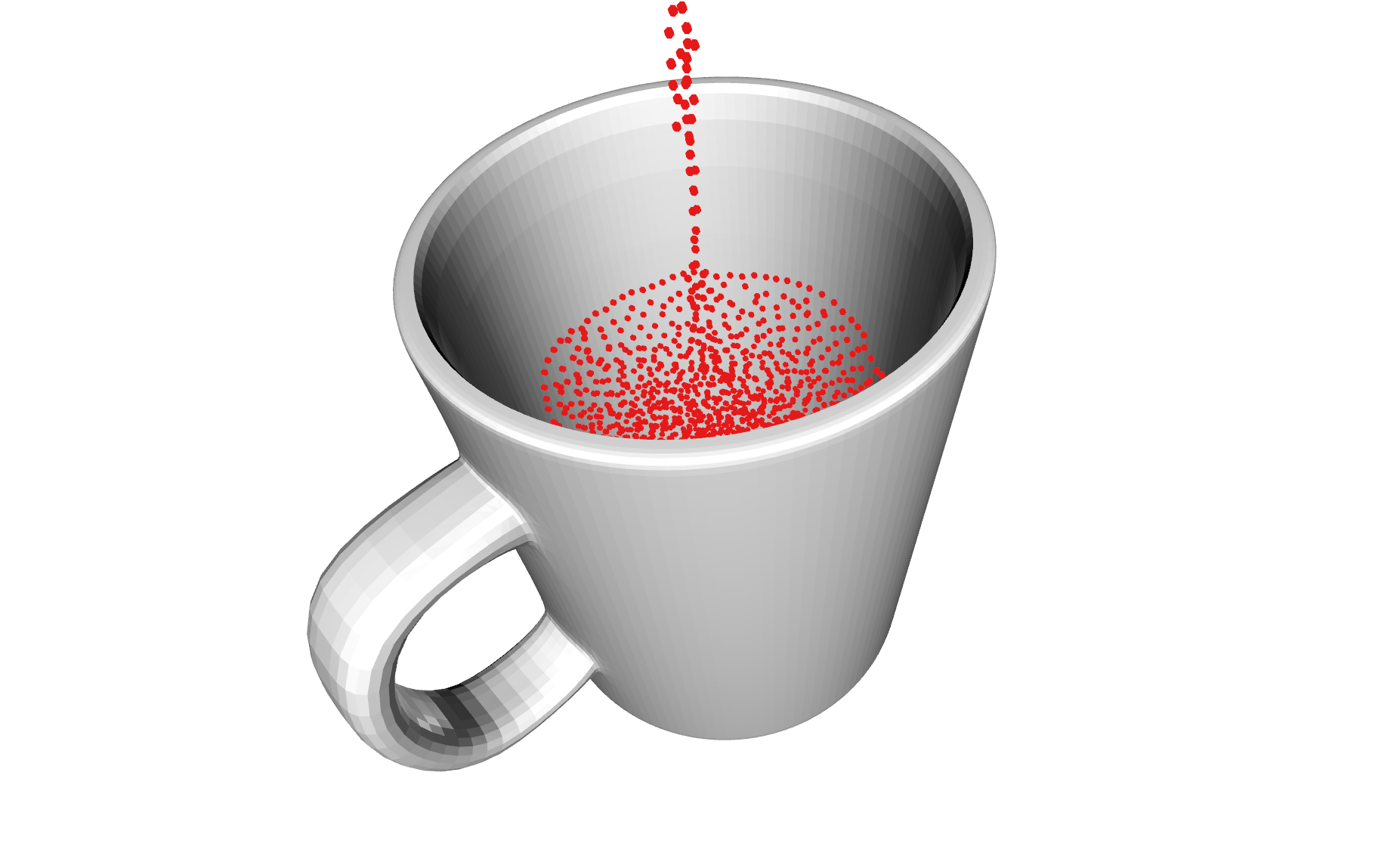}  
        \end{subfigure}
        
        \begin{subfigure}{1.0\textwidth}
        \includegraphics[trim=18cm 6cm 16cm 2cm, clip,width=\linewidth]{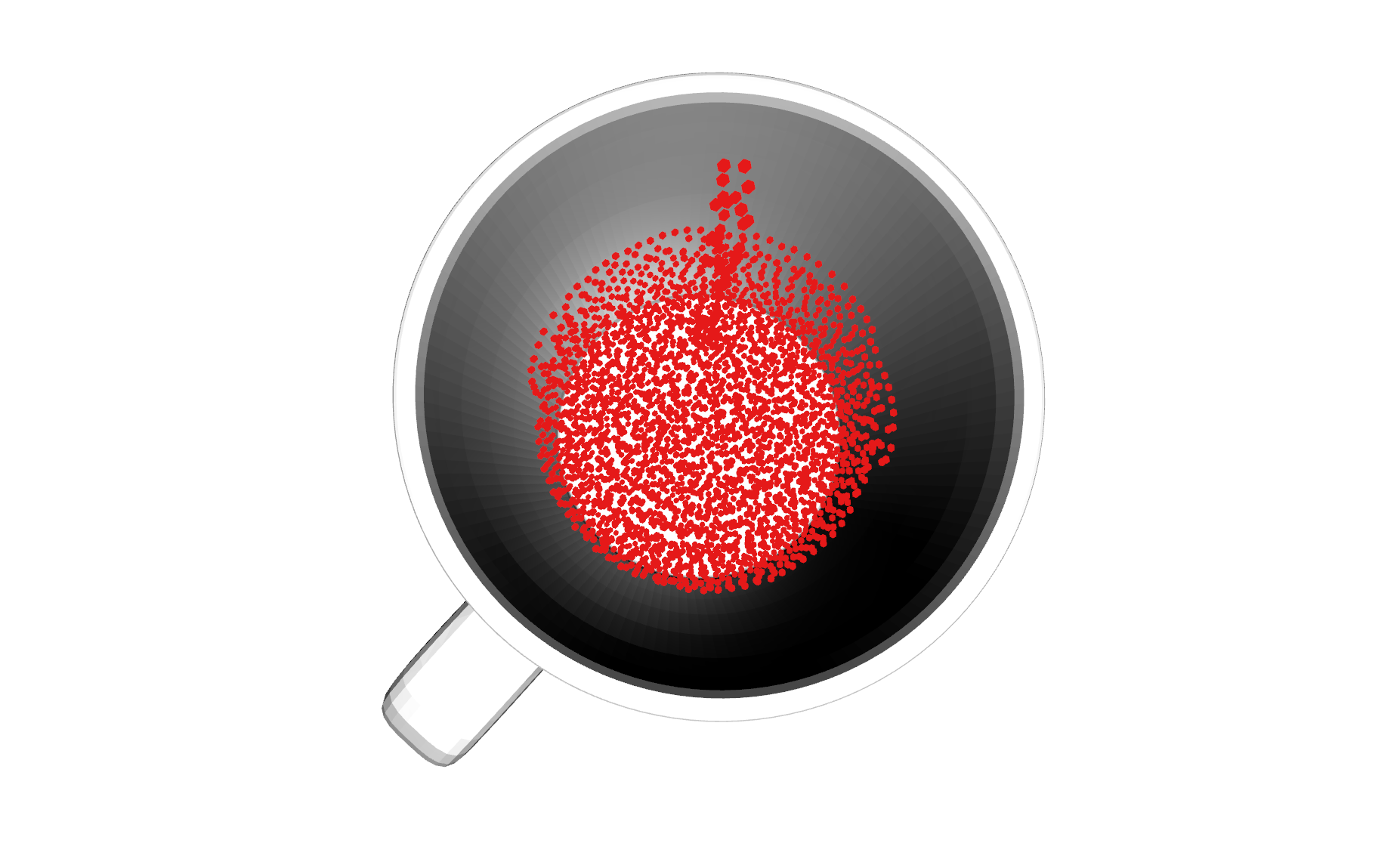}  
        \end{subfigure}
    \end{subfigure}%
    \hspace{1pt}
    \begin{subfigure}{0.155\linewidth}
        \centering
        \begin{subfigure}{1.0\textwidth}
        \includegraphics[trim=13cm 5cm 19cm 0cm, clip,width=\linewidth]{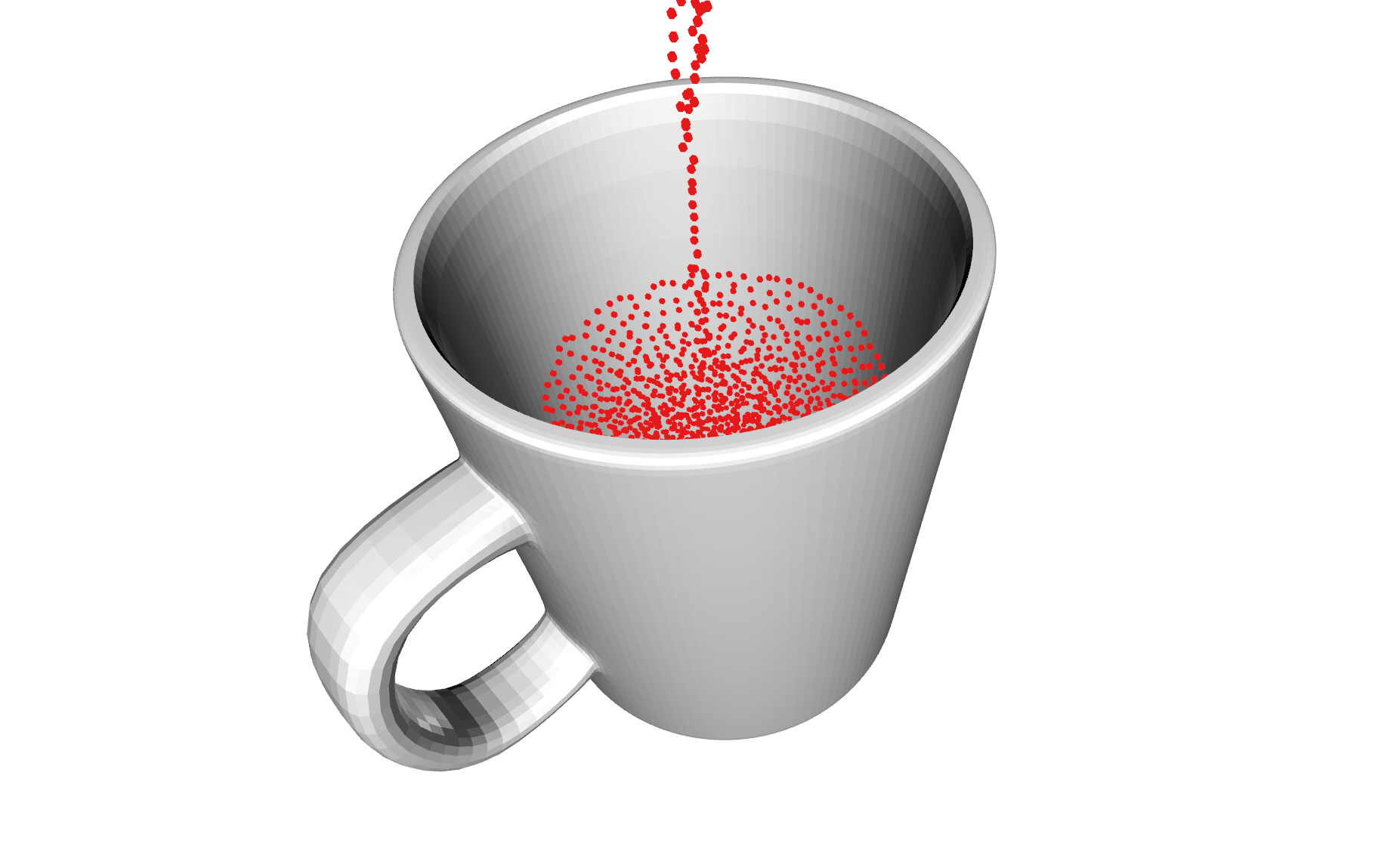}  
        \end{subfigure}
        
        \begin{subfigure}{1.0\textwidth}
        \includegraphics[trim=18cm 6cm 16cm 2cm, clip,width=\linewidth]{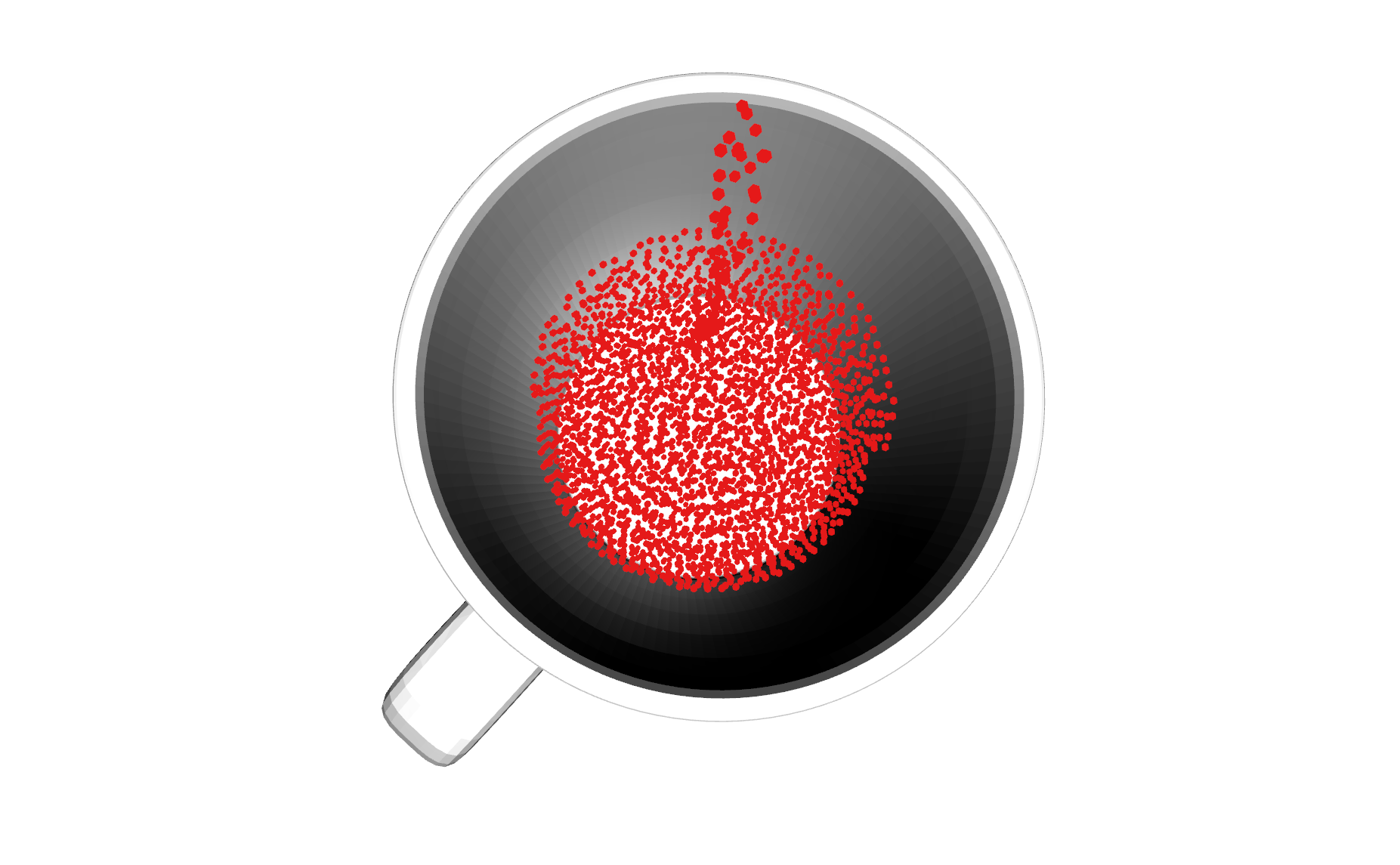}  
        \end{subfigure}
    \end{subfigure}%
    
    \caption{From left to right the image columns are an image from the Pouring Milk dataset being reconstructed with: no density constraint, Schenck \& Fox constraints \cite{schenck2018spnets},  DSS \cite{yifan2019differentiable}, our approach, and our source estimation technique.
    The first row of renderings have the virtual camera positioned similar to the raw image showing how from that perspective, the particles in red line up with the real image of the liquid.
    The second row shows a birds-eye-view perspective and how our proposed approaches properly reconstruct the liquid in 3D.
    The no density constraint and DSS \cite{yifan2019differentiable}  comparisons are unable to properly reconstruct due to over-fitting on the image loss and fail to make inferences in the occluded region.
    Meanwhile Schenck \& Fox constraints \cite{schenck2018spnets} constraints went unstable and splashed particles outside the mug.}
    \label{fig:human_pour_qualitative_results}
\end{figure*}

\begin{figure}[t]
    \centering
    \begin{subfigure}{0.31\linewidth}
        \setlength{\fboxsep}{0pt}
        \setlength{\fboxrule}{1.5pt}
        \fbox{\includegraphics[trim=10cm 25cm 10cm 1cm, clip, width=\linewidth]{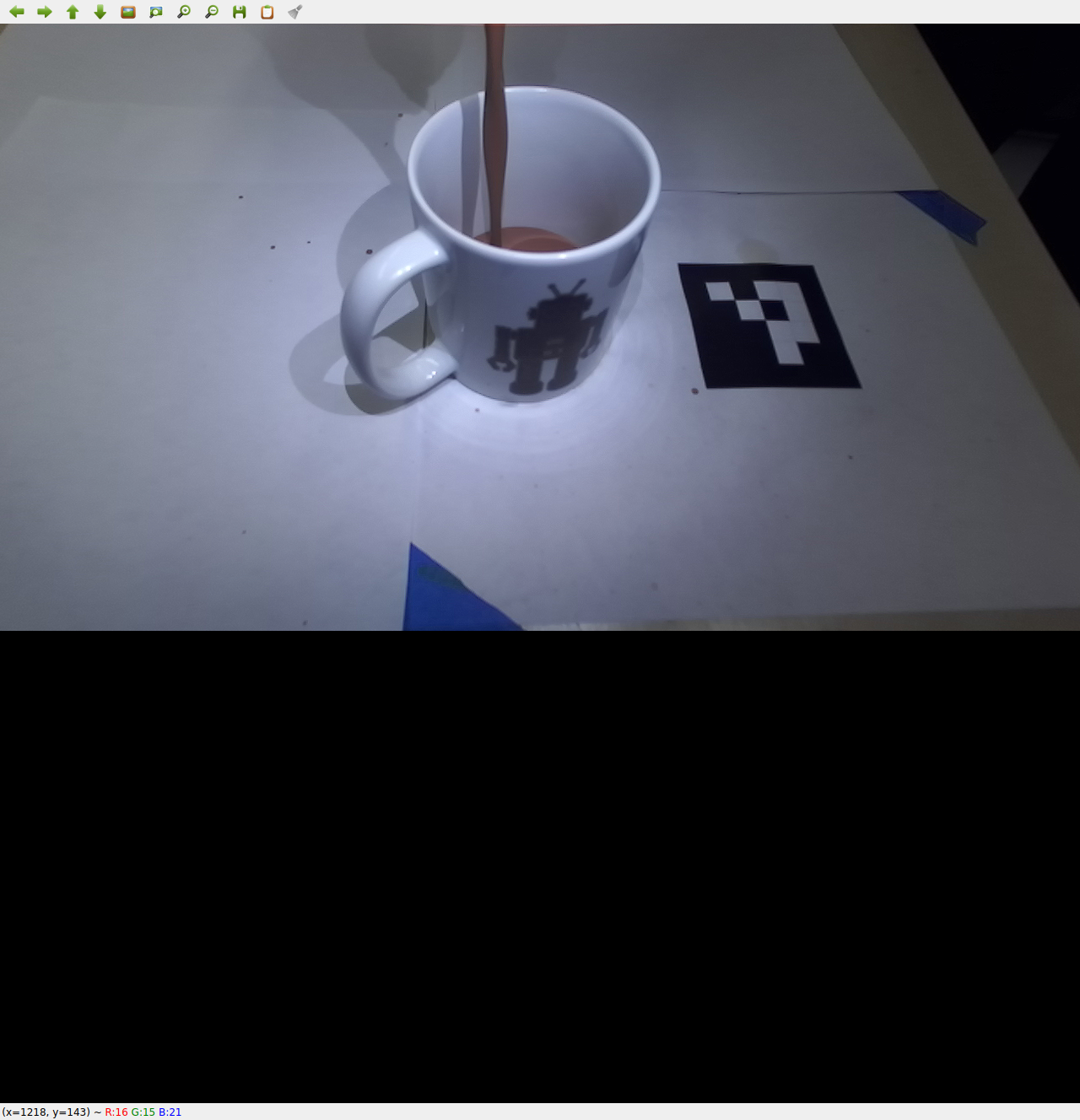}}
    \end{subfigure}%
    \hspace{2pt}
    \begin{subfigure}{0.31\linewidth}
        \includegraphics[trim=10cm 5cm 18cm 5cm, clip, width=\linewidth]{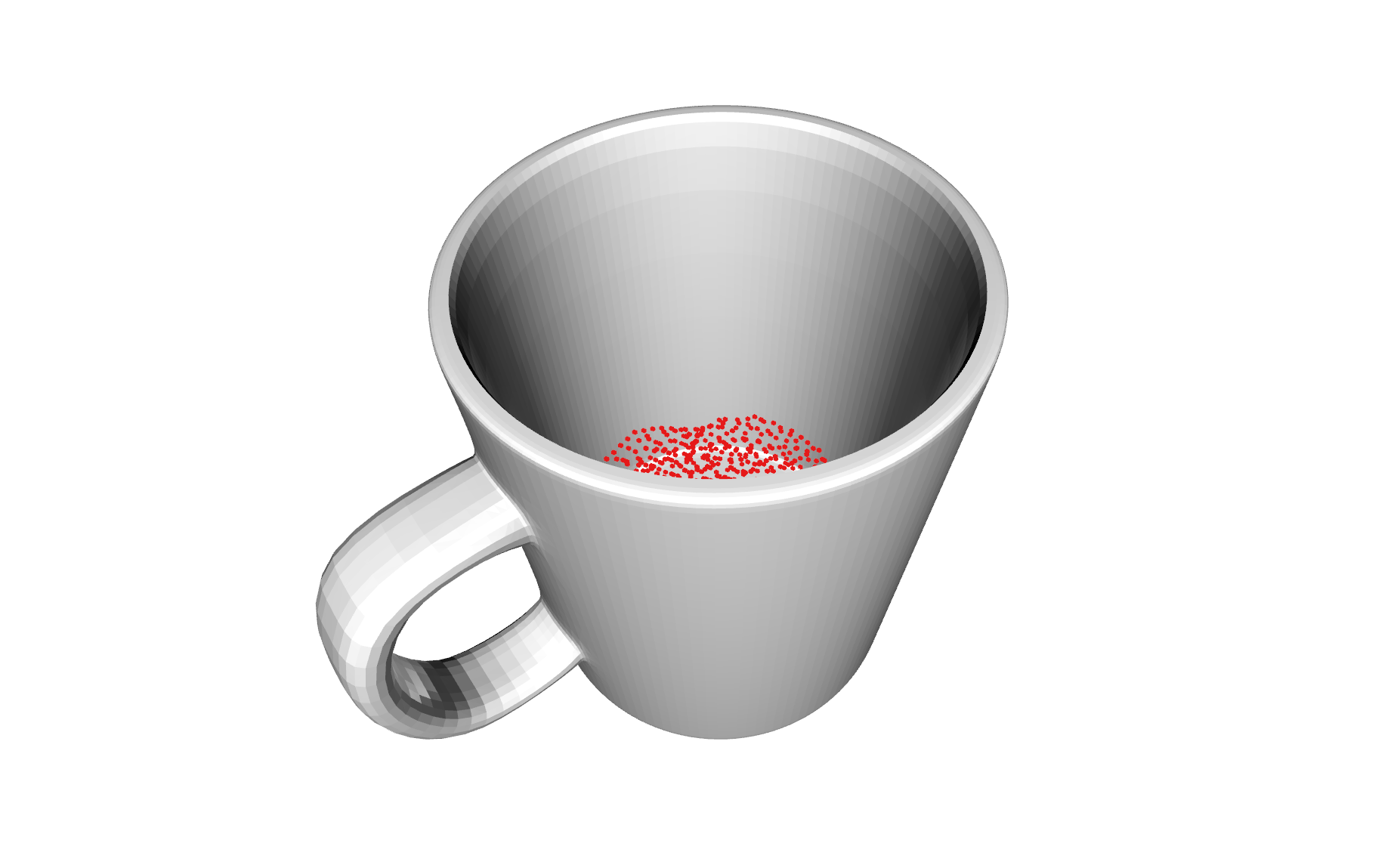}
    \end{subfigure}
    \hspace{2pt}
    \begin{subfigure}{0.31\linewidth}
    \includegraphics[trim=10cm 5cm 18cm 5cm, clip, width=\linewidth]{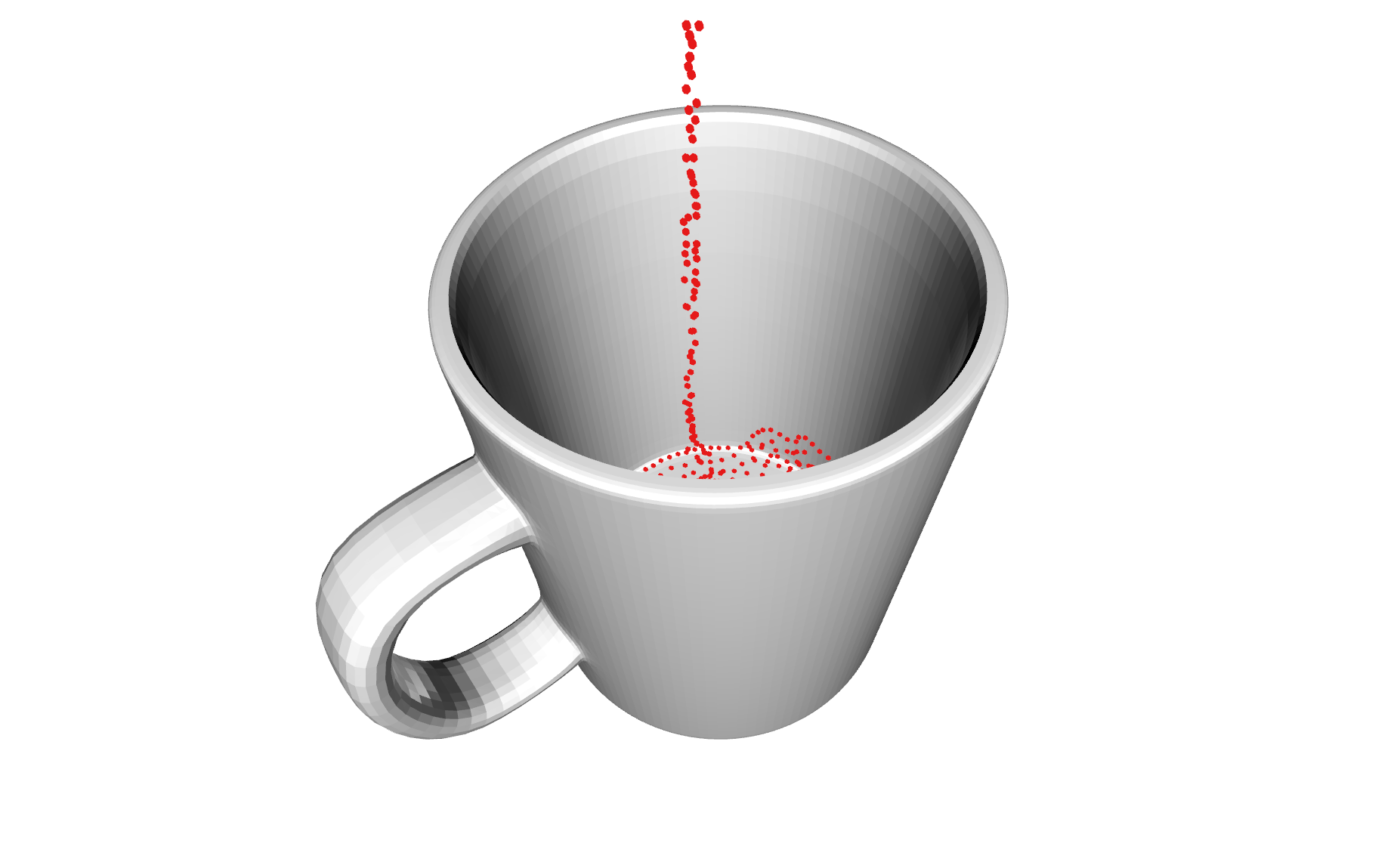}
    \end{subfigure}
    \caption{The left-most image is from the Pouring Milk dataset being reconstructed with the uniform comparison (middle) and our method (right).
    Our method is able to reconstruct the falling stream, unlike the uniform comparison, due to the novel particle insertion and removal approach.
    }
    \label{fig:qualitative_result_uniform}
\end{figure}

When removing the collision constraint (i.e. No Constraints and No Collision comparisions), the particle prediction also had to be turned off otherwise the particles will fall forever due to gravity.
Without the density or collision constraint, the method is equivalent to differentiable rendering \cite{lassner2021pulsar} thus giving a baseline comparison.
Schenck \& Fox proposed their own position-based liquid constraints for constant density \cite{schenck2018spnets} (i.e. replacing $\mathbf{C}_\rho$) which were integrated into this method for comparison.
We also compared with another recently developed differentiable renderer for point-based geometry called Differentiable Surface Splatting (DSS) \cite{yifan2019differentiable}.
Lastly, a source estimation technique is implemented to highlight how the proposed method can be extended.
Implementation details for the Schenk \& Fox, DSS, and Our Source comparisons are given the supplementary material.

Videos and convergence statistics of the comparison study on all the datasets are in the supplementary materials, and a few highlights are given here.
Quantitative results from the Simulated Fountain dataset are shown in Fig. \ref{fig:simulated_results}, and it shows how effective our proposed approach is in a turbulent, long scene.
Fig. \ref{fig:cavity_qualitative_results} shows results from the Endoscopic Trails and how our proposed methods leverage liquid dynamics to fit the cavity shape correctly.
From the Milk Pouring experiments, results are shown in Fig. \ref{fig:human_pour_qualitative_results} and \ref{fig:qualitative_result_uniform} which indicate that our proposed method is able to infer liquid in occluded regions and reconstruct the falling stream.

\section{Discussion and Conclusion}
% NOTE DISCUSSION OF LIMITATIONS: https://cvpr2022.thecvf.com/author-guidelines

Our method is the first approach to reconstructing liquids with only knowledge of the collision environment, gravity direction, and 2D surface detections.
We limited the scope to 2D surface detections because a liquids color is too variable from reflections and refractions.
Our experiments highlight the generalizability of our approach through the wide range of liquids (simulated, water, and milk) and cameras (narrow \& wide field of view and 15, 24 \& 30 fps).
In the supplementary material video, consistent particle flow is observed when using the source estimation extension.
We envision that the source estimation extension will be beneficial in downstream robotic automation applications such as robotic bar tending \cite{wu2020can} and managing hemostasis in surgery \cite{richter2021autonomous} where prediction of the liquid is required.

We found that the density constraint, collision constraint, and prediction are crucial to inferring beyond the 2D image loss as seen in Fig. \ref{fig:cavity_qualitative_results} and \ref{fig:human_pour_qualitative_results}.
Furthermore in longer and more turbulent scenes, the lack of liquid properties can cause instabilities and blow up the mixed-integer optimizer (greater than 10,000 particles).
The density constraint can be switched with other constraints that reflect a liquid incompressibility and other liquid properties, such as Schenck \& Fox's constraints \cite{schenck2018spnets}.
However, we were unable to stabilize Schenck \& Fox's constraints and found the constraint in (\ref{eq:density_constraint}) and its solver to be stable on all of our datasets.
Similar is true for the differentiable rendering, and we found Pulsar to be more robust than DSS in our application since DSS requires normals which we observed are not consistently generated.
Our particle insertion and removal strategy was effective and even able to insert particles to reconstruct a falling stream as seen in Fig. \ref{fig:qualitative_result_uniform}.

There is a large quantity of hyper-parameters in our method, but this is expected when solving a mixed-integer, optimization problem.
Nevertheless, we found a set that generalizes over our diverse datasets, and the interaction radius, $h$, adjusts the effective resolution of our reconstruction (i.e. smaller $h$ gives a denser reconstruction).
An artifact that we observed in our reconstruction approach is over fitting to the image loss during the Pouring Milk experiment.
The top layer of milk is slightly lopsided which is best seen in Fig. \ref{fig:mesh_generation} in the supplemental materials.
This is the result of a challenging trade-off between reliance on observations (i.e. image loss) and dynamics (i.e. liquid prediction).
In future work, we intend on solving the trade-off by modifying (\ref{eq:optimization_problem}) to incorporate multiple timesteps, and hence optimizing with the dynamics.
Furthermore, the dynamics can be incorporated in a differentiable manner through learned graph neural networks which have shown promise in particle based physics \cite{sanchez2020learning}.

{\small
\bibliographystyle{ieee_fullname}
\bibliography{egbib}
}

\newpage

\section{Supplementary Material}

\subsection{DSS Rendering}

For one of our experimental comparisons, we used Differentiable Surface Splatting (DSS) \cite{yifan2019differentiable} to minimize the image loss.
DSS renders each point as a circle, which projects to an ellipse, where the circle's normal is the surface normal.
Surface normals for a particle-represented liquid are computed using the \textit{color field} \cite{muller2003particle} which is
\begin{equation}
    \label{eq:color_field}
    c (\mathbf{p}^i) = \sum \limits_{j=1}^N \frac{1}{\rho_i(\mathbf{p})} W(||\mathbf{p}^i-\mathbf{p}^j||, r)
\end{equation}
at $\mathbf{p}^i \in \mathbb{R}^3$.
% The color field's gradient represents the change of color.
% Therefore, the surface particles can be selected as the particles whose color field has a large gradients at their location.
% So the set of surface points, $\{\tilde{\mathbf{p}}^k\}^{N_s}_{k=1}$, are chosen from $\{\mathbf{p}^i\}^N_{i=1}$ which meet the condition
% \begin{equation}
%     \label{eq:threshold_for_surface}
%     \left| \left| \frac{ \partial c(\tilde{\mathbf{p}}^k) }{ \partial \tilde{\mathbf{p}}^k } \right| \right| > \gamma_h
% \end{equation}x
% where $\gamma_h$ is a set threshold.
The surface normals should point outwards from the reconstructed liquid which results in a negative change in color field.
Therefore the normal is set to:
\begin{equation}
    \label{eq:original_surface_normal}
    \mathbf{n}^i = - \left( \frac{ \partial c(\mathbf{p}^i) }{ \partial \mathbf{p}^i } \right) \bigg/ \left| \left| \frac{ \partial c(\mathbf{p}^i) }{ \partial \mathbf{p}^i } \right| \right|
\end{equation}
for particle $\mathbf{p}^i$.
% Surface particles and normals are selected and computed in line \ref{alg:get_surface} in Algorithm \ref{alg:main_outline}.
To compute the liquid volume color's gradient, the following expression is used:
\begin{equation}
    \frac{ \partial c(\mathbf{p}^i) }{ \partial \mathbf{p}^i } = \sum \limits_{j=1}^N \frac{1}{\rho_i}
    \frac{\partial W(||\mathbf{p}^i-\mathbf{p}^j||, r)}{\partial ||\mathbf{p}^i-\mathbf{p}^j||} 
    \left( \frac{\mathbf{p}^i - \mathbf{p}^j}{ ||\mathbf{p}^k-\mathbf{p}^j|| } \right)
    \label{eq:derivative_of_color_field}
\end{equation}
after applying the chain rule to (\ref{eq:color_field}).

Laplacian smoothing is applied for more consistent normals, similar to \cite{yu2013reconstructing}, by averaging the particle positions the color field is being evaluated about in (\ref{eq:color_field}).
The particle averages are computed as
\begin{equation}
    \label{eq:laplacian_smoothing}
    \overline{\mathbf{p}}^j = (1 - \lambda_l) \mathbf{p}^j + \lambda_l \frac{ \sum \limits_{k=1}^N \mathbf{p}^k W(||\mathbf{p}^j-\mathbf{p}^k||, r)} {\sum \limits_{k=1}^N W(||\mathbf{p}^j-\mathbf{p}^k||, r)}
\end{equation}
where $\lambda_l$ is the Laplacian average weight and $\overline{\mathbf{p}}^j$ replaces $\mathbf{p}^j$ in (\ref{eq:color_field}).
In low particle count situations, the normal computation in (\ref{eq:original_surface_normal}) can produce undesirable effects such as artifacts on the edges of the liquids.
To account for this, the evaluation of points $\mathbf{p}^i$ in (\ref{eq:original_surface_normal}) are given a small offset towards the virtual camera which will eventually render the surface.
The offset is computed as:
\begin{equation}
    \Delta \overline{\mathbf{p}}^i = \lambda_c \frac{
    \mathbf{c} - \overline{\mathbf{p}}^i}{||\mathbf{c} - \overline{\mathbf{p}}^i||}
\end{equation}
where $\mathbf{c} \in \mathbb{R}^3$ is the position of the virtual camera, $\lambda_c$ is the amount of the offset, and $\Delta \overline{\mathbf{p}}^i$ is added to $\mathbf{p}^i$ in (\ref{eq:original_surface_normal}).

The particle position, normal pairs, $\{\mathbf{p}^i, \mathbf{n}^i \}_{i=1}^N$, are directly fed into the DSS which renders each point, $\mathbf{p}^i$, as a circle whose plane is tangent to its normal, $\mathbf{n}^i$.
The circles are projected to ellipses, denoted as $\mathcal{E}(\mathbf{p}^i, \mathbf{n}^i)$, and averaged with their neighboring projected circles, hence being called Elliptical Weighted Averaging (EWA).
In the problem formulation for this work, we assume only knowledge of an observed visibility mask $\mathbb{I}$.
Therefore, we simplify the rendering by not conducting the EWA and only render a surface mask from the projected ellipses.
Written mathematically, the masked image at pixel $[u,v]^\top$ from a single particle and normal pair is:
DSS computes each 
\begin{equation}
    \label{eq:single_particle_mask}
    h_{u,v} (\mathbf{p}^i, \mathbf{n}^i ) = \begin{cases} 1  & \mbox{if } [u,v]^\top \in \mathcal{E}(\mathbf{p}^i, \mathbf{n}^i) \\
    0 & \mbox{if $\mathbf{p}^i$ is occluded }\\
    0 & \mbox{otherwise}  \end{cases}
\end{equation}
The rendered surface is evaluated as a summation of all the masked images from (\ref{eq:single_particle_mask}):
\begin{equation}
    \label{eq:surface_rendering}
    \hat{\mathbb{I}}_{u,v}(\mathbf{p}) = \eta_i \sum \limits_{i=1}^{N_s} h_{u,v} (\mathbf{p}^i, \mathbf{n}^i)
\end{equation}
where $\eta_i$ normalizes the pixel value.
Finally, gradients of the rendered liquid surface with respect to particle positions are computed using the approximation presented by Yifan et al. to minimize the image loss \cite{yifan2019differentiable}.
The normal smoothing values are set to $\lambda_l = 0.2$ and $\lambda_c = 0.2r$, and the original proposed kernels are used for (\ref{eq:color_field}) \cite{muller2003particle} and (\ref{eq:laplacian_smoothing}) \cite{yu2013reconstructing}.
The gradient step size and its threshold for detecting a local-minima are set to $\alpha_\mathbb{I} = 10^{-4}$ and $\lambda_s = 0.2$ respectively.

\subsection{Schenck and Fox Constraints}

Schenck and Fox previously proposed liquid position constraints that represent: pressure, cohesion, and surface tension \cite{schenck2018spnets}.
These constraints replaced the proposed density constraint from (\ref{eq:density_constraint}) for comparison in our experiments.
This is done by replacing $\Delta \mathbf{p}_\rho$ to solve (\ref{eq:density_constraint}) in lines (\ref{alg:density_constraint_1}) and (\ref{alg:density_constraint_2}) in Algorithm \ref{alg:main_outline} with:
\begin{equation}
    \Delta \mathbf{p}_p + \alpha_c \Delta \mathbf{p}_c + \alpha_s \Delta \mathbf{p}_s
\end{equation}
where $\Delta \mathbf{p}_p, \Delta \mathbf{p}_c, \Delta \mathbf{p}_s$ solve the pressure, cohesion, and surface tension constraints respectively and $\alpha_c, \alpha_s$ are the cohesion and surface tension weights respectively.
Refer to the original paper for exact expressions to the constraint solutions \cite{schenck2018spnets}.
The cohesion and surface tension weights are optimized for in the original work to conduct real-to-sim registration.
However, this cannot be done with our problem setup because it requires prior information on the amount of liquid volume there is (i.e. how many particles there are).
Therefore, instead the weights are preset to $\alpha_c = 0.05$ and $\alpha_s = 0$ (the surface tension constraint only yielded unstable behavior so it was turned off).

\begin{figure*}[t]
    \centering
    \begin{subfigure}{0.24\linewidth}
        \includegraphics[trim=16cm 0cm 16cm 0cm, clip, width=\linewidth]{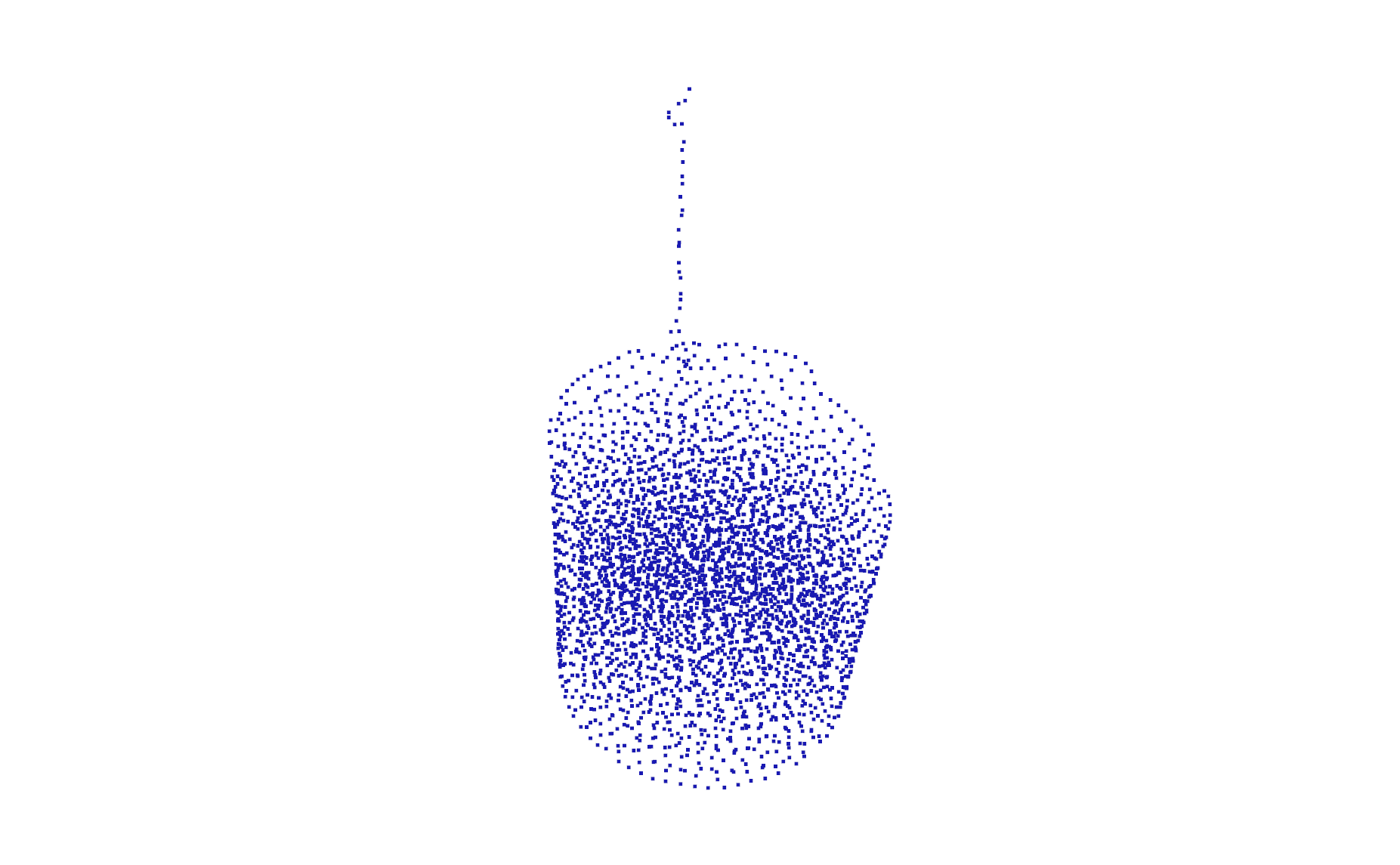}  
    \end{subfigure}%
    \hspace{2pt}
    \begin{subfigure}{0.24\linewidth}
        \centering
        \includegraphics[trim=16cm 0cm 16cm 0cm, clip, width=\linewidth]{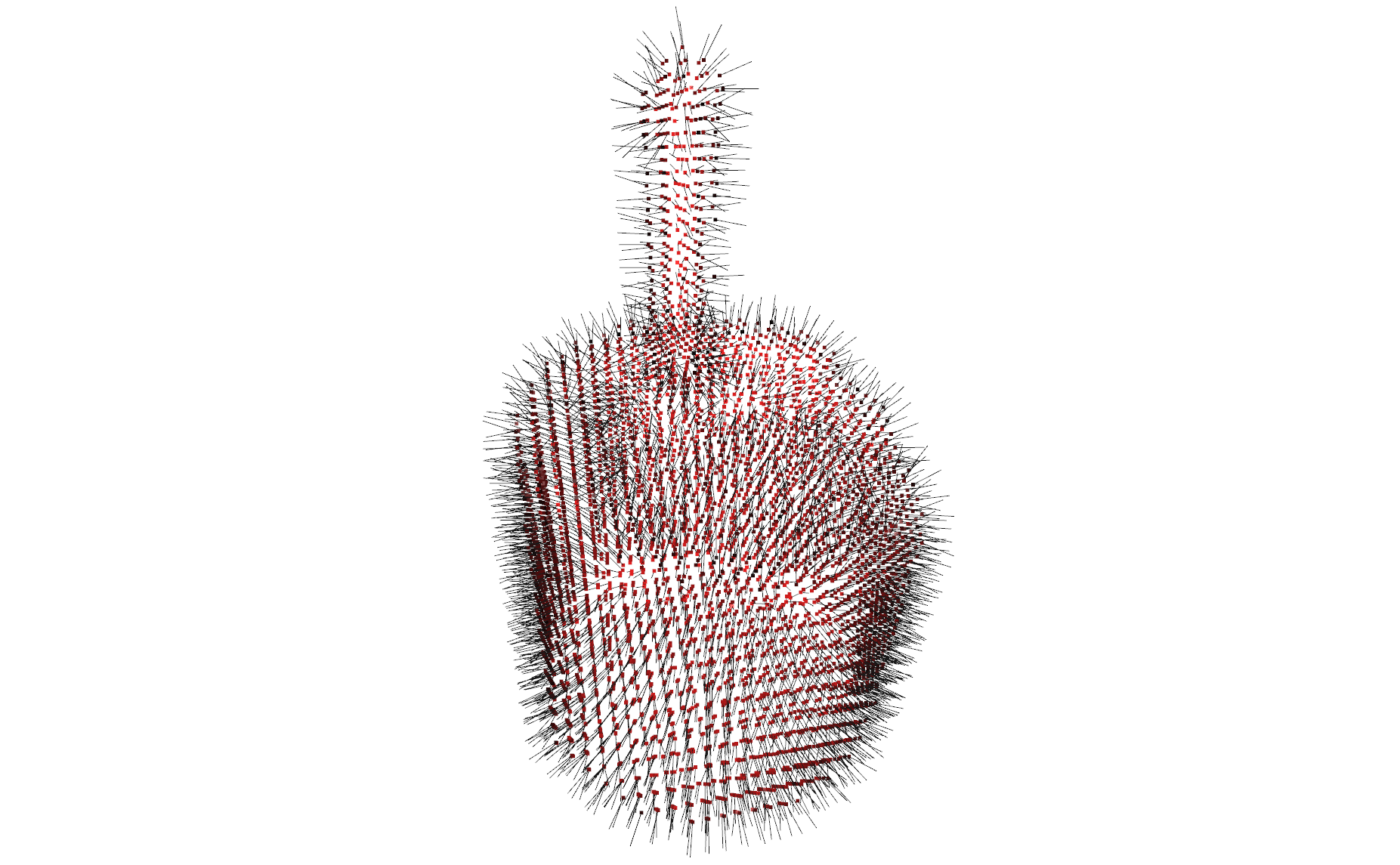}  
    \end{subfigure}%
    \hspace{2pt}
    \begin{subfigure}{0.24\linewidth}
        \centering
        \includegraphics[trim=16cm 0cm 16cm 0cm, clip, width=\linewidth]{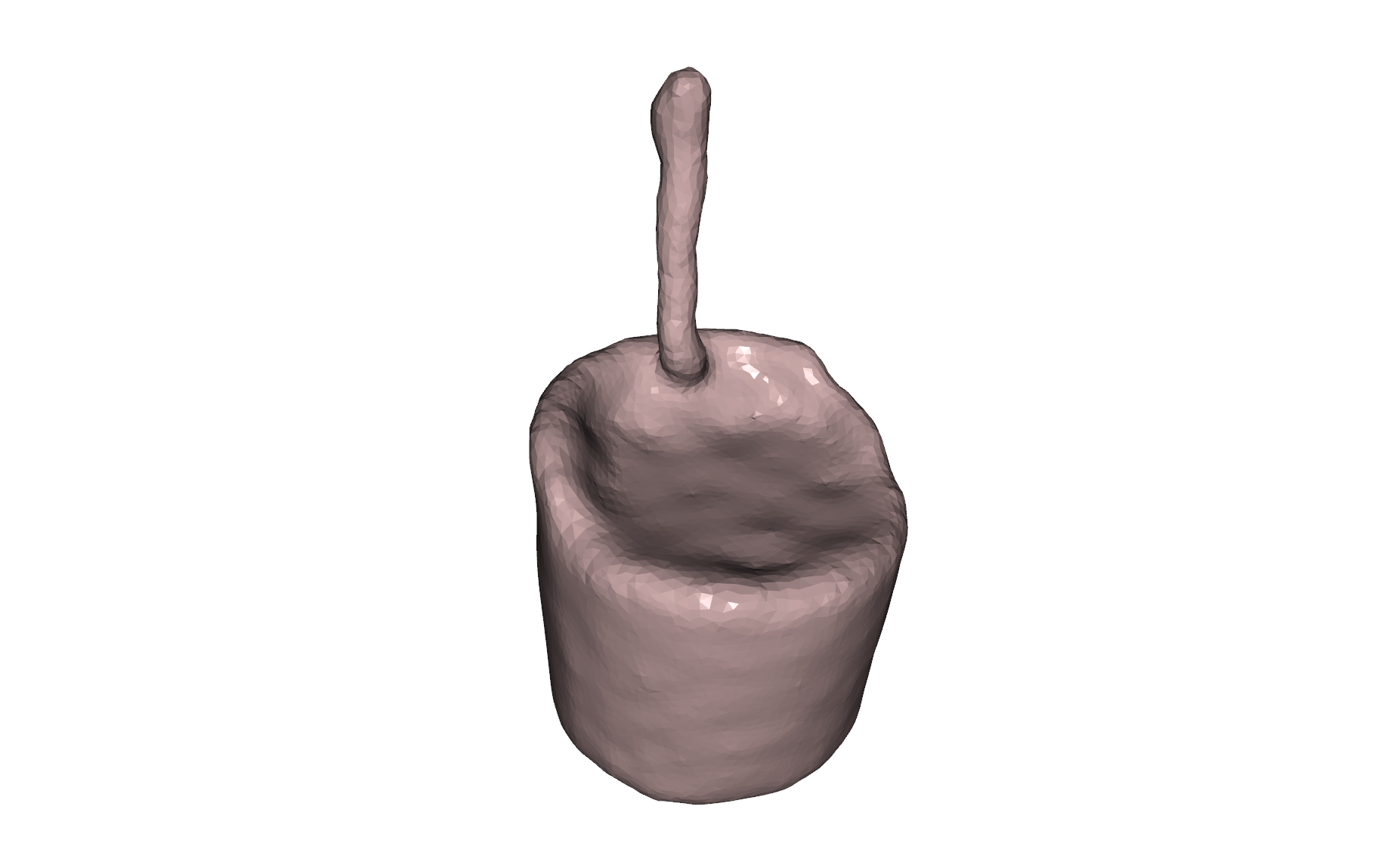}  
    \end{subfigure}%
    \hspace{2pt}
    \begin{subfigure}{0.24\linewidth}
        \centering
        \includegraphics[trim=16cm 0cm 16cm 0cm, clip, width=\linewidth]{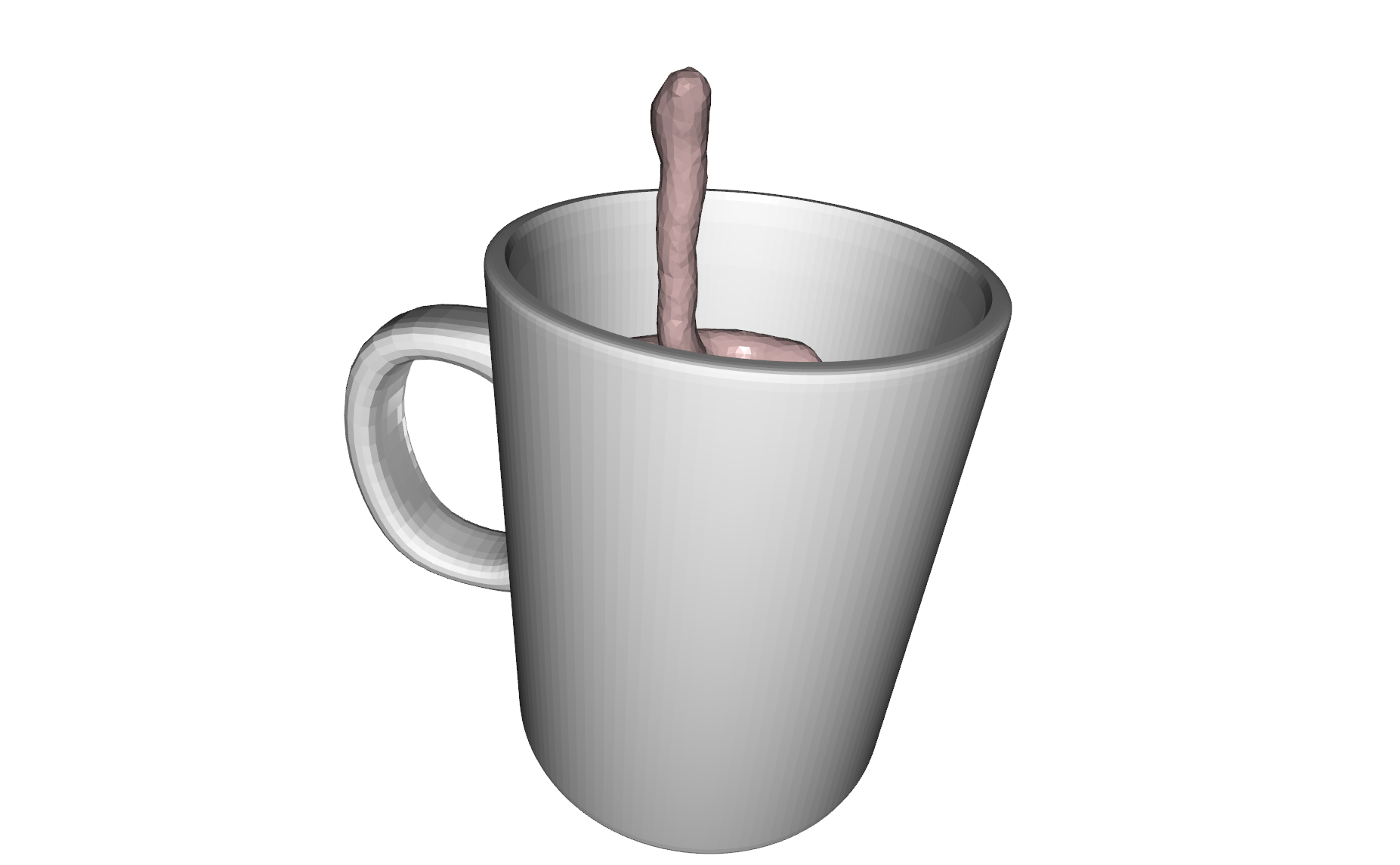}  
    \end{subfigure}%
    \hspace{2pt}
    \caption{From left to right, the sequence shows the steps for mesh generation from the reconstructed liquid. The left-most figure is the reconstructed liquid in particle representation. The next figure shows the densely generated surface points and normals from the reconstructed liquid. The last two figures show the generated surface mesh from the surface points and normals without and with the collision mesh.}
    \label{fig:mesh_generation}
\end{figure*}

\subsection{Source Estimation}

A simple single, static source estimation technique is developed to highlight how the proposed method can be extended.
Let $\hat{\mathbf{s}}_t \in \mathbb{R}^3$ be the estimated liquid source location in the camera frame at time $t$ and particles are inserted according to an estimated flow rate of $\hat{f}_t$ particles per timestep at the source location.
% The particle insertion from source is done in lines \ref{alg:insert_source_particles_1} and \ref{alg:insert_source_particles_2} in Algorithm \ref{alg:main_outline} where $\mathbf{p}_s = [\mathbf{p}^1_s, \dots, \mathbf{p}^{\hat{f}_t}_s]$.
Note that no velocity prediction is conducted for the inserted particles as there is no initial velocity.
This reduces the number of parameters to estimate to $\hat{\mathbf{s}}_t$ and $\hat{f}_t$.

To update the liquid source location, $\hat{\mathbf{s}}_t$, we compare the source particle locations after completing the optimization in (\ref{eq:optimization_problem}) against their pre-optimized location.
Let the initial and optimized particle locations emitted from the source be denoted as $\mathbf{p}^{n}_s \in \mathbb{R}^3$ for $n=1,\dots, \hat{f}_t$ and $\mathbf{p}^{n*}_s \in \mathbb{R}^3$ for $n=1,\dots, \hat{f}_t$ respectively.
Then the update rule given to the source location is:
\begin{equation}
    \hat{\mathbf{s}}_{t+1} = \hat{\mathbf{s}}_{t} + \frac{\alpha_\mathbf{\hat{s}}}{\hat{f}_t} \sum \limits_{n=1}^{\hat{f}_t} (\mathbf{p}^{i*}_s - \mathbf{p}^{i}_s)
\end{equation}
where $\alpha_\mathbf{\hat{s}}$ is adjusted according to:
\begin{equation}
    \alpha_\mathbf{\hat{s}} = 1 / \sum \limits_{i=1}^{t-1} \hat{f}_i
\end{equation}
so the source becomes less adjusted as more particles have been inserted since the source is assumed stationary.

The liquid source rate, $\hat{f}_t$, has an integer effect on the reconstruction, and we adjust it at every time step based on how many particles are duplicated or removed during the optimization of (\ref{eq:optimization_problem}) after inputting the source particles for that timestep.
The expression is:
\begin{equation}
    \label{eq:estimate_source_rate}
    \hat{f}_t =  \alpha_{\hat{f}} \Delta N_t + \hat{f}_{t-1} - \lambda_f
\end{equation}
where $\Delta N_t$ is the cumulative increase of particles (e.g. could be negative if particles are removed) at timestep $t$, $\alpha_{\hat{f}}$ adjusts the reaction rate to the insertion/removal of particles, and $\lambda_f$ is a constant decay rate.
Note that $\hat{f}_t$ is estimated as a non-integer value, however is applied as an integer by rounding (i.e. only an integer number of particles can be inserted per timestep).
The decay rate, $\lambda_f$ is used ensures stability by driving the flow rate to 0 when no new information from $\Delta N_t$ can be leveraged.
The reaction rate and decay rates are set to $\alpha_f = 0.1$ and $\lambda_f = \alpha_{\hat{f}}/2$ respectively.

\begin{figure*}[t]
    \centering
    \begin{subfigure}{0.98\linewidth}
        \includegraphics[trim=0cm 3.1cm 2.2cm 0cm, clip, width=\linewidth]{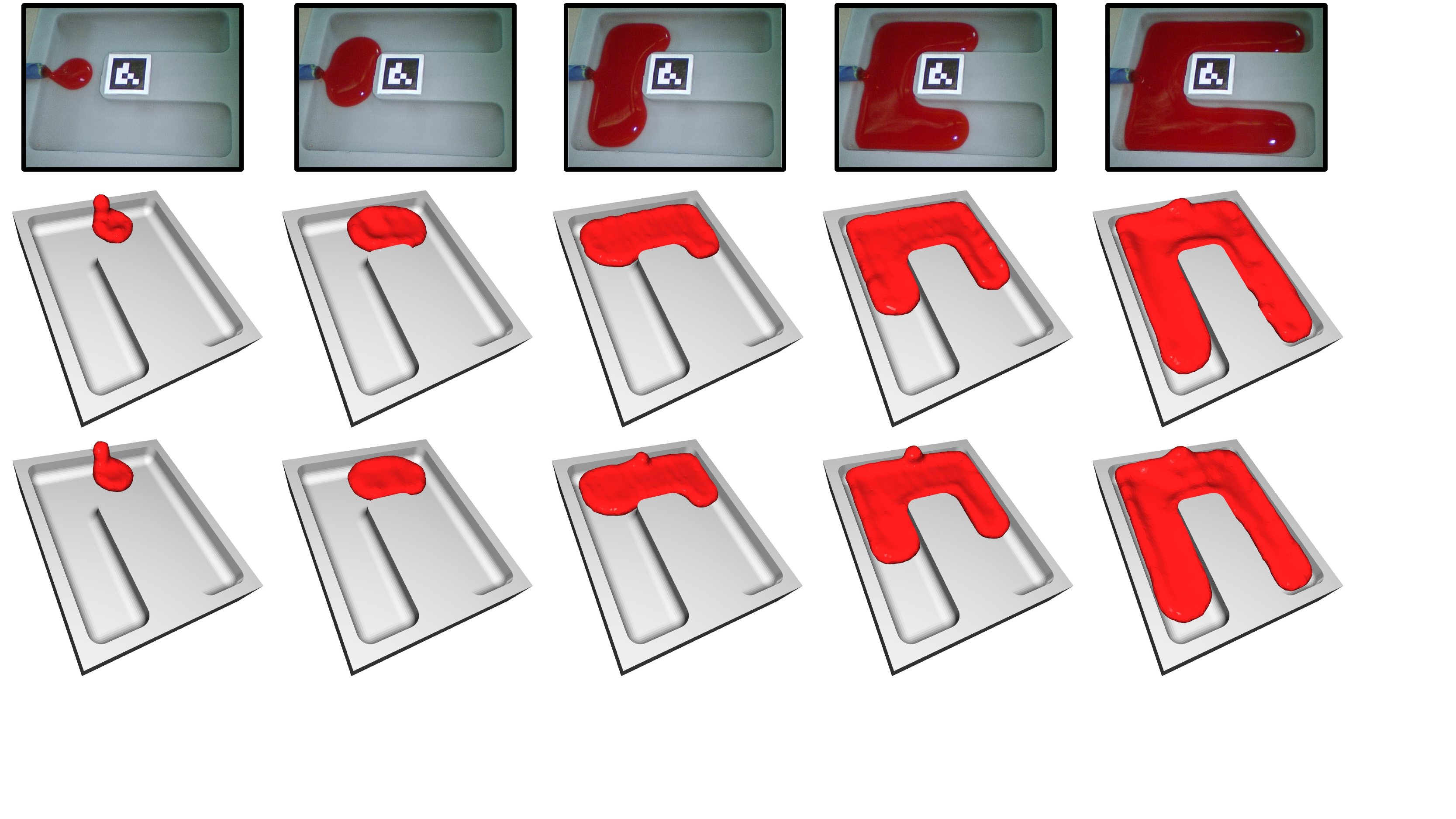}  
    \end{subfigure}%
    \caption{Sequence of reconstruction results from Endoscopic Trail 3 where the rows from top to bottom show: endoscopic image, our complete approach, and our source approach.}
    \label{fig:sequence_cavity}
\end{figure*}

\begin{figure*}[t]
    \centering
    \begin{subfigure}{0.98\linewidth}
        \includegraphics[trim=0cm 9.2cm 1cm 0cm, clip, width=\linewidth]{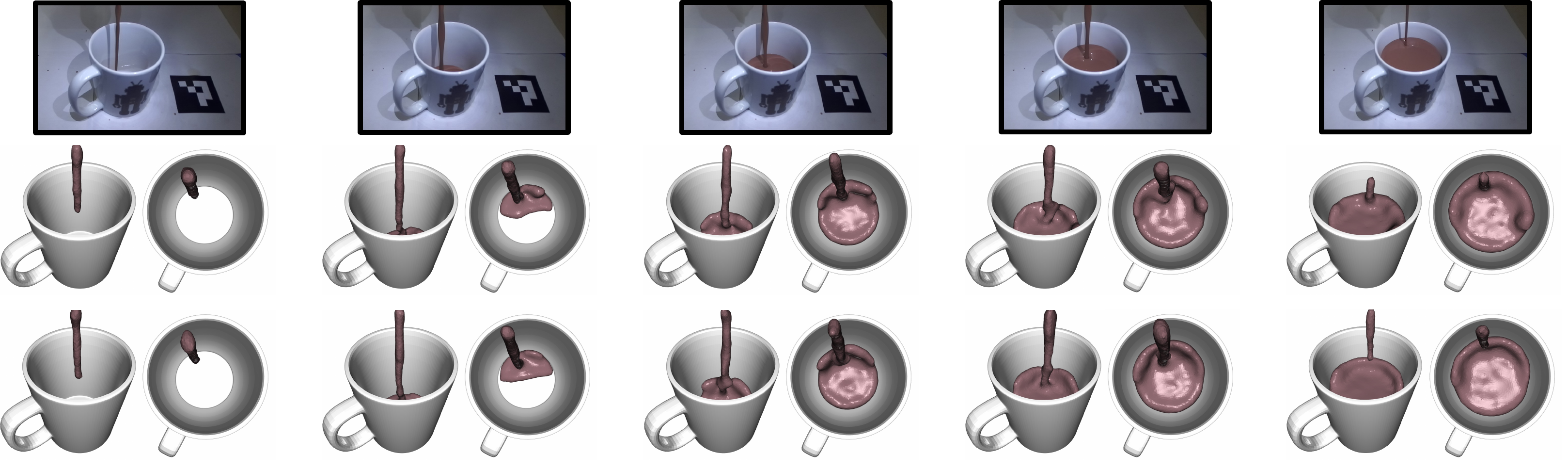}  
    \end{subfigure}%
    \caption{Sequence of reconstruction results from Pouring Milk dataset where the rows from top to bottom show: image, our complete approach, and our source approach.}
    \label{fig:sequence_pour}
\end{figure*}

\subsection{Mesh Generation}

For visualization purposes, the reconstructed liquid can be converted to a surface mesh.
A dense, uniformly spaced, grid of 3D points is generated.
Surface points, $\mathbf{g}^k$, from the grid points are then selected by thresholding the gradient of the color field \cite{muller2003particle}:
\begin{equation}
     \partial c(\mathbf{g})/ \partial \mathbf{g}^k  \geq \lambda_g
\end{equation}
where the color field, $c(\cdot)$, is defined in (\ref{eq:color_field}) and $\lambda_g$ is the threshold.
The surface normals for each surface point is computed the same as (\ref{eq:original_surface_normal}).
The collection of surface points and normals are then converted to a mesh using Open3D's implementation of \cite{zhou2018open3d} Poisson surface reconstruction \cite{kazhdan2006poisson}.
Fig. \ref{fig:mesh_generation} shows an example of this process.
The grid points, which the surface points are selected from, are spaced at 3mm, the gradient threshold, $\lambda_g$, is set to 0.5, and the depth for Poission surface reconstruction is set to 12.
Note that figures of particles and mesh renderings in this paper are done with Open3D \cite{zhou2018open3d}.

\begin{table*}    
    \centering
    \fontsize{8}{10}
    \begin{tabular}{ c | c | c | c | c | c}
     \fontsize{10}{12}{\textbf{Method}} & Simulation & Endo Trail 1 &  Endo Trial 2 & Endo Trail 3 & Pouring   \\ \hline
     No Constraints \cite{lassner2021pulsar} & $-0.03 \pm 0.12$ & $-0.25 \pm 0.20$  & $-0.09 \pm 0.23$  & $-0.01 \pm 0.29$ & $-0.08 \pm 0.14$  \\  \hline
     No Density & $0.89 \pm 3.7$ & $0.48 \pm 0.65$ & $1.0 \pm 0.94$ & $0.91 \pm 1.0$ &  $66 \pm 85$  \\  \hline
     No Collision & $(-5.26 \pm 61)10^{-5}$ & $(-3.2 \pm 4.9) 10^{-3}$ & $(-1.4 \pm 3.3) 10^{-3}$ & $(-3.7 \pm 8.7) 10^{-4}$ & $(-0.33 \pm 11) 10^{-3}$  \\ \hline
     Schenck \& Fox \cite{schenck2018spnets} & $(-2.3 \pm 83)10^{-2}$ & $-0.11 \pm 0.04$ & $-0.12 \pm 0.05$ & $-0.11 \pm 0.04$ & $0.21 \pm 0.37$ \\ \hline
     DSS \cite{yifan2019differentiable} & $(-1.3 \pm 25)10^{-3}$& $(-4.6 \pm 8.4) 10^{-3}$  & $(-4.9 \pm 6.6) 10^{-3}$ & $(-6.4 \pm 6.4) 10^{-3}$ & $(-5.2 \pm 32) 10^{-3}$  \\ \hline
     Uniform & $(-0.05 \pm 14)10^{-3}$ & $(-2.1 \pm 4.5)10^{-3}$ & $(-2.2 \pm 3.4)10^{-3}$ & $(-1.4 \pm 3.8)10^{-3}$ & $(1.3 \pm 7.3)10^{-3}$  \\ \hline
     No Prediction & $(-0.51 \pm 9.9)10^{-3}$ &$(-3.1 \pm 4.6)10^{-3}$  & $(-1.5 \pm 3.4)10^{-3}$ & $(-3.5 \pm 5.0)10^{-3}$ & $(-4.6 \pm 14)10^{-3}$ \\ \hline
     Ours & $(1.2 \pm 13)10^{-2}$ &$(-1.7 \pm 3.8)10^{-3}$  & $(-2.3 \pm 4.1)10^{-3}$ & $(-1.2 \pm 3.0)10^{-3}$ & $(-0.08 \pm 4.4)10^{-3}$  \\ \hline
     Our Source & $(-1.6 \pm 15)10^{-2}$ & $(-1.2 \pm 3.5)10^{-3}$ & $(-1.9 \pm 4.5)10^{-3}$ & $(-1.6 \pm 8.5)10^{-3}$ & $(-0.24 \pm 7.3)10^{-3}$  \\ 
    \end{tabular}
    \caption{Mean and standard deviation of the density constraint, defined in (\ref{eq:density_constraint}), for the real life experiments. The density constraint ensures incompressibility for the reconstructed liquid and should be 0 when the constraint is satisfied. These results show that when applying our constraint solver, the incompressibility property is met. Meanwhile Schenck \& Fox's constraints were unable to reach similar performance.}
    \label{table:cavity_results_density}
\end{table*}

\clearpage

\begin{table*}[ht!] 
    \vspace{-6.5in}    
    \centering
    \fontsize{9}{11}
    \begin{tabular}{ c | c | c | c | c | c }
     \fontsize{10}{12}{\textbf{Method}} & Simulation & Endo Trail 1 &  Endo Trial 2 & Endo Trail 3 & Pouring \\ \hline
     No Constraints \cite{lassner2021pulsar} & $0.469 \pm 0.133$ & $0.907 \pm 0.034$ & $0.904 \pm 0.058$ & $0.871 \pm 0.063$ & $0.798 \pm 0.167$ \\ \hline
     No Density  & $0.822 \pm 0.096$ & $0.915 \pm 0.055$ & $0.874 \pm 0.214$ & $0.919 \pm 0.025$ & $0.867 \pm 0.054$ \\  \hline
     No Collision  & $0.410 \pm 0.241$ & $0.904 \pm 0.026$ & $0.899 \pm 0.048$ & $0.761 \pm 0.177$ & $0.869 \pm 0.078$  \\ \hline
     Schenck \& Fox \cite{schenck2018spnets}  & $0.217 \pm 0.113$  & $0.437 \pm 0.319$  & $0.882 \pm 0.052$ & $0.830 \pm 0.100$ & $0.039 \pm 0.221$  \\ \hline
     DSS \cite{yifan2019differentiable}  & $0.759 \pm 0.123$ & $0.916 \pm 0.042$ & $0.909 \pm 0.081$ & $0.917 \pm 0.032$ &  $0.815 \pm 0.079$  \\ \hline
     Uniform  & $0.826 \pm 0.127$ & $0.900 \pm 0.056$ & $0.891 \pm 0.061$ & $0.891 \pm 0.071$ & $0.576 \pm 0.205$  \\ \hline
     No Prediction  & $0.896 \pm 0.049$ & $0.904 \pm 0.031$ & $0.899 \pm 0.046$ & $0.905 \pm 0.020$ & $0.890 \pm 0.084$ \\ \hline
     Ours  & $0.889 \pm 0.049$ &   $0.902 \pm 0.054$  & $0.905 \pm 0.051$ & $0.910 \pm 0.026$ &  $0.849 \pm 0.071$ \\ \hline
     Our Source  & $0.891 \pm 0.041$ & $0.911 \pm 0.034$ & $0.908 \pm 0.055$ & $0.913 \pm 0.026$ &  $0.843 \pm 0.061$ \\ 
    \end{tabular}
    \caption{Mean and standard deviation of IoU for the real life experiments. The results show that our reconstruction approach is able to achieve comparable image loss performance as the best from No Constraints, No Density, No Collision and No Prediction comparisons. This implies that our approach is effective at converging in image loss with additional constraints (density and collision) and prediction.}
    \label{table:cavity_results_iou}
\end{table*}

\end{document}